\pdfoutput=1

\documentclass[11pt]{article}

\usepackage[final]{acl}

\usepackage{times}
\usepackage{latexsym}

\usepackage[T1]{fontenc}

\usepackage[utf8]{inputenc}

\usepackage{microtype}

\usepackage{inconsolata}

\usepackage{graphicx}

\usepackage{booktabs}
\usepackage{amsmath}
\usepackage{subcaption}
\usepackage{multirow}
\usepackage{tcolorbox}
\usepackage{xcolor}
\newcommand{\hlr}[1]{\begingroup\setlength{\fboxsep}{0.6pt}\colorbox{red!20}{#1}\endgroup}

\title{SOS: Analyzing Surface over Semantics in multilingual multimodal AI}

\title{DeepForm}

\title{Core or Cover: Analysis of Surface over Semantics in multilingual multimodal AI}

\title{\emph{SoS}: Analysis of Surface over Semantics in Multilingual Multimodal AI}

\title{\emph{SoS}: Analysis of Surface-over-Semantics \\in Multilingual Text-To-Image Generation}

\author{
  \textbf{Carolin Holtermann\textsuperscript{1}},
  \textbf{Florian Schneider\textsuperscript{2}},
  \textbf{Anne Lauscher\textsuperscript{1}}
\\
  \textsuperscript{1}Trustworthy AI Lab, University of Hamburg \\
  \textsuperscript{2}Language Technology Group, University of Hamburg \\
\\
  \texttt{carolin.holtermann@uni-hamburg.de}
}

\begin{document}

\maketitle
\begin{abstract}
Text-to-image (T2I) models are increasingly employed by users worldwide. However, prior research has pointed to the high sensitivity of T2I towards particular input languages -- when faced with languages other than English (i.e., different surface forms of the same prompt), T2I models often produce culturally stereotypical depictions, prioritizing the surface over the prompt's semantics. Yet a comprehensive analysis of this behavior, which we dub \emph{Surface-over-Semantics (SoS)}, is missing. We present the first analysis of T2I models' SoS tendencies. To this end, we create a set of prompts covering 171 cultural identities, translated into 14  languages, and use it to prompt seven T2I models. To quantify SoS tendencies across models, languages, and cultures, we introduce a novel measure, and analyze how the tendencies we identify manifest visually. We show that all but one model exhibit strong surface-level tendency in at least two languages, with this effect intensifying across the layers of T2I text encoders. Moreover, these surface tendencies frequently correlate with stereotypical visual depictions.
\textcolor{red}{\textit{**Warning: This paper contains discussions about stereotypes**}}


\end{abstract}

\section{Introduction}
With the public release of text-to-image (T2I) tools like Dall-E~\cite{pmlr-v139-ramesh21a}, automatically generating images via prompting, e.g., for marketing purposes, has become a widely adopted practice for many users worldwide. 
Given that only about 20\% of the global population speaks English fluently\footnote{\url{https://www.ethnologue.com/insights/most-spoken-language/}}, many of these users may prompt the systems in languages other than English. Accounting for this variety, while most T2I models are still primarily trained and tested on English only~\cite[e.g., Stable Diffusion v2.1;][]{sdmodel}, several multilingual models have been presented~\cite[e.g., Kandinsky-3;][]{vladimir-etal-2024-kandinsky3}, enabling wider linguistic inclusion. 

While intuitively, semantically equivalent prompts should produce similar outputs, previous work demonstrated that T2I models are sensitive to the particular input language: given the same prompt and its translation, the output may be \emph{highly} different -- often reflecting cultural stereotypes associated with the input languages~\cite[e.g.,][]{ventura2024navigatingculturalchasmsexploring}. As such, \citet{homoglyphs} found that exchanging a single prompt character with a homoglyph from a different script may activate depictions of cultural elements tied to that character. 
Importantly, these studies demonstrate the existence of a \emph{tension between the surface form of an input (i.e., its language, its script, etc.), and the semantics of the prompt (i.e., the actual description of what should be visualized)}. However, so far, all studies operate on a handful of languages and/or cultures only, and, surprisingly, they fail to quantify the tendencies between prompt surface and semantics. This gap directly hinders further research on controlling how much the model is guided by either of the two, and the development of globally inclusive and culturally fair T2I models.


\begin{figure*}
    \centering
    \includegraphics[width=1\linewidth]{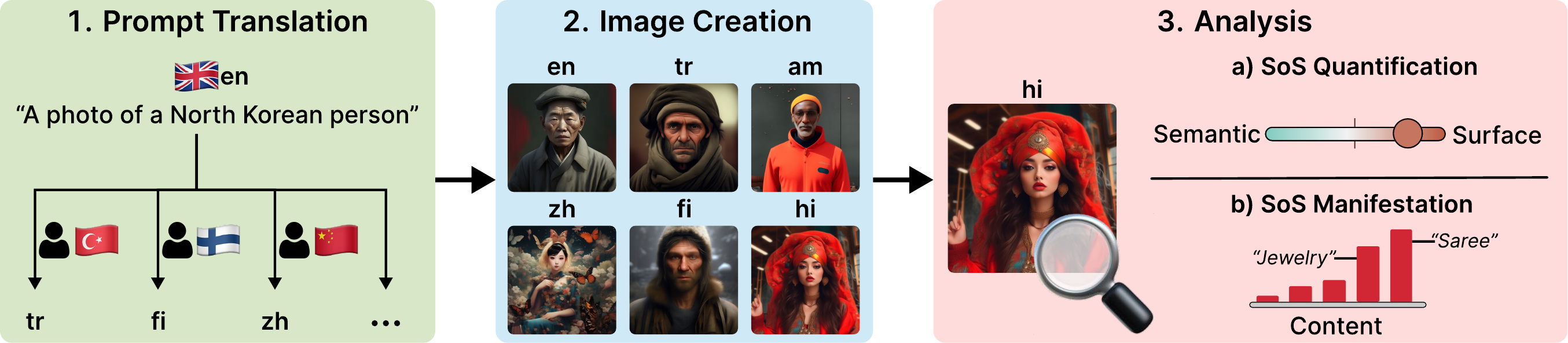}
    \caption{\textbf{Overview of our evaluation setup:} (1) Constructing and translating each prompt into 13 other languages; (2) Generating corresponding images using one of 7 T2I models; (3) Analyzing output images through the SoS Score, a color analysis, and an analysis of commonly occurring descriptive terms.}
    \label{fig:fig1}
\end{figure*}

We conduct the first systematic analysis of a model's alignment with the surface form of a prompt versus its semantic meaning. To do this, we create a dataset covering 171 cultures and 14 languages, with translations provided by native speakers. Using this dataset, we generate images with seven T2I models and analyze their reliance on the input language vs. a culture mentioned. To quantify the models' \emph{Surface-over-Semantics (SoS)} tendencies, we introduce a novel evaluation measure -- SoS score -- which allows us to compare models and languages. We then examine how bias toward a prompt's surface form manifests in visual outputs. Our study framework (Figure \ref{fig:fig1}) enables us to answer the following research questions (RQs):

\noindent \textbf{(RQ1) \emph{Can we quantify the tension between surface and semantics?}} Yes. Lacking a suitable method to analyze SoS tendencies of models, we propose the \emph{SoS score} (\S\ref{sec:rq1}), an embedding-based method that measures the similarity of T2I's generations towards average representations of a surface form and a semantic content of the input prompts. We show that it effectively captures SoS tensions in line with human perception.

\noindent \textbf{(RQ2) \emph{Do SoS tendencies differ across different models and languages?}} Yes. Applying our measure to quantify SoS tendencies~(\S\ref{sec:rq2}), we find that all but one of the tested models exhibit strong surface tendencies in at least two, and up to six, languages. These biases are amplified when generations are conditioned on representations from the upper layers of the text encoders in T2I models.

\noindent \textbf{(RQ3) \emph{Are observed SoS tendencies manifested in the concrete visual depictions generated?}} Yes. We couple our measurements with complementary analyses based on Visual Question Answering (VQA) (\S\ref{sec:rq3}). We find that languages with strong surface tendencies according to their SoS scores often trigger culturally stereotypical depictions.

\section{Related Work}
\paragraph{(Cultural) Bias in T2I Models}
Numerous studies have explored bias in T2I models; for a comprehensive overview, see \cite{wan2024surveybiastexttoimagegeneration}.

\noindent Most prior research focused on sociodemographic biases in T2I, examining dimensions such as gender, race, age, and nationality \cite{socialbiasesT2I, amplifydemographicstereotypes, ungless-etal-2023-stereotypes}. Mitigation techniques include fine-tuning on diversified datasets through counterfactual data augmentation \cite{brinkmann2023multidimensionalanalysissocialbiases}, and training on newly generated diverse instances \cite{esposito2023mitigatingstereotypicalbiasestext}.
Cultural biases have only recently gained attention. \citet{jha-etal-2024-visage} studied visual cultural stereotypes in depictions of individuals from different cultural backgrounds. \citet{kannen2025aestheticsculturalcompetencetexttoimage} introduced a benchmark to assess cultural awareness and diversity, identifying significant performance gaps for certain cultures. Similarly, \citet{partialityandmisco} presented a dataset of cultural concepts, including reference images to evaluate and fine-tune T2I models. However, these studies remain monolingual, overlooking the effect of the prompting language on image generation.

\noindent\textbf{Language-Induced Bias in AI Models} 
Another line of research explores how prompting language affects the output of generative models, particularly in LLMs.
\citet{romanou2024includeevaluatingmultilinguallanguage} proposed a benchmark to evaluate the factual and combinatorial knowledge of LLMs across 44 languages. Other studies explored shifts in models' cultural values due to multilingual model fine-tuning \cite{choenni-etal-2024-echoes}, variations in moral norms across languages \cite{haemmerl-etal-2023-speaking}, prompt language-specific stereotypes \cite{neplenbroek2024mbbqdatasetcrosslingualcomparison}, and LLM safety across multiple languages \cite{wang-etal-2024-languages}. Additionally, \citet{naous-etal-2024-beer} found a Western cultural bias when prompting LLMs in Arabic,  i.e., models favoring the generation of entities associated with Western traditions and values.

\noindent In contrast, work on language-induced bias in T2I remains limited. \citet{friedrich2024multilingualtexttoimagegenerationmagnifies} showed how multilingual prompting can amplify gender stereotypes in T2I. \citet{homoglyphs} found that minor character substitutions in textual prompts with non-Latin homoglyphs can induce strong cultural biases.
However, their analysis is restricted to a limited set of homoglyphs, thereby covering only a narrow subset of cultural variations. \citet{ventura2024navigatingculturalchasmsexploring} prompted T2I models with cultural concepts while translating parts of the prompt into ten languages. Their analysis relies on CLIP-based measures and VQA to assess cultural representation and national associations in images. We evaluate a broader and more diverse range of cultures and languages, and propose a novel evaluation measure that operates independently of textual descriptions.

\section{Dataset and Overall Setup}
We create a new multilingual prompt dataset in three steps: we (1) select 171 diverse cultures and 13 diverse languages (plus English); (2) define English prompt templates which we instantiate with explicit mentions of cultures; (3) let native speakers translate the prompts. Using these prompts, we generate images with seven state-of-the-art T2I systems and them with CLIP-like models~\cite{radford2021learning} for subsequent analyses.\footnote{We will release all data and code at \url{https://github.com/TAI-HAMBURG/Surface-Over-Semantics}.}

\vspace{0.3em}
\noindent\textbf{Selection of Cultures} 
Building on \citet{jha-etal-2024-visage}, we use the cultural identity groups compiled in the SeeGULL dataset \cite{jha-etal-2023-seegull}, which lists 176 terms. We refine this set by removing duplicates (e.g., \emph{``Nepali''} and \emph{``Nepalese''}) and standardizing names (e.g., \emph{``Netherlanders''} to \emph{``Dutch''}). These modifications result in a final set of 171 unique cultural identity groups ($C$).

\vspace{0.3em}
\noindent\textbf{Selection of Languages} 
Next, we select 13 languages for translating the English prompts, based on two key criteria: \textbf{(1) Model Coverage:}~while most T2I models do not advertise multilingual support, we include three models in our analysis that explicitly support a small set of non-English languages, to enable fair comparison. 
\textbf{(2) Linguistic Diversity:}~to ensure the robustness and generalizability of our findings, we increase the typological diversity of our dataset. Therefore, following \citet{ploeger2024typological}, we incorporate not only languages from diverse language families but also those that achieve a typological feature coverage of more than 81\% of the typological features recorded in the Grambank corpus \citep{grambank}. We are thus confident that our dataset provides a comprehensive basis for evaluating multilingual performance. Specifically, we include seven Indo-European Languages: English (en), Russian (ru), German, (de), French (fr), Italian (it), Spanish (es), and Hindi (hi); two Afro-Asiatic languages: Arabic (ar) and Amharic (am); and one language each from the Sino-Tibetan (Chinese (ch)), Turkic (Turkish (tr)), Koreanic (Korean (ko)), Japonic (Japanese (ja)), and Uralic (Finnish (fi)) language families. 

\vspace{0.3em}
\noindent\textbf{Template Construction and Initialization} 
To account for variations due to the exact prompt formulation, we define three base templates that differ slightly in wording, shown in Table \ref{tab:prompttemplates}. Each template consists of an introductory phrase (e.g., ``\emph{A photo of}''), a cultural identity term $c_i \in C$, and one of three person terms $p_j \in P$: \emph{``person''}, \emph{``woman''}, \emph{``man''}. Finally, initializing the templates with all possible $P \times C$ combinations yields 1,539 unique English prompts. 

We deliberately constrain prompt structure across prompts to minimize cross-linguistic variation. Nevertheless, we perform a robustness analysis with respect to prompt formulation, which shows low result deviations, indicating that more diverse prompt formulations would likely yield similar results. Details are provided in Appendix \ref{app:robustness}.

Note that for this analysis, we focus on depictions of people for several reasons. First, we consider stereotypes in depictions of people to be particularly harmful to users and socially consequential. Second, this choice lets us leverage cultural stereotypes identified by prior work, which mostly focuses on people. Third, given the limited transparency around training data, we expect depictions of people to be common in the pretraining corpus, reducing concerns about data scarcity that might confound our results. However, to verify the generalizability of our findings beyond human subjects, we perform additional experiments for an object category (\emph{'house'}).

\setlength{\tabcolsep}{18pt}
\begin{table}[t]
    \centering
    \small
    \begin{tabular}{cl}
    \toprule
        \textbf{ID}  &  \textbf{Template}\\
        \midrule
        a & \emph{A photo of a \textit{\{$c_i$\}}  \textit{\{$p_j$\}}}\\
        b & \emph{A \textit{\{$c_i$\}}  \textit{\{$p_j$\}}}\\
        c & \emph{A photorealistic image of a \textit{\{$c_i$\}}  \textit{\{$p_j$\}}}\\
        \bottomrule
    \end{tabular}
    
    \caption{\textbf{Base templates} used for the construction of our data set. Upon initialization, the placeholders \{$c_i$\} and \{$p_j$\} are filled with a cultural identity term (e.g., \emph{``German''}) and a person term (e.g., \emph{``person''}).}
    \label{tab:prompttemplates}
\end{table}

\vspace{0.3em}
\noindent\textbf{Prompt Translation}
We hire native speakers fluent in both English and the target language to translate the prompts. Annotators are recruited via the authors' own networks or an annotation platform,\footnote{\url{www.prolific.com}}, and fairly compensated (>10USD/hour). We explicitly instruct them to provide concise translations and to preserve distinctions between female, male, and gender-neutral person terms. We provide details on the languages, annotation task, and annotator demographics in Appendix \ref{app:annotation}.

\vspace{0.3em}
\noindent\textbf{Image Generation} 
Finally, we prompt seven T2I models to create an image for each prompt using a fixed random seed of 42. We deliberately include state-of-the-art models primarily designed for English usage as well as multi- and bilingual models. Specifically, we evaluate the following multilingual models: AltDiffusion-m9~\cite[\textsc{AD};][]{Ye2023AltDiffusionAM}, Kandinsky-2-1 \cite[\textsc{K21};][]{razzhigaev-etal-2023-kandinsky} and Kandinsky-3~\cite[\textsc{K3};][]{vladimir-etal-2024-kandinsky3}; alongside models not explicitly advertised as multilingual: Stable Diffusion v2.1~\cite[\textsc{SD21};][]{sdmodel}, Stable Diffusion XL~\cite[\textsc{SDXL};][]{podell2023sdxlimprovinglatentdiffusion}, Stable Diffusion 3~\cite[\textsc{SD3};][]{esser2024scalingrectifiedflowtransformers} and FLUX.1-dev~\cite[\textsc{FX};][]{flux2023}. An overview of model architectures is provided in Appendix \ref{app:dataset_exp_setup}.

\vspace{0.3em}
\noindent\textbf{Image and Text Embeddings}
All embeddings used in this work are obtained using a LAION CLIP model (\textsc{CLIP-ViT-bigG-14-laion2B-39B-b160k}). We validated all results from the experiments in \S\ref{sec:rq1} and \S\ref{sec:rq2} with an alternative CLIP-based model and non-CLIP-based models, yielding similar outcomes (see Appendix \ref{app:add_sos_score}).

\

\section{How to quantify the tension between surface and semantics?}
\label{sec:rq1}
While previous work either showed that T2I models often produce stereotypical depictions given a cultural identity term (i.e., \emph{semantic} tendency), or culturally biased outputs given a particular language (i.e., \emph{surface} tendency), we focus on the tension between surface and semantics. 

\subsection{Surface-over-Semantics (SoS) Score}
\paragraph{Motivation}
Lacking a suitable method to analyze SoS tendencies, we propose a new score, inspired by existing embedding-based similarity measures. In the realm of language-vision studies, the CLIPScore~\citep{hessel-etal-2021-clipscore} is perhaps the most widely used. It measures the similarity between an image and a textual description in a shared embedding space, thereby aligning semantically equivalent inputs. For T2I models, a higher CLIPScore indicates a ``better'' representation of the input.

However, the traditional CLIPScore has two limitations: {\color{black}
(\emph{i}) it measures image–caption alignment and is therefore only meaningful for higher-quality generations—precluding analysis of SoS dynamics across T2I layers, where early text-encoder outputs often resemble abstract noise; and (\emph{ii}) it is ultimately constrained by the representation quality of the languages within the CLIP model~\citep{radford2021learning, saxon-wang-2023-multilingual}, which may pose a non-negligible confounding factor, especially for resource-lean languages.} Instead, we propose an evaluation measure that operates solely on the generated images (and is thus reduces language-dependency), is robust to both lower- and higher-quality generations (and thus broadly applicable), and directly captures the SoS tension by comparing outputs against references representing surface form and semantic meaning.

\vspace{0.3em}
\noindent\textbf{Definition} Our score compares each output image $o_{c,l}$ generated for language $l$, and cultural identity $c$ with two reference vectors: $\Vec{\textnormal{sur}_l}$ representing surface-form generations, and $\Vec{\textnormal{sem}_c}$, representing semantic identity generations. Concretely, let $\Vec{\textnormal{sur}_l} = \frac{1}{|O_{l}|} \sum_{o \in O_{l}} \Vec{e_o}$ be the averaged embedding vector $\Vec{e_o}$ of all images $o \in O_{l}$ generated from input prompts in language $l$, e.g., \emph{Finnish}. Conversely, let  $\Vec{\textnormal{sem}_c} = \frac{1}{|O_{c}|} \sum_{o \in O_{c}} \Vec{e_o}$ be the average vector of all embeddings $\Vec{e_o}$ for images $o \in O_{c}$ that were generated for input prompts that target a specific cultural identity $c$, e.g., \emph{depicting German individuals}. {\color{black} To mitigate model-specific biases, we estimate these reference vectors by pooling images across all models rather than per model.} For an individual output $o_{c,l}$ (e.g., for the prompt \emph{``A German person''}), the $\textnormal{SoS}_{c,l}(o_{c,l})$ score is defined as the difference between its embedding similarities with the two reference vectors $\Vec{\textnormal{sem}_{c}}$ and $\Vec{\textnormal{sur}_{l}}$:

\vspace{-0.6em}
\begin{small}
\begin{equation}
   \textnormal{SoS}_{c,l}(o_{c,l}) = \cos(\Vec{\textnormal{sem}_{c}}, \Vec{e_o}) - \cos(\Vec{\textnormal{sur}_{l}}, \Vec{e_o})\,,
\end{equation} 
\end{small}

\vspace{-0.3em}
\noindent with $\Vec{e_o}$ the embedding of output $o_{c,l}$. The score lies in $[-1,1]$: negative values indicate stronger alignment with the surface, positive values with the semantics of the input. Finally, the joint SoS-score for a language–culture pair $(c,l)$ is the mean of $\textnormal{SoS}_{c,l}(o_{c,l})$ over all outputs $o_{c,l}$ generated for $(c,l)$ (i.e., across the base templates and person terms).

\subsection{Validation}
\paragraph{Human Validation} To validate our score, we conduct a human annotation study on a subset of generated images. After computing average SoS per culture–language–model, we use stratified sampling to select 50 representative groups spanning the SoS range. For each group, annotators evaluate all images from nine prompts (3 base templates $\times$ 3 person terms), for a total of 450 images.

Validating the score through human judgment is challenging, as assessments are inherently subjective and influenced by cultural background. To minimize biases, we take several steps: (1) We recruit three annotators with diverse cultural backgrounds (German, Indian, and Chinese) to ensure broader perspectives. (2) We design a simple task in which each image is paired with five options: one referring to the cultural identity mentioned (\emph{semantics}), one corresponding to a culture matching prompt language (\emph{surface}), and three randomly sampled alternatives from the remaining 169 cultures. Annotators select the option that, in their view, best describes the image. (3) We determine the final label by majority vote and compare it with the label predicted based on the SoS score. 

The annotation task yielded a moderate Fleiss' $\kappa$ of $55.1\%$, reflecting the inherent difficulty of the annotation task. We attribute this primarily to two factors. First, following prior work, we use country names as cultural proxies, though culture is multidimensional and does not map neatly to national boundaries. Second, while our diverse annotator pool introduces valuable multiple perspectives, it also produces varying intuitions and familiarity levels with the cultural options. Crucially, a finer-grained analysis shows substantially higher agreement on our main dimensions of interest, surface and semantic tendencies (up to 77\% Cohen's $\kappa$), while most disagreement is concentrated in the culture distractor labels. This pattern validates that annotation quality is sufficient for our proposed metric, particularly along its core dimensions. Further details on the annotation task and validation appear in Appendix \ref{app:annotation}. 

\vspace{0.3em}
\noindent\textbf{Comparison with CLIPScore} 
We compare the SoS scores against an annotation procedure based on CLIPScores. Following \citet{ventura2024navigatingculturalchasmsexploring} and \citet{homoglyphs}, we use a simple caption template (\emph{`A photo of a $c_i$ person'}), filling $c_i$ with either the cultural identity from the original prompt or the identity matching the prompt language. Next, we embed the image and the candidate captions, calculate the CLIPScores following \citet{hessel-etal-2021-clipscore}, and select the identity with the higher score. 


\vspace{0.3em}
\noindent\textbf{Results} 
Table \ref{tab:validation_results} summarizes the validation results. 
Overall, the SoS score achieves $74.0\%$ accuracy compared with human annotations, slightly below CLIPScore’s $78.2\%$, but shows higher precision ($94.8\%$) in capturing surface-level tendencies. A closer analysis reveals lower SoS accuracy for English prompts, likely due to greater image diversity and less reliable reference embeddings, while CLIPScore underperforms on all non-European languages. These results confirm the robustness of our approach, while at the same time providing wider applicability. In the Appendix \ref{app:robustness}, we provide the full per-language breakdown and additional experiments with a multilingual CLIP model, which still underperforms SoS on non-English languages.


\setlength{\tabcolsep}{10pt}
\begin{table}[t]
    \centering
    \small
    \begin{tabular}{lccc}
    \toprule
        \textbf{Comparison}  &  \textbf{Acc} & $\mathbf{P_{\textnormal{sur}}}$ & $\mathbf{P_{\textnormal{sem}}}$ \\
        \midrule
        SoS score & 74.0\% & 94.8\% & 84.8\% \\
        CLIPScore  & 78.2\% & 86.8\% & \textbf{95.4}\% \\
        \midrule
        SoS  $\setminus$  en & \textbf{79.1}\% & \textbf{94.8}\% & 94.8\%\\
        CLIPScore  $\setminus$ en & 76.3\% & 86.8\% & 94.3\% \\
         \bottomrule
    \end{tabular}
    \caption{\textbf{SoS score validation results:} Comparing SoS score and CLIPScore to human judgments, reporting accuracy (Acc) and precision (P) for predicting SoS tendencies, with and without English prompts.}
    \label{tab:validation_results}
\end{table}


\section{Which SoS tendencies can be observed across different models and languages?}
\label{sec:rq2}
We study SoS across models and languages, and its evolution throughout different model layers. 

\subsection{Analysis Setup}
\paragraph{SoS Tendencies per Model and Layer-Wise Analysis} With a suitable measure at hand (\S\ref{sec:rq1}), we first quantify the overall SoS tendencies of each model given a cultural identity and language combination. Next, we investigate the evolution of these tendencies throughout the T2I models' text encoders. To this end, we apply DiffusionLens~\cite{toker-etal-2024-diffusion}. 
Thus, we extract the prompt representation from selected layers of the text encoder, perform layer normalization, and feed these representations into the diffusion model. In our experiments, we generate images using the representation from every fourth layer. Due to resource constraints, we perform these experiments exemplarily for two models with distinct encoders: \textsc{SD21}, which utilizes OpenCLIP ViT-H, and \textsc{K3}, which employs Flan-UL2. 
As we are mostly interested in analyzing the effect of multilingual prompting, we perform the layer-wise analysis across all languages of the dataset except English. To ensure robustness, we generate images using four random seeds per layer. Since images based on earlier layers are not directly comparable with those of subsequent layers, we calculate the semantic and surface reference vectors separately for each layer and model.

\vspace{0.3em}
\noindent\textbf{SoS Tendencies per Language}
To analyze SoS tendencies across languages, we calculate language-specific SoS scores and assess potential inter-language correlations. Therefore, we compute Pearson correlations of the SoS scores for images generated using all language combinations.

\begin{figure}
    \centering
    \includegraphics[width=0.8\linewidth]{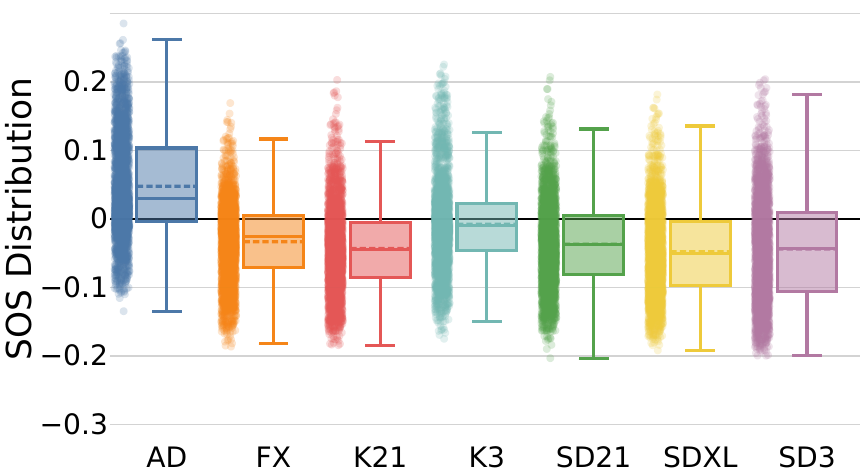}
    \caption{\textbf{SoS Score Distribution} with upper and lower quantiles, along with the mean (dotted line) and median (solid line) for each model.}
    \label{fig:sos_model_comparison}
\end{figure}

\subsection{Results}
\paragraph{Models are mostly guided by the surface form of a prompt with AD as the exception.}
We present the distribution of SoS scores per model across all cultures and languages in Figure \ref{fig:sos_model_comparison}. Although all models except AD ($0.05$) show a negative mean SoS score of as low as $-0.048$ for SDXL, we observe slight differences in their value distribution. While AD features mostly positive SoS scores, the mean differs more from the median compared to the other models, indicating the presence of notable outliers with a surface tendency. 
The bilingual K3 exhibits the smallest negative mean SoS score ($-0.007$), with values nearly normally distributed around zero. In contrast, its predecessor, K21, also bilingual, shows a stronger inclination towards the surface with a mean score of $-0.043$. Intriguingly, despite SD3’s strong overall surface tendency, defined as having a median SoS score at or below the 25th percentile across model–language pairs, some images display the opposite, showing a more pronounced semantic tendency with an outlier corrected value of $0.18$. The model's wide SoS value range suggests a tendency to amplify stereotypes, whether in surface or semantic alignment.

\begin{figure}
    \centering
    \includegraphics[width=\linewidth]{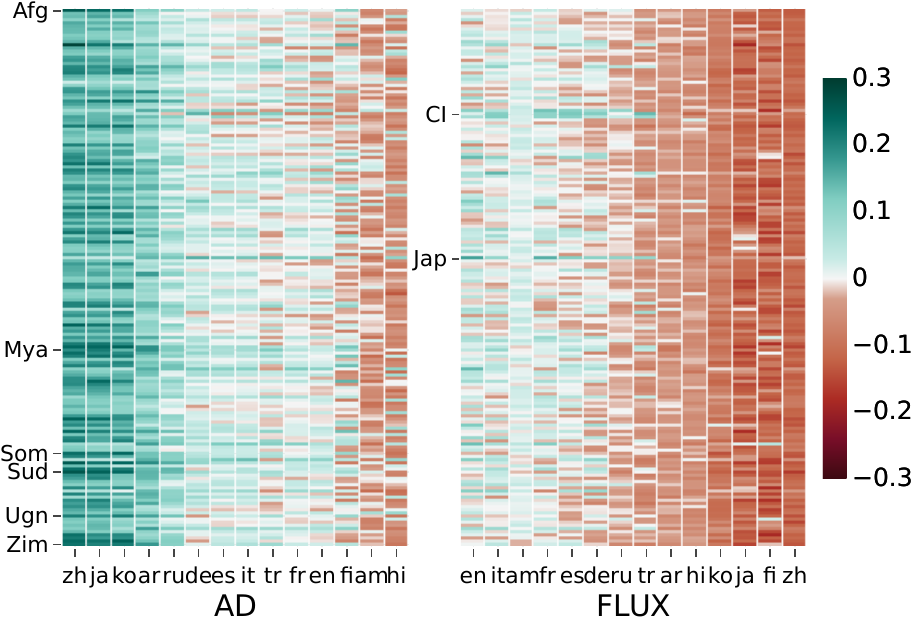}
    \caption{\textbf{SoS Score Heatmap} averaged across templates and person terms for AltDiffusion (left) and FLUX (right). Rows depict each culture, and columns are sorted by the mean SoS score per language.}
    \label{fig:sos_small}
\end{figure}

\vspace{0.3em}
\noindent\textbf{All but one model exhibit strong surface tendencies in at least two languages.} 
Figure \ref{fig:sos_small} shows SoS scores for the multilingual AD and monolingual FX models across different cultural identities (y-axis) and languages (x-axis). We provide additional results for the remaining models in uncompressed form, as well as analyses with alternative embeddings~\citep[e.g., DINO;][]{oquab2024dinov2learningrobustvisual} in Appendix~\ref{app:add_sos_score}.

We find that all models except AD exhibit a strong surface tendency in at least two languages. We define surface tendency as \textit{strong} for a given model-language pair if the median SoS score falls at or below the 25th percentile of median SoS scores across all model-language pairs. Overall, AD demonstrates a semantic tendency for most languages and cultures. However, for Hindi, Amharic, and Finnish prompts, the generations are more guided by the input language, which aligns with the fact that these languages were not part of the model's explicit training. For AD, SoS tendencies also vary by cultural identity, whereas FX displays a more uniform pattern, with the main exceptions of Japanese and Christmas Island. Notably, certain cultures-- such as Myanmar, Sudanese, Afghan, and Zimbabwean -- show a greater semantic tendency for AD compared to other cultures, a pattern particularly pronounced for African cultures. FX shows strongly negative SoS tendencies for Korean, Finnish, Japanese, and Chinese, while displaying a weaker bias for Amharic compared to AD.

\begin{figure}
     \centering
     \begin{subfigure}[b]{0.8\linewidth}
         \centering
         \includegraphics[width=1\textwidth]{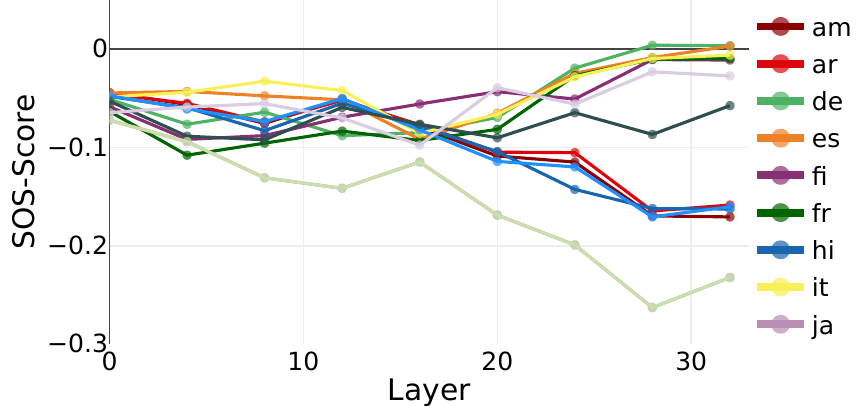}
         \caption{Kandinsky-3}
     \end{subfigure}
     \hfill
     \begin{subfigure}[b]{0.8\linewidth}
         \centering
         \includegraphics[width=\textwidth]{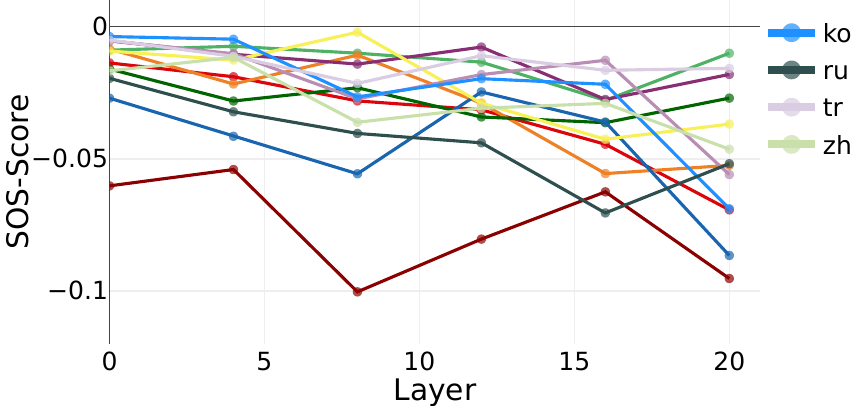}
         \caption{Stable Diffusion 2.1}
     \end{subfigure}
        \caption{\textbf{Averaged SoS scores per language across different text encoder layers.} Colors indicate input languages, shown for (a) K3 and (b) SD21.}
        \label{fig:layer_analysis}
\end{figure}

\vspace{0.3em}
\noindent\textbf{Negative SoS tendencies become more pronounced in later text encoder layers.} We present the results of our layer-wise analysis for \textsc{SD21} and \textsc{K3} in Figure \ref{fig:layer_analysis} (average SoS score per layer and language, aggregated across seeds and cultural identities). Interestingly, both models exhibit only a slightly negative score in early layers, except for Amharic in SD21. For later layers, the bias toward the prompt surface form becomes more pronounced. Notably, for \textsc{K3}, the SoS scores for European languages tend to shift toward neutrality from layer 20 onward, whereas those for most Asian languages move toward a more negative SoS score. This pattern could be explained by the finding of \cite{rogers-etal-2020-primer}, who found that semantic encoding primarily occurs in later text encoder layers. Our results thus suggest that when a text encoder lacks sufficient exposure to a particular language, it is more likely to default to surface-level cues rather and that exactly this tendency is further amplified than capturing deeper semantic structures.

\begin{figure}[t]
    \centering
    \includegraphics[width=0.8\linewidth]{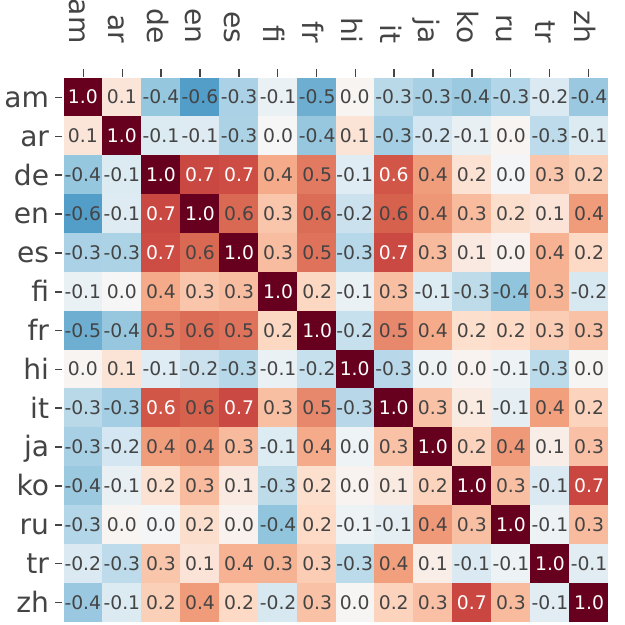}
    \caption{\textbf{Pearson correlation of SoS scores between languages} across all models and cultural identities.}
    \label{fig:corr_all_models}
\end{figure}

\vspace{0.3em}
\noindent\textbf{SoS score correlations highlight strong Latin script similarities.}
Figure \ref{fig:corr_all_models} presents the Pearson correlations for SoS scores obtained across models and cultures for all language pairs. We provide additional results and per-model breakdowns in Appendix \ref{app:linguistic_similarities}. Overall, Amharic shows strong negative correlation with most languages, especially English, while Arabic exhibits very weak ($0.1$) to negative ($-0.4$) correlations with other languages, marking them as outliers. In contrast, Indo-European languages spoken in Europe, except Russian, show strong correlations above 0.5, with very high correlations of 0.7 between Italian and Spanish, but also English and German. This indicates a similar SoS tendency for languages in the Latin script. 
Interestingly, Chinese shows a high correlation with Korean of $0.7$, while its correlation with other languages is significantly lower. One might argue that this could be influenced by general patterns due to script similarity; however, the correlation between Chinese and Japanese is much lower at $0.3$. In fact, Japanese exhibits a higher correlation with Russian than with Chinese and also differs in its distribution within the vector space. Thus, the extent to which models adhere to semantic or surface tendencies is not necessarily dictated by similarities in scripts. Note that this analysis captures only similarities in the score distribution patterns, not in their visual realization. The same SoS tendency can manifest in generated images in many different ways.

\paragraph{Similar patterns emerge for other concepts.} 
To evaluate the robustness of our findings and the extent to which the SoS score generalizes beyond the \textit{person} concept, we replicate the analysis for a second, culturally grounded concept, namely \textit{house}. Using an analogous prompt template with three paraphrased variants, we compute SoS scores for \textit{house} across a subset of seven languages. The resulting scores are strongly correlated with those for \textit{person} across models (Pearson $r=0.876$), indicating that the SoS tendencies observed across languages and cultures are not specific to depictions of people and further supporting the robustness of the metric. We provide the results in Appendix \ref{app:robustness}.

\section{How do negative SoS tendencies manifest in concrete visual depictions?}
\label{sec:rq3}
{\color{black}After quantifying models' tendencies toward the prompt’s surface form rather than its semantic meaning, we next analyze how this tendency manifests visually by examining textual descriptions of images generated in languages showing greater surface-level associations.

\subsection{Analysis Setup}
\paragraph{VQA Analysis} Following \citet{jha-etal-2024-visage}, we generate and analyze image descriptions, but without restriction to predefined terms. We prompt the VQA model \texttt{Qwen2-VL-7B-Instruct} with \textit{``Describe this image in detail''} (1024-token limit), and tokenize the outputs to extract unique terms per image. To identify visual cues distinctive to a language–model pair, i.e., tokens that occur more frequently when a model is prompted in a certain language, we apply the Fighting Words method by \citet{Monroe_Colaresi_Quinn_2017} with inverse document frequency (IDF) smoothing to downweight generic terms. Concretely, we treat all terms identified per language-model pair as a document, compute IDF token values, and weight token counts accordingly. We then calculate the weighted log-odds ratios per token and document using the pooled set of terms for all other languages as the prior distribution. We rank terms by z-scores and extract the top 10 significant terms per document.

To validate the precision of the VQA-derived terms, we conduct a human annotation on a subset of images. First, we remove terms challenging to visually discern from the list (e.g., \emph{``updo''}). Next, we stratifiedly sample 300 images across all language-model pairs, along with six candidate terms per image, which include the correct ones plus distractors. An annotator selects the terms that best describe the image, and we measure precision as the proportion of true positives among detected terms. We achieve 96.4\% precision, with most errors involving attributes that are inherently difficult to visualize (e.g., \emph{``narrow''}), confirming the general suitability of VQA for our task. 

\paragraph{Coverage of stereotypical terms} 
To examine whether the identified depictions reflect cultural stereotypes, we compare the terms strongly associated with a given language–model combination against the visual stereotypes from the SeeGULL dataset \cite{jha-etal-2024-visage}. We merge all stereotypes of cultures associated with the respective input language and report proportional coverage. Further details are provided in the Appendix~\ref{app:vqa_analysis}.}

\begin{figure}
    \centering
    \includegraphics[width=0.8\linewidth]{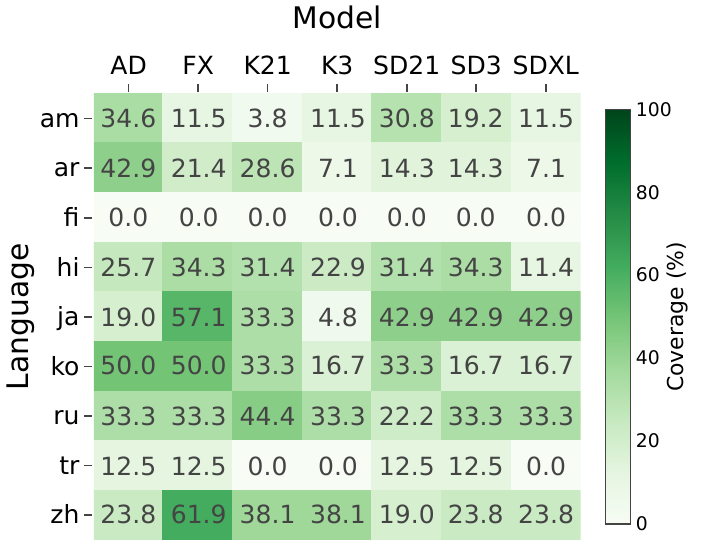}
    \caption{\textbf{Coverage of SeeGULL stereotypes}, showing the percentage of visual stereotypes of the SeeGULL dataset detected in the VQA analysis.}
    \label{fig:seagul_cov}
\end{figure}

\subsection{Results}

\vspace{0.3em}
\noindent\textbf{Negative SoS scores are associated with common cultural stereotypes.}
Figure~\ref{fig:seagul_cov} reports the percentage coverage of visual stereotypes for each language–model pair, with the corresponding term lists in Table~\ref{tab:seagull_terms}. Stereotypically associated terms appear in nearly all pairs, but coverage varies substantially by both language and model. For instance, FLUX generations in Chinese cover 61.9\% of Chinese visual stereotypes, whereas none of the Finnish stereotypes are represented. This discrepancy partly reflects the limited representation and cultural diversity of stereotypes in certain languages within the dataset. In Finnish, for instance, the sole visual stereotype is \textit{sauna}, while in Turkish, nearly half of the stereotypes capture highly similar concepts (e.g., \textit{greasy}, \textit{angry}, \textit{fiery}, \textit{violent}). Moreover, associations learned by T2I models may differ from human ones. These findings underscore the need for less restrictive and more nuanced approaches to analyze distinctive associative patterns.

\begin{figure}
    \centering
    \includegraphics[width=0.9\linewidth]{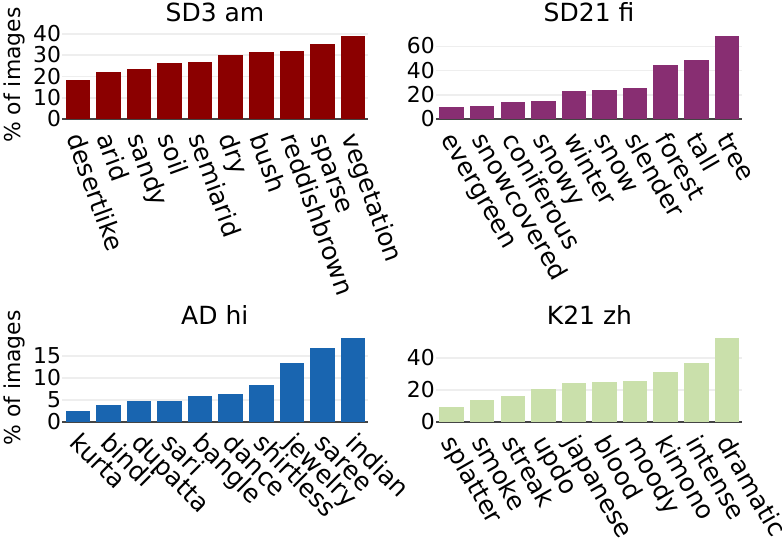}
    \caption{\textbf{VQA analysis results}, showing most frequent terms in image descriptions and their frequency (\%).}
    \label{fig:vqa_results}
\end{figure}

\noindent\textbf{Surface-level associations go beyond common stereotypes.} Figure~\ref{fig:vqa_results} presents, for selected model–language pairs, the top 10 significant ($z>1.96$) distinguishable VQA terms together with the percentage of images in which each term appears. For comprehensive per-language results across all models, see Appendix~\ref{app:vqa_analysis}. The results reveal pronounced and systematic patterns. For Finnish, models frequently generate snowy forest scenes; for example, SD depicts \emph{trees} in more than 68\% of images generated from Finnish prompts. Moreover, more than $20\%$ of images generated by SD3 for Amharic contain terms such as \textit{arid},\textit{sandy}, or \textit{bush}, resembling patterns identified by \citet{jha-etal-2024-visage} for African countries. Alarmingly, over 24.7\% K21-generated images for Chinese prompts include the term \textit{blood}. We find similar patterns for Japanese, Korean, and Arabic prompts, with images showing individuals with scars and blood, which raises concerns about the model's safety. Furthermore, we find that image descriptions for generations by K3 for Hindi, Japanese, Chinese, Korean, Arabic, and Amharic frequently contain terms like \textit{butterflies}, \textit{fantastical}, and \textit{dreamlike} (over $30\%$), suggesting that languages unfamiliar to the model are associated with this appearance. Finally, although AD exhibits the least average surface tendency, image descriptions for images prompted in Hindi still frequently include \textit{saree} ($>16\%$) and \textit{Indian} ($>19\%$), indicating persistent cultural associations. Notably, these cultural associations appear across all models except K3. Ultimately, our results suggest that relying on fixed stereotype lists can be limiting and that both SoS and Fighting Words provide valuable tools for uncovering such manifestations more comprehensively.

\section{Discussion and Conclusion}
The landscape of multilingual, open T2I models is not only scarce, users may face risks when prompting in their native languages.
While prior work shows that prompt language can induce stereotypical biases in T2I outputs \cite{ventura2024navigatingculturalchasmsexploring}, existing evaluation methods rely on a reference language whose representation may be low quality and biased (cf. CLIPScore). We address this issue with SoS, which does not rely on comparisons to textual inputs but solely on the generated images, thereby reducing language-specific biases and providing a more objective assessment.

Our analysis reveals that all but one model exhibit strong surface-level tendency in at least two languages, yet some cultures experience disproportionately high stereotyping. This tendency intensifies in the later layers of the text encoder, suggesting that enhanced multilingual training may reduce these effects. Additionally, measures like SoS could serve as a metric to guide model training, encouraging image generation that aligns more closely with fair outputs. Our open VQA-based analysis uncovers how these biases manifest visually, often reinforcing stereotypical biases associated with a culture linked to the input language. Additionally, the color analysis suggests that even stylistic elements can vary with prompt language.

The desirable SoS score is highly context-dependent. For culturally grounded prompts (e.g., \emph{house of the president}), language-specific grounding may be appropriate, while prompts like \emph{a medieval knight} should be guided by semantic meaning, independent of linguistic surface form. Thus, we advocate for future systems to dynamically adapt preferred scores based on prompt type.

\section*{Limitations}
We acknowledge the following limitations of our work: First, we deliberately limited the main scope of our analysis to images depicting persons. We decided to set this focus as the creation of images depicting people is a particularly sensitive application, where the amplification of stereotypical biases can lead to especially harmful outcomes. Furthermore, expanding the dataset to cover additional concepts would have significantly increased the experimental scope, limiting the depth of our analysis and potentially reducing the quality of our dataset. {\color{black}However, our complementary analysis on the concept 'house' yields results that are highly correlated with the main findings, demonstrating the robustness and generalizability of our approach across other concepts.} Second, although we aimed to include a typologically diverse set of languages in our dataset, our selection currently covers only a small fraction of the languages of the world. Still, we are convinced that most of the trends we find are generalizable to more languages. Third, our validation showed that the SoS score does not remain entirely reliable for English prompts, with a particular weakness in assessing semantic tendencies. Importantly, while measures similar to the SoS score may guide image generation toward more neutral and fair predictions, it is not an exhaustive fairness measure, and it needs to be coupled with other types of evaluations. {\color{black}Furthermore, the SoS score is not intended to measure the overall quality or factual accuracy of the generated image, but to reveal surface or semantic level tendencies in the model's image generations. In practical applications, it is thus essential to combine SoS with complementary metrics that assess generation quality and alignment with the prompt.} Ultimately, we call for considering the specific context and use case of an application for arriving at a meaningful interpretation.

\section*{Ethical Considerations}
To validate our findings, we collected human annotations, asking annotators to assign the most likely culture to each image. We acknowledge that each annotator brings their own (biased) views based on their individual cultural backgrounds and personal experiences. Although we hired annotators from three different backgrounds, we acknowledge that a more diverse workforce would better capture the full spectrum of cultural perspectives.

\section*{Acknowledgements}
The work of Carolin Holtermann and Anne Lauscher is funded by the Excellence Strategy of the German Federal Government and the Federal States.

\bibliography{custom}
\clearpage

\appendix


\section*{Appendix  Overview}
As the appendix is extensive, we provide a brief outline of its contents to facilitate navigation and enable a quick overview.

\noindent
\textbf{\hyperref[app:dataset_exp_setup]{A \quad Dataset and Experimental Setup}}

\textit{Additional information about the dataset creation and models used.}

\noindent
\textbf{\hyperref[app:annotation]{B \quad Annotation}}

\textit{Details on the annotation processes, including annotator demographics and annotation tasks.}

\noindent
\textbf{\hyperref[app:add_sos_score]{C \quad Additional SoS Score Results}}

\textit{Additional SoS results for all T2I models and using different embedding models.}

\noindent
\textbf{\hyperref[app:robustness]{D \quad Robustness Analysis}}

\textit{Experiments supporting the robustness of the score across other concepts and different prompt formulations.}

\noindent
\textbf{\hyperref[app:linguistic_similarities]{E \quad Linguistic Similarities}}

\textit{Analysis of image embedding distributions and SoS score language correlation analysis per T2I model.}

\noindent
\textbf{\hyperref[app:colour_analysis]{F \quad Colour Analysis}}

\textit{Experiments on colour similarities in image generation across prompt languages.}

\noindent
\textbf{\hyperref[app:vqa_analysis]{G \quad VQA Analysis}}

\textit{More fine-grained results on the VQA analysis, including distributions per language-model combination and detailed comparison to SeeGULL terms.}

\noindent
\textbf{\hyperref[app:example_images]{H \quad Example images}}

\textit{Example images generated by the T2I models.}

\onecolumn
\section{Dataset and Experimental Setup} \label{app:dataset_exp_setup}

\paragraph{Data Statement}
The main basis of our datasets is the list of cultural identity groups obtained from \cite{jha-etal-2023-seegull}, which was published under the CC-BY-4.0 license. 

\begin{table}[h]
    \centering
    \begin{tabular}{l|p{10cm}}
    \toprule
        Topic & Explanation \\
        \midrule
        Name &  SoS Evaluation Dataset \\
        Date Created & December 2024\\
        Languages Covered & English, Russian, German, French, Italian, Spanish, Hindi, Arabic , Amharic , Chinese, Turkish, Korean, Japanese, Finnish \\
        Purpose & The dataset was created to evaluate the language-induced biases of text-to-image models across cultural identity terms. It is intended for analyzing model biases and not for training purposes. \\
        Source Prompts & The original prompts (in English) were manually curated to reflect a diverse set of cultural identities and different genders. \\
        Translation Method & All prompts were translated by native speakers of each target language. Translators were instructed to preserve the gender mentioned in the prompts but adapt the structure whenever they see fit to preserve fluency. The translation task is shown below. \\
        Annotator Demographics & Translators were chosen based on their fluency in the respective target language and English. \\
        Limitations & Translators are not professional translators but contacted over a crowdsourcing platform or hired from within the authors network. The languages covered in the dataset are not exhaustive.  \\
        \bottomrule
    \end{tabular}
    \caption{Data Statement of our dataset.}
\end{table}

\paragraph{Experimental Setup}
All experiments within this work were run on a single NVIDIA A6000 GPU.

\begin{table*}[h]
    \centering
    \small
    \begin{tabular}{l|l|c|c}
    \toprule
      Modelname &   & Text Encoder & Diffuser \\
      \midrule
        FLUX Dev & FX &CLIP-G/14, CLIP-L/14, T5 XXL & MM-DiT \\
        AltDiffusion-m9 & AD & XLM-R & Latent Diffusion U-Net\\
        Kandinsky-2-1  & K21 & XLM-Roberta-Large-Vit-L-14, CLIP ViT-L/14 & Latent Diffusion U-Net\\
        Kandinsky-3  & K3 & Flan-UL2 & Latent Diffusion U-Net\\
        Stable Diffusion XL  & SDXL & CLIP ViT-L \& OpenCLIP ViT-bigG & Latent Diffusion U-Net\\
        Stable Diffusion v2.1 & SD21 & OpenCLIP ViT-H &  Latent Diffusion U-Net\\
        Stable Diffusion v3 & SD3 & CLIP-G/14, CLIP-L/14, T5 XXL & MM-DiT\\
        \bottomrule
    \end{tabular}
    \caption{Explanation of T2I model architectures.}
    \label{tab:models}
\end{table*}

\begin{table}[h]
\small
    \centering
    \begin{tabular}{ll c c}
    \toprule
      \multicolumn{2}{c}{\textbf{Language}}   & \textbf{L. Family}  & \textbf{Model Coverage}\\
       \midrule
       en & English  &   & all \\
       ru & Russian  &   & AD, K3, K21 \\
       de & German  &   &  AD \\
       fr & French  &   Indo-European  & AD \\
       it & Italian  &  & AD \\
       es & Spanish  &    & AD \\
       hi & Hindi  &     & - \\
       \midrule
       ar & Arabic & 	\multirow{2}{*}{Afro-Asiatic}  & AD\\
       am & Amharic &  & -\\       
       \midrule
       zh & Chinese  &   Sino-Tibetan  & AD \\
       tr & Turkish  &   Turkic  & - \\
       ko & Korean  &   Koreanic  & AD \\
       ja & Japanese  &   Japonic  & AD \\
       fi & Finnish  &   Uralic  & - \\
        \bottomrule
    \end{tabular}
    \caption{Languages we cover in our study together with their support by the T2I models we analyze (AltDiffusion-m9 (\textsc{AD}), Kandinsky-2-1 (\textsc{K21}), Kandinsky-3 (\textsc{K3})).}
    \label{tab:languages}
\end{table}

\twocolumn
\clearpage
\section{Annotation} \label{app:annotation}
This section details the two annotation tasks that were performed for our experiments. First, the translation of the template prompts into the 13 other languages. Second, the human annotation of the generated images by the T2I models, which was performed to validate the SoS scores.

\subsection{Native Speaker Translation}
\paragraph{Translation Task} 

We present the description of the translation task given to all native speakers to translate the prompts. To facilitate the process, we provide the annotators with machine-translated examples that they can adapt or keep.

\begin{tcolorbox}[colframe=blue!50!black, colback=blue!5!white, 
                  title=Translation Task, fonttitle=\bfseries]
Thank you for participating in this translation validation task. 

Task Description: You are provided with an Excel file containing sentences in English in column 'prompt in English' and their automatic translations to xxx in column 'prompt translation'. The sentences are easy and template-based, containing different cultural identifiers and identifiers of persons (person, man, \& woman). Your task is to validate the translations and insert '1' if the translation is correct, and insert the correct translation if the translation provided is incorrect.
Note: Please make sure that the translations of, e.g., 'A French man' and 'A French person' differ from one another and that one refers to a male person while the other refers to a gender-neutral person.
\end{tcolorbox}

\paragraph{Noise Reduction} 
To ensure consistency and reduce translation noise across languages, we explained the templated prompt format to each annotator and provided automatically translated examples as well as English examples as references. Annotators were instructed to preserve the structure where possible but also freely rephrase prompts if the direct translation sounded unnatural or unidiomatic in their language. In several cases, annotators explicitly reported that the automatic translation was correct but unnatural, and we adopted their reformulations in such cases. We therefore expect translation awkwardness to be low. Moreover, we explicitly asked annotators to maintain clear distinctions between ``person'', ``woman'', and ``man'' in their language to ensure comparability.

\paragraph{Annotator Demographics} 
We recruited annotators who are native speakers of thirteen different languages. An overview of their demographic backgrounds is provided in Table \ref{tab:native_speakers}.

\setlength{\tabcolsep}{4pt}
\begin{table*}[h]
\centering
\small
\begin{tabular}{l|lllllll}
    \toprule
 \textbf{Language}&  \textbf{Fluent languages}    &  \textbf{Age}&  \textbf{Sex}&  \textbf{Ethn.}&  \textbf{Country of birth}&  \textbf{Country of residence}&  \textbf{Nationality}\\
 \midrule
 Korean                      &  en,ko               &  33          &  Female      &  Asian                        &  Korea                    &  Canada                       &  Canada              \\
 Amharic                     &  am,en              &  56          &  Female      &  Black                        &  Ethiopia                 &  U.S.                &  U.S.       \\
 Arabic                      &  ar,en               &  27          &  Male        &  Other                        &  Sweden                   &  UK               &  UK      \\
 Hindi                       &  en,hi,ur,pa   &  34          &  Male        &  Asian                        &  India                    &  Mexico                       &  India               \\
 French                      &  fr,de,en,ru&  34          &  Female      &  White                        &  France                   &  Germany                      &  French              \\
 German                      &  de,en               &  28          &  Female      &  White                        &  Germany                  &  Germany                      &  German              \\
 Italian                     &  it,en              &  30          &  Male        &  White                        &  Italy                    &  UK                           &  Italian             \\
 Finnish &  fi,de,en       &  31 &  Female      &  White &  Finnland &  Germany &  Finnish    \\   
  Chinese &  zh,en       &  27 &  Male &  Asian &  Mainland China &  Germany &  Chinese    \\
  Turkish &  tr,en,de &  29 &  Male &  White &  Turkey &  Germany &  German    \\
  Russian &  ru,en &  30-40 &  Male &  White &  Russia &  Russia &  Russian    \\
  Spanish &  es,en,de &  55 &  Male &  White &  Germany &  Spain &  Spanish    \\
  Japanese &  en,ja,de &  27 &  Female &  Asian &  Japan &  Germany &  Japanese    \\
 \bottomrule
\end{tabular}
\caption{Demographic information of the native speakers who translated all prompts.}
\label{tab:native_speakers}
\end{table*}

\subsection{SoS Score Validation}

\paragraph{Annotator Demographics} We recruited three annotators (A1, A2, A3) with diverse cultural backgrounds for the SoS score validation task. All annotators had to be fluent in English. We will list more information about the annotators' demographics in Table~\ref{tab:annotator_demos}.

\setlength{\tabcolsep}{4pt}
\begin{table}[t]
\centering
\small
\begin{tabular}{l|c|c|c}
\toprule
\textbf{Attribute} & \textbf{A1} & \textbf{A2} & \textbf{A3} \\
\midrule
Age & 27 & 26 & 30 \\
Gender & m & m & m \\
Cult. Background & Chinese & German & Indian \\
Employ. type & PhD Cand. & BSc Student & Part-time \\
Years lived in & \multirow{2}{*}{>20} & \multirow{2}{*}{>20} & \multirow{2}{*}{>20} \\
country of birth &  &  &  \\
\bottomrule
\end{tabular}
\caption{Demographic information of the three annotators.}
\label{tab:annotator_demos}
\end{table}

\paragraph{Annotation Task} We present the task description, as well as an exemplary representation of the Excel table structure that the annotators received for the task. In addition to the annotation task, the annotators received a content warning upfront that indicated that: (1) they will be confronted with AI-generated images, (2) the images might depict cultural stereotypes amplified by AI, (3) the images might contain AI-generated disturbing images.

\begin{tcolorbox}[colframe=green!50!black, colback=blue!5!white, 
                  title=Cultural Validation Task, fonttitle=\bfseries]
Thank you for participating in this translation validation task. 

Task Description: In this study, you will receive a CSV file containing images and a list of five cultures. Your task is to examine each image and select the culture from the provided list that you believe best corresponds to the image. You will make your selection using the dropdown menu in the last column of the file.
Note: Since the images are AI-generated, some may appear unusual. Additionally, some images may depict landscapes or animals rather than people or cultural artifacts. Even if you are unsure, try to select the culture you would most likely associate with the picture.
\vspace{5mm}

\begin{minipage}{0.95\linewidth}
\begin{minipage}[t]{0.28\linewidth}
\centering
\small
\vspace{10mm}
\fbox{\parbox[c][1.5cm][c]{1.5cm}{\centering \textit{Image}\\[-2pt]\footnotesize ($300\times300$)}}
\vspace{2mm}
\end{minipage}\hfill
\begin{minipage}[t]{0.67\linewidth}
\textbf{Possible cultures}\\
\footnotesize Arabic, Senegalese, Algerian, Afghan, South Sudanese\\[6pt]
\textbf{Question}\\
\footnotesize Which culture from the list would you most likely assign to the image?\\[6pt]
\textbf{Selection}\\
(dropdown)
\end{minipage}
\end{minipage}

\end{tcolorbox}

\paragraph{Annotator Agreement} 
We calculate inter-annotator agreement statistics across the three annotators shown in Table \ref{tab:annotator_agreement}. Overall, we observe a moderate agreement between all annotators with a Fleiss' $\kappa$ of 0.551 and consistent pairwise cosine similarities. We believe that this score is driven by the inherent difficulty of the annotation task rather than shortcomings of the methodology. First, in line with prior work, we resort to using country names as proxies for certain cultures. However, culture is multidimensional and does not map neatly to national boundaries. Second, we intentionally recruited annotators with diverse cultural backgrounds. While this introduces multiple perspectives, it also results in differing intuitions about the task and varying levels of familiarity with the five cultural identity options provided. Third, the set of five cultural identity options presented for each image is sampled at random (with the exception of surface and semantic culture), meaning that we do not explicitly control for the degree of similarity or dissimilarity among the options in a given instance.

\begin{table}[h]
\centering
\small
\begin{tabular}{lc}
\toprule
\textbf{Agreement Measure} & \textbf{Score} \\
\midrule
Fleiss' $\kappa$ (all annotators) & 0.551 \\
Cosine similarity (Annotator 1 vs.\ 2) & 0.586 \\
Cosine similarity (Annotator 1 vs.\ 3) & 0.560 \\
Cosine similarity (Annotator 2 vs.\ 3) & 0.625 \\
\bottomrule
\end{tabular}
\caption{Inter-annotator agreement statistics across the annotators.}
\label{tab:annotator_agreement}
\end{table}

To better assess the reliability of our labels, we compute category-specific agreement scores, that is, agreement measured separately for each label, and report the results in Table \ref{tab:cat_spec_agreement}. This analysis shows substantially higher agreement among annotators when they label surface and semantic tendencies, which constitute our primary dimensions of interest, compared to the remaining auxiliary categories that were included as distractors. Taken together, these findings indicate that the annotation quality is sufficient to support the validation of our proposed metric, especially along the surface and semantic tendencies.

\begin{table}[h]
\centering
\small
\begin{tabular}{lccc}
\toprule
   & \multicolumn{3}{l}{\textbf{Agreement of Annotators on label}} \\ 
\textbf{Label}  & A1\&A2       & A1\&A3      & A2\&A3      \\
\midrule
Sem. Tendency & 0.71                   & 0.61                  & 0.77                  \\
Sur. Tendency  & 0.69                   & 0.54                  & 0.66                  \\
Other culture     & 0.25                   & 0.38                  & 0.32                 
\end{tabular}
\caption{Category-specific agreement statistics across the annotators.}
\label{tab:cat_spec_agreement}
\end{table}

\clearpage
\section{Additional SoS score results} \label{app:add_sos_score}
\begin{figure}[h]
    \centering
    \includegraphics[width=0.7\linewidth]{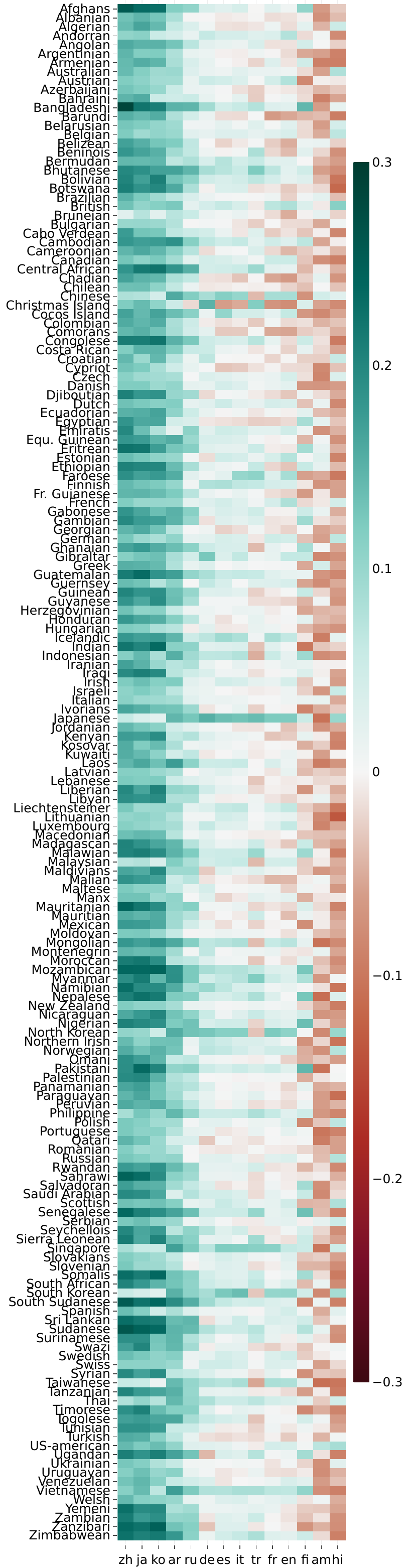}
    \caption{Heatmap of the AD SoS scores based on LAION embeddings.}
    \label{fig:sos_altdif}
\end{figure}

In this section, we present additional SoS results that are omitted from the main body for space reasons. Specifically, in Section~\ref{app:sos_laion} we report uncompressed SoS heatmaps analogous to Figure~\ref{fig:sos_model_comparison}, but extended to all models to enable a more fine-grained inspection of SoS scores for individual cultural identities. Overall, we find that AD (Figure~\ref{fig:sos_altdif}) exhibits the strongest semantic-level tendencies, followed by K3 (Figure~\ref{fig:sos_k3}), while the remaining models show predominantly surface-level tendencies. Nevertheless, we consistently observe for certain cultural identities, including Japanese, Singaporean, and Chinese, stronger semantic-level biases across models. 

Then, in Sections~\ref{app:sos_siglip} and~\ref{app:sos_dino}, we present compressed SoS heatmaps computed using two additional image embedding models, SIGLIP and DINO. Notably, DINO is not trained with textual supervision and is therefore text-independent. Despite differences in the absolute SoS value ranges, both embedding models exhibit patterns consistent with those observed in the main analysis, demonstrating the robustness of our findings across different embedding models.

Finally in Section \ref{app:sos_gender}, we provide an additional analysis of differences in SoS tendencies across the three person terms (male, female, and neutral) for each language. Here, we find differences for some of the input languages, motivating future research in the direction of intersectional biases within T2I image generations.

\subsection{SoS score results across models computed with LAION}\label{app:sos_laion}

We present the full heatmaps of the SoS results across cultures for each of the models.

\begin{figure}[h]
    \centering
    \includegraphics[width=0.7\linewidth]{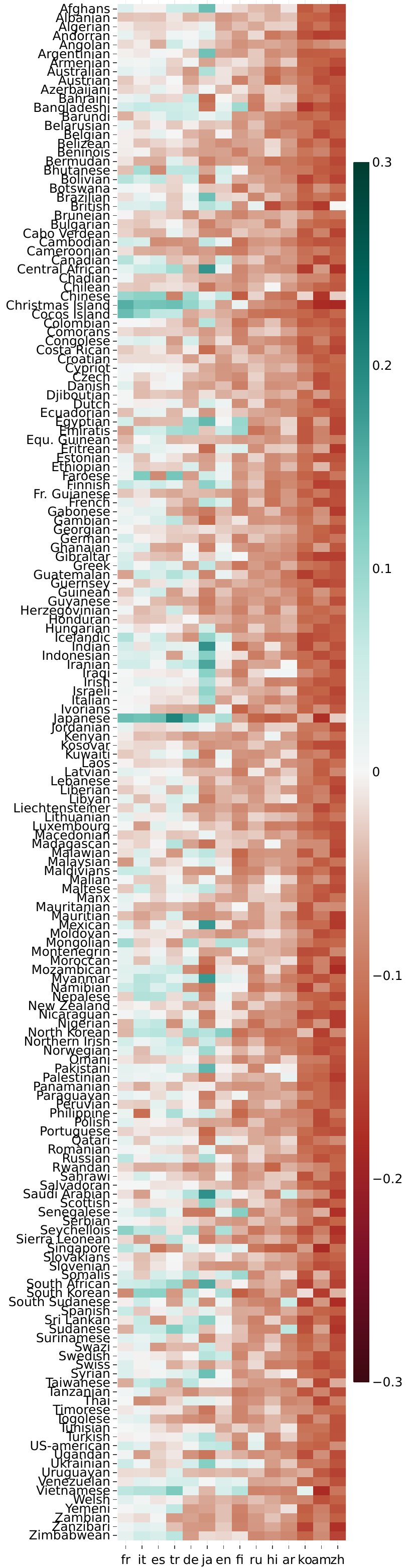}
    \caption{Heatmap of the K21 SoS scores based on LAION embeddings.}
    \label{fig:sos_k21}
\end{figure}

\begin{figure}[h]
    \centering
    \includegraphics[width=0.7\linewidth]{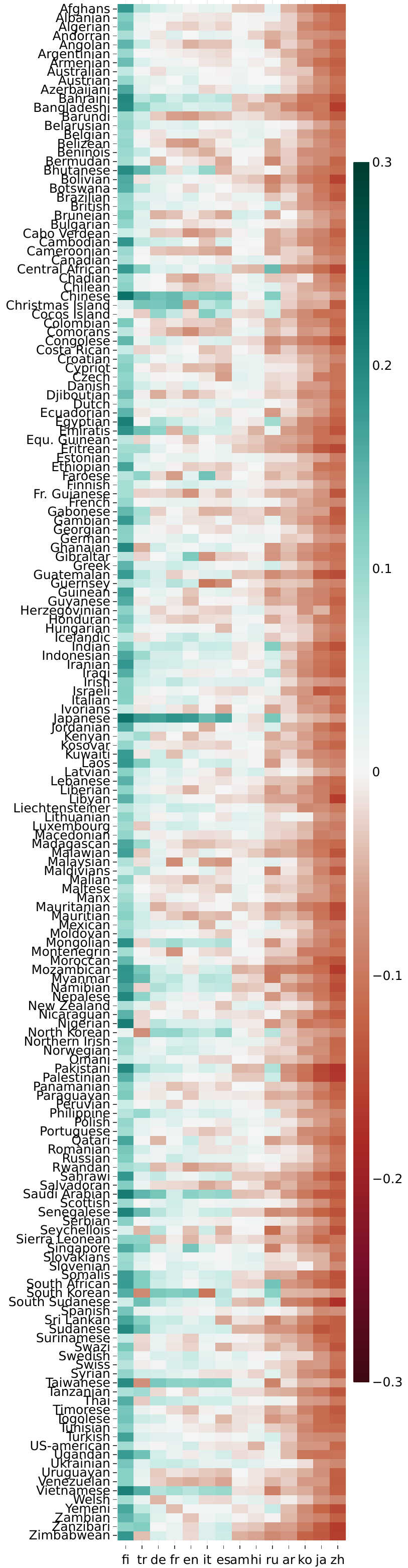}
    \caption{Heatmap of the K3 SoS scores based on LAION embeddings.}
    \label{fig:sos_k3}
\end{figure}

\begin{figure}[h]
    \centering
    \includegraphics[width=0.7\linewidth]{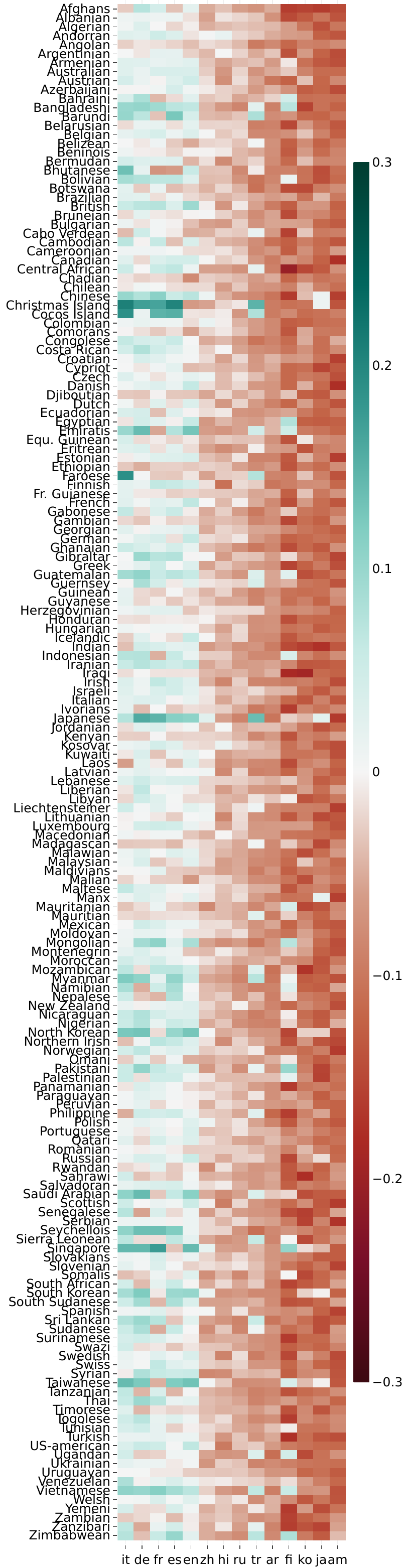}
    \caption{Heatmap of the SD21 SoS scores based on LAION embeddings.}
    \label{fig:sos_sd21}
\end{figure}

\begin{figure}[h]
    \centering
    \includegraphics[width=0.7\linewidth]{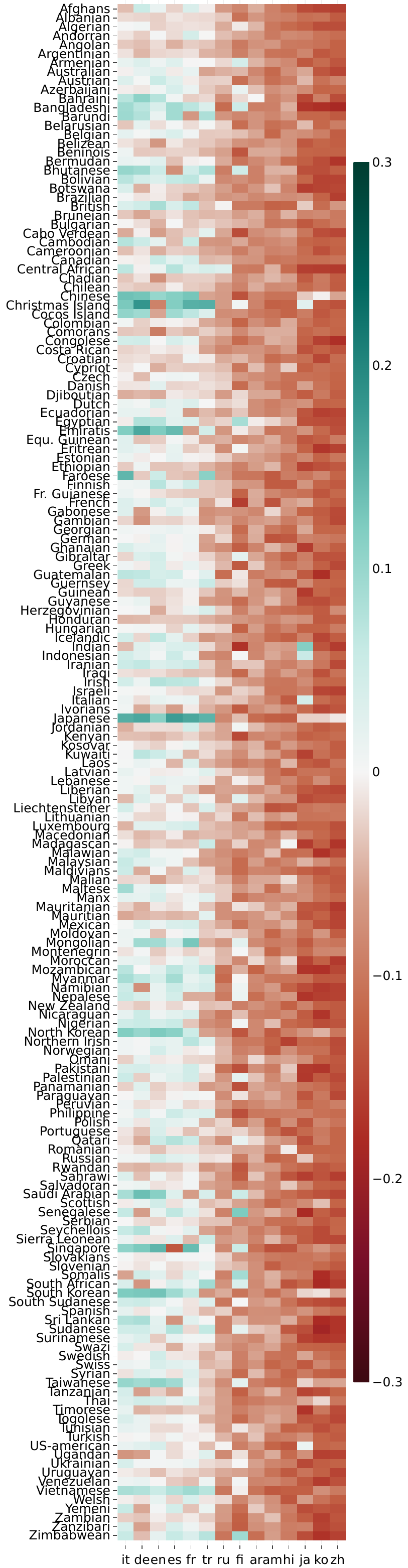}
    \caption{Heatmap of the SDXL SoS scores based on LAION embeddings.}
    \label{fig:sos_sdxl}
\end{figure}

\begin{figure}[h]
    \centering
    \includegraphics[width=0.7\linewidth]{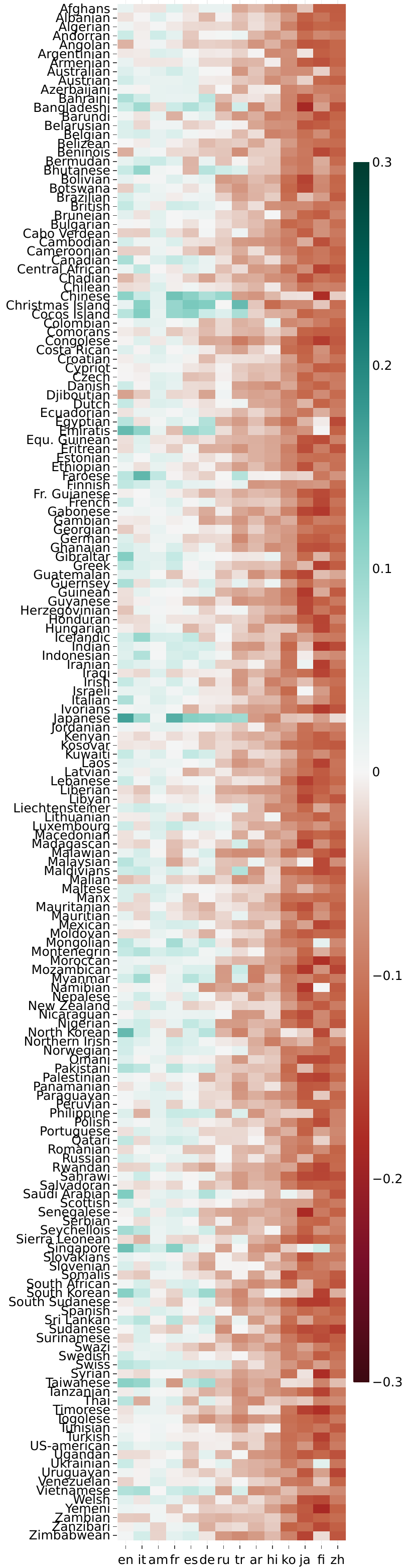}
    \caption{Heatmap of the FX SoS scores based on LAION embeddings.}
    \label{fig:sos_flux}
\end{figure}

\begin{figure}[h]
    \centering
    \includegraphics[width=0.7\linewidth]{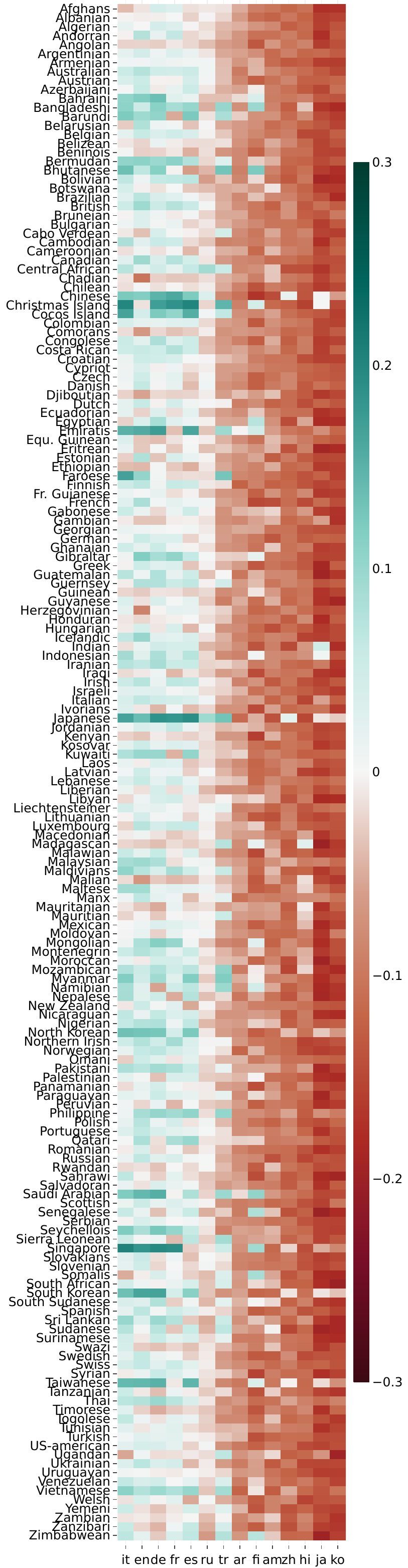}
    \caption{Heatmap of the SD3 SoS scores based on LAION embeddings.}
    \label{fig:sos_sd3}
\end{figure}

\clearpage

\subsection{SoS score validation results with SIGLIP} \label{app:sos_siglip}

Using the image embedding model SIGLIP \textsc{siglip-so400m-patch14-384} to embed the images, we obtain smaller SoS score values but see the same patterns per language and model. This proves the robustness of our findings.

\begin{figure}[h]
    \centering
    \includegraphics[width=0.5\linewidth]{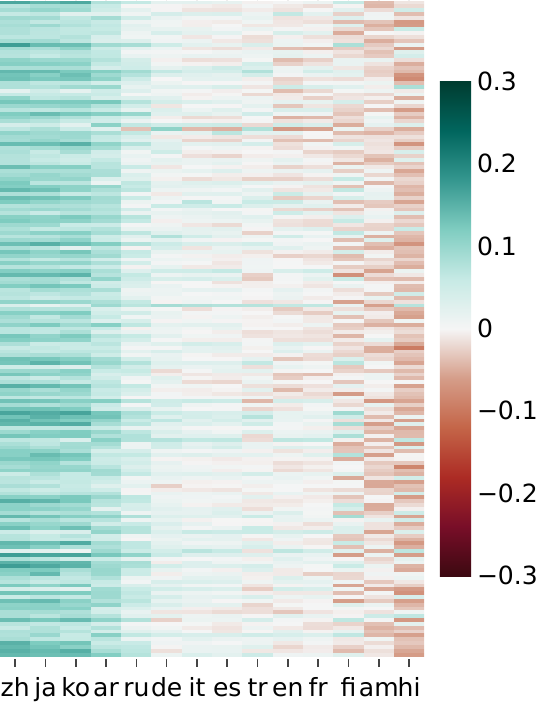}
    \caption{Heatmap of the AD SoS scores based on SIGLIP embeddings.}
\end{figure}

\begin{figure}[h]
    \centering
    \includegraphics[width=0.5\linewidth]{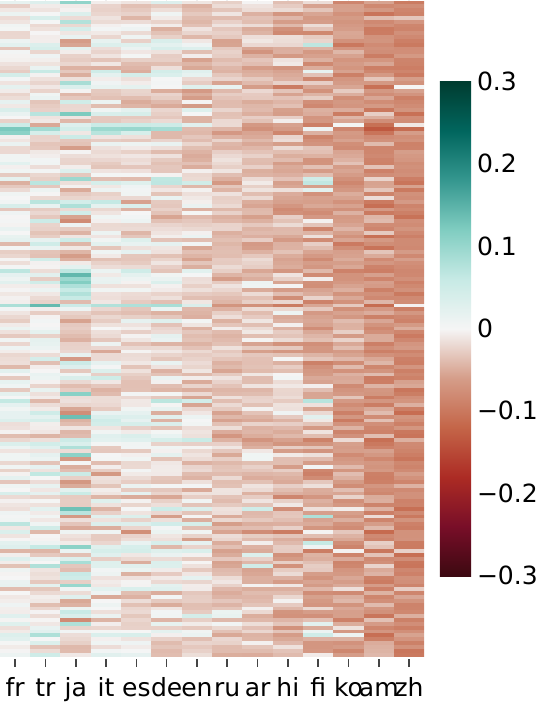}
    \caption{Heatmap of the K21 SoS scores based on SIGLIP embeddings.}
\end{figure}

\begin{figure}[h]
    \centering
    \includegraphics[width=0.5\linewidth]{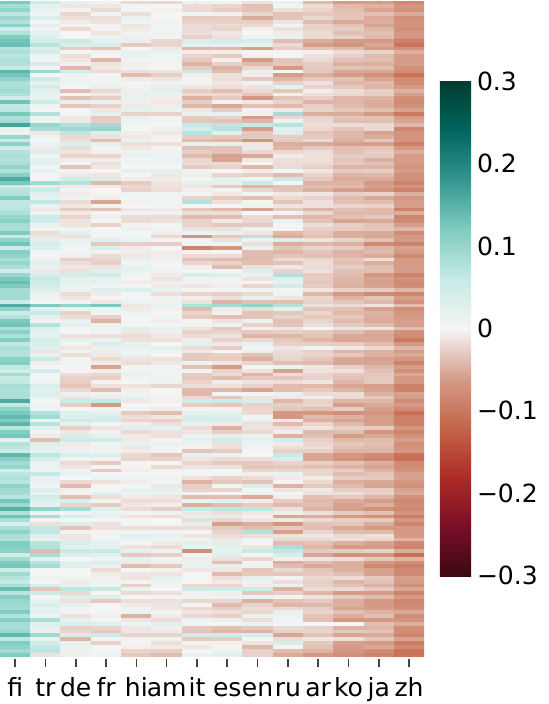}
    \caption{Heatmap of the K3 SoS scores based on SIGLIP embeddings.}
\end{figure}

\begin{figure}[h]
    \centering
    \includegraphics[width=0.5\linewidth]{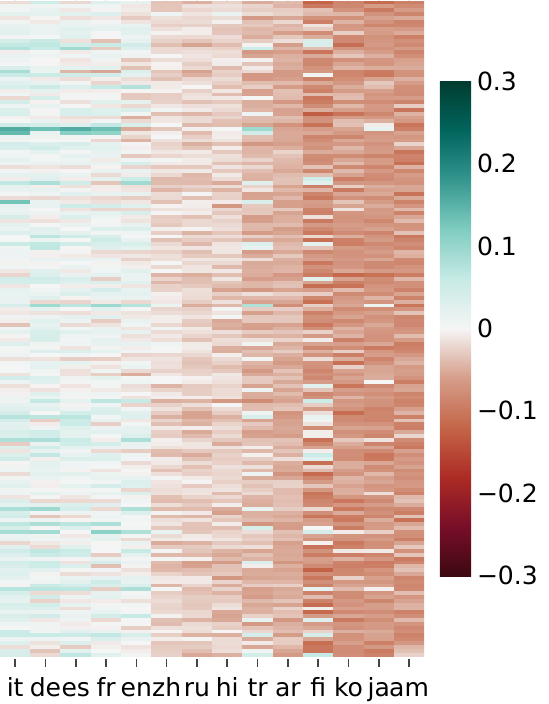}
    \caption{Heatmap of the SD21 SoS scores based on SIGLIP embeddings.}
\end{figure}

\begin{figure}[h]
    \centering
    \includegraphics[width=0.5\linewidth]{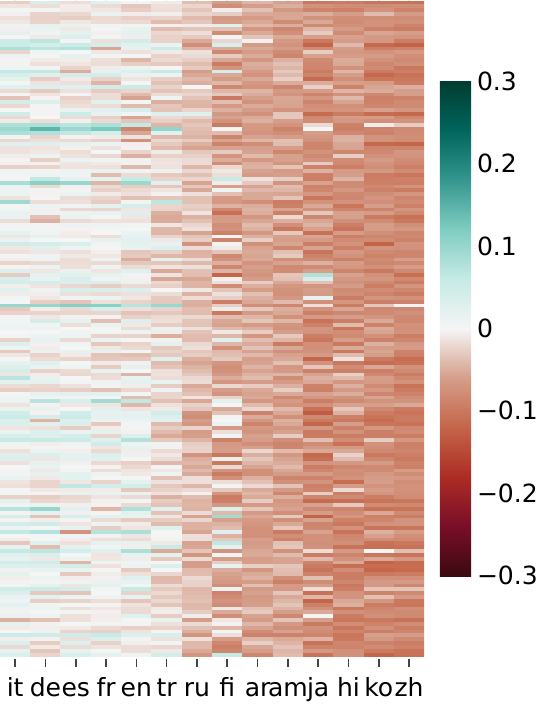}
    \caption{Heatmap of the SXL SoS scores based on SIGLIP embeddings.}
\end{figure}

\begin{figure}[h]
    \centering
    \includegraphics[width=0.5\linewidth]{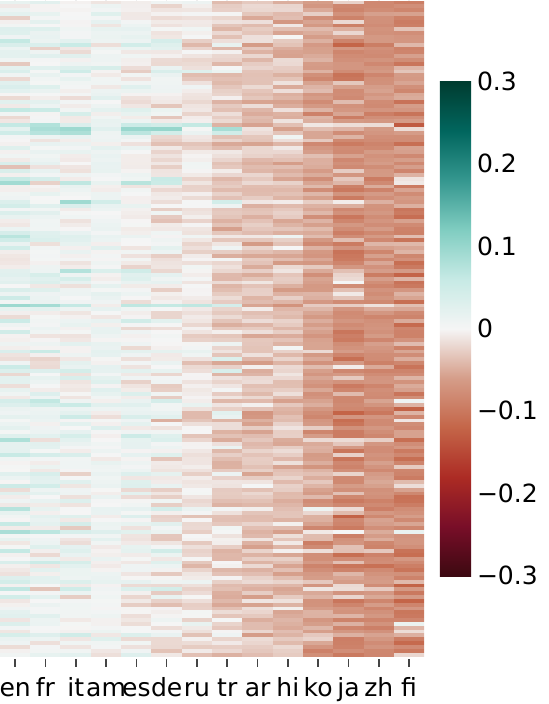}
    \caption{Heatmap of the FX SoS scores based on SIGLIP embeddings.}
\end{figure}

\begin{figure}[h]
    \centering
    \includegraphics[width=0.5\linewidth]{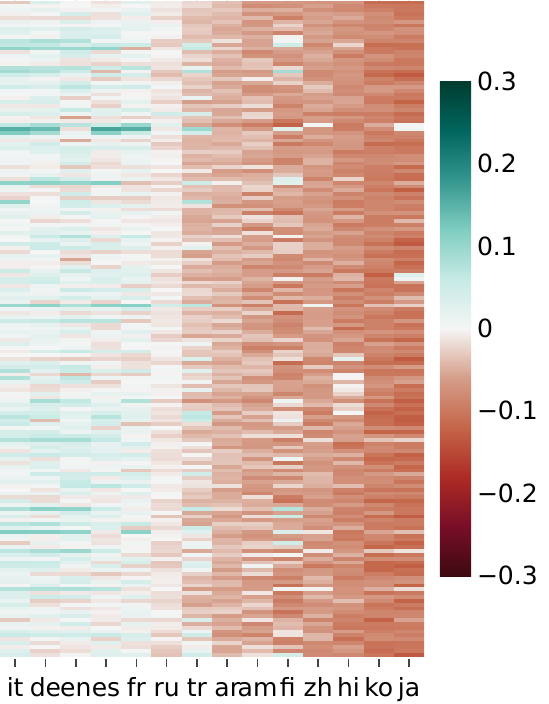}
    \caption{Heatmap of the SD3 SoS score based on SIGLIP embeddings.}
\end{figure}

{\color{black}  
\clearpage
\newpage

\subsection{SoS score validation results with DINO} \label{app:sos_dino}

Even though we are aiming to eliminate language biases by purely relying on the visual encodings of the CLIP model, without using any textual input to compare against, the CLIP model is inherently biased due to its training on a combination of image and text with predominantly English textual inputs. To account for this, we validate our results using the DINO image encoder \cite{oquab2024dinov2learningrobustvisual}, which was trained without textual supervision. We present all results created using the \texttt{facebook/dinov2-with-registers-giant} model below:
}

\begin{figure}[h]
    \centering
    \includegraphics[width=0.5\linewidth]{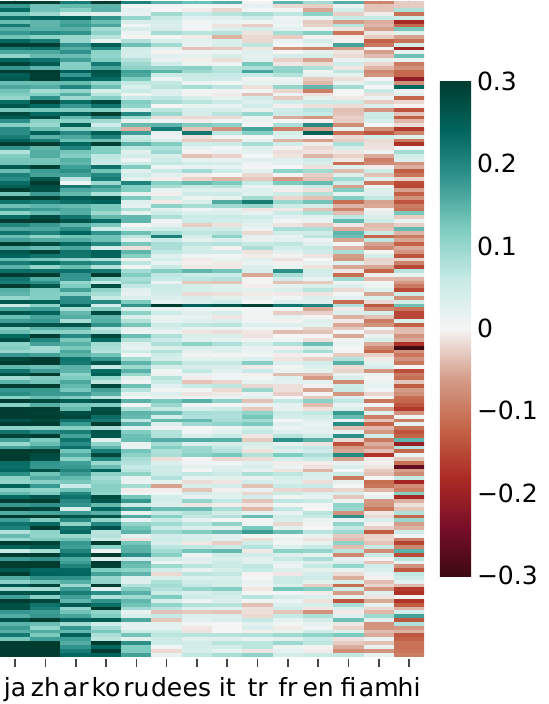}
    \caption{Heatmap of the AD SoS scores based on DINO embeddings.}
    \label{fig:dino-am9}
\end{figure}

\begin{figure}[h]
    \centering
    \includegraphics[width=0.5\linewidth]{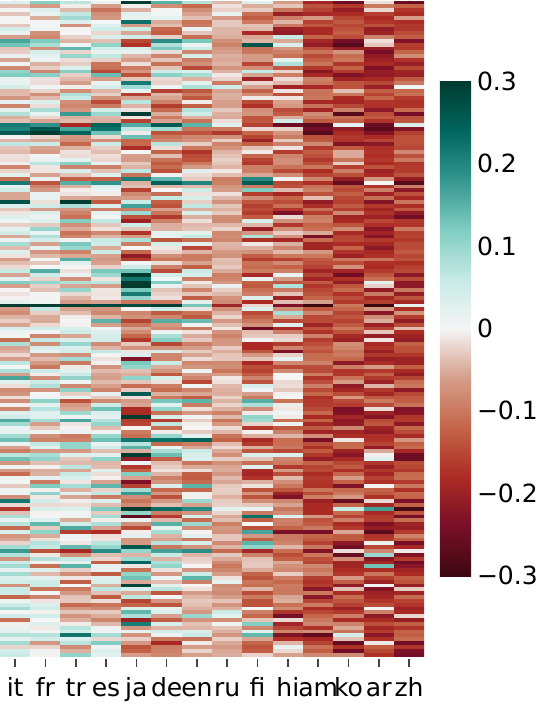}
    \caption{Heatmap of the K21 SoS scores based on DINO embeddings.}
    \label{fig:dino-k21}
\end{figure}

\begin{figure}[h]
    \centering
    \includegraphics[width=0.5\linewidth]{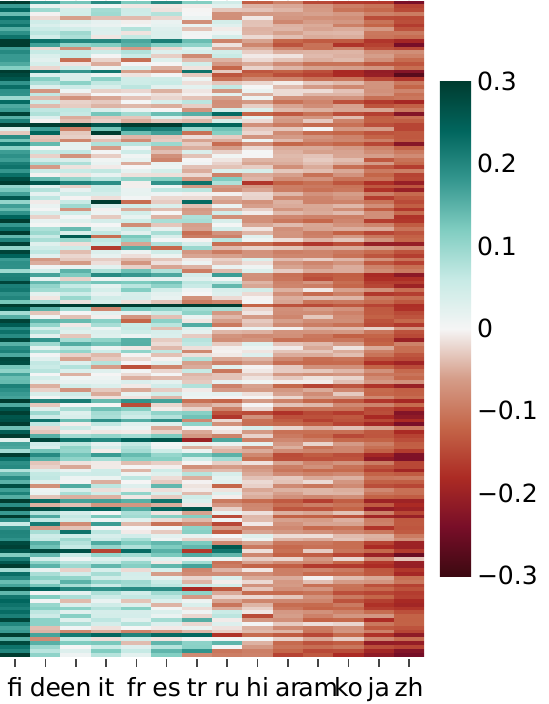}
    \caption{Heatmap of the K3 SoS scores based on DINO embeddings.}
    \label{fig:dino-k3}
\end{figure}

\begin{figure}[h]
    \centering
    \includegraphics[width=0.5\linewidth]{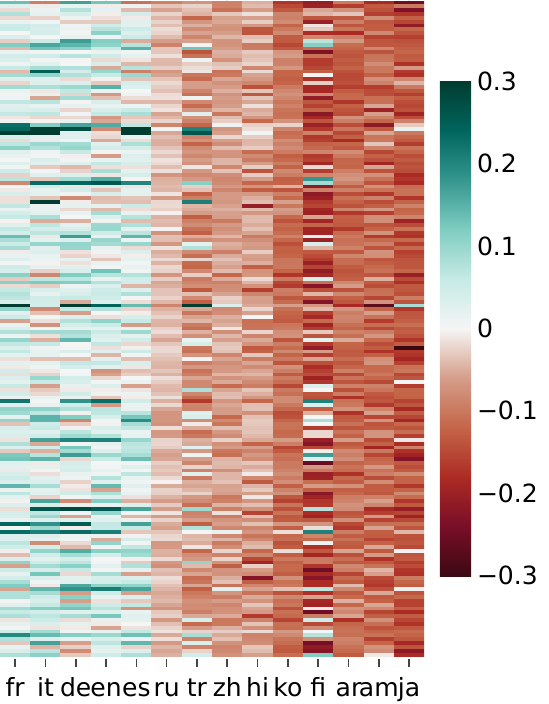}
    \caption{Heatmap of the SD21 SoS scores based on DINO embeddings.}
    \label{fig:dino-sd21}
\end{figure}

\begin{figure}[h]
    \centering
    \includegraphics[width=0.5\linewidth]{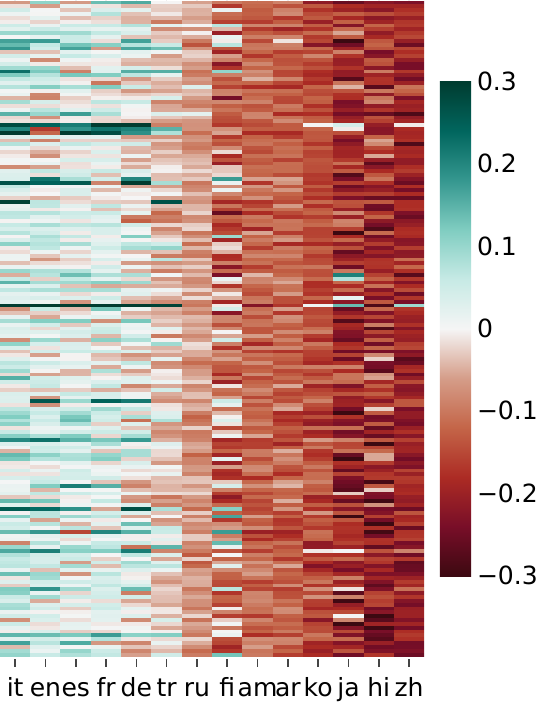}
    \caption{Heatmap of the SDXL SoS scores based on DINO embeddings.}
    \label{fig:dino-sdxl}
\end{figure}

\begin{figure}[h]
    \centering
    \includegraphics[width=0.5\linewidth]{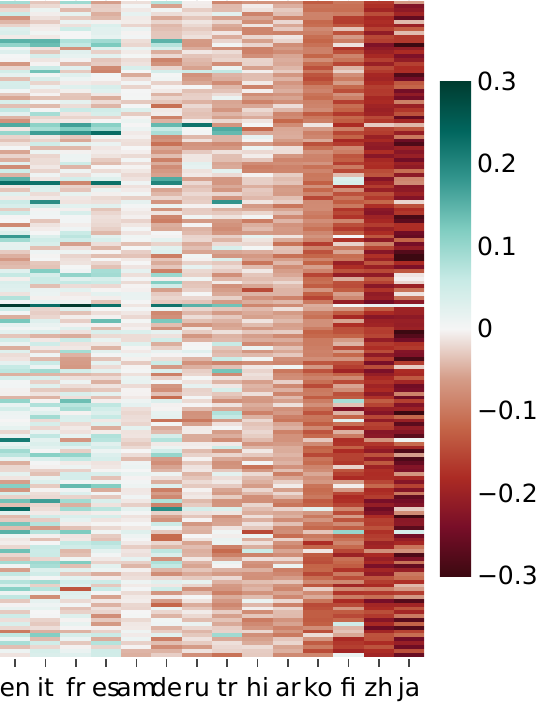}
    \caption{Heatmap of the FX SoS scores based on DINO embeddings.}
    \label{fig:dino-fx}
\end{figure}

\begin{figure}[h]
    \centering
    \includegraphics[width=0.5\linewidth]{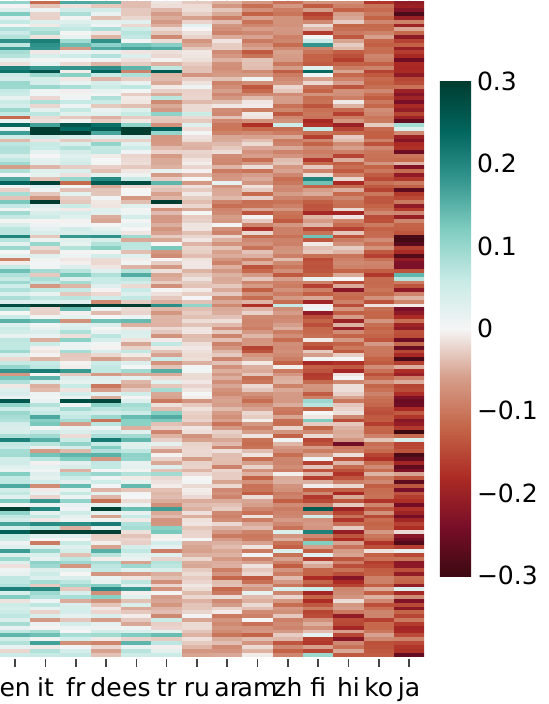}
    \caption{Heatmap of the SD3 SoS scores based on DINO embeddings.}
    \label{fig:dino-sd3}
\end{figure}

\clearpage

\subsection{Analysis of SoS score differences per person term} \label{app:sos_gender}
\begin{figure}[h]
    \centering
    \includegraphics[width=1\linewidth]{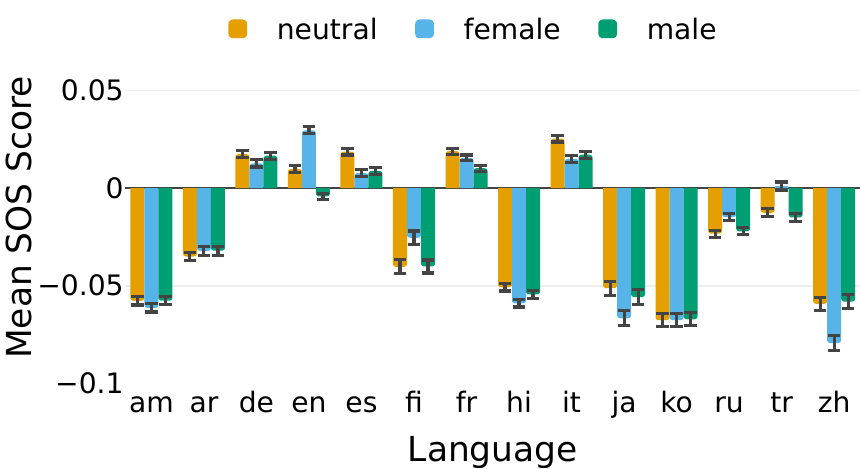}
    \caption{SoS score differences across genders. We present the SoS score mean along with the confidence interval for each gender and language, averaged over all models.}
    \label{fig:sos_gender_comparison}
\end{figure}

\textbf{Gender-related biases in T2I models vary across languages, with distinct regional patterns affecting SoS tendencies.}
To analyze the intersectional bias between gender and culture and find out whether the SoS tendency is more salient for a certain gender, we aggregate the SoS score per gender across all models. Figure \ref{fig:sos_gender_comparison} shows the mean SoS score per language and gender with confidence intervals. For some cultures, such as Amharic, Arabic, and Korean, we observe no significant differences in SoS scores across gender identities. However, in other cultures, notable patterns emerge. Our aggregation shows that for Chinese, Japanese, and Hindi (i.e., Asian languages), images associated with female identities exhibit a significantly more negative SoS score compared to male and neutral identities, indicating a stronger reliance on surface-level linguistic features. Conversely, for Russian, Turkish, and Finnish (i.e., Eastern European languages), we observe the opposite effect: female identities are less biased toward the surface form of the prompt compared to their male and neutral counterparts. These findings suggest that gender-related biases in T2I models are not uniform across languages but instead follow distinct regional patterns, reinforcing the need for a more nuanced evaluation of intersectional biases in T2I models.

\clearpage

\section{Robustness Analysis} \label{app:robustness}
In this section, we present additional experiments that further support the robustness of our findings. Specifically, we demonstrate that our main results are robust to variations in terms of variations due to prompt formulation and generalize beyond a single concept. To this end, we (1) report additional SoS results for a second concept (\textit{house}), which exhibit high correlation with the results obtained for the primary concept (\textit{person});(2) assess consistency across different prompt formulations; (3) provide per-language performance comparisons between SoS and CLIPScore, revealing substantial gains for multiple languages; and (4) report additional results using mCLIP, which, while improving over CLIPScore, does not match the performance of SoS.

\subsection{Concept Generalization}

\begin{figure*}[h]
    \centering
    \includegraphics[width=0.8\linewidth]{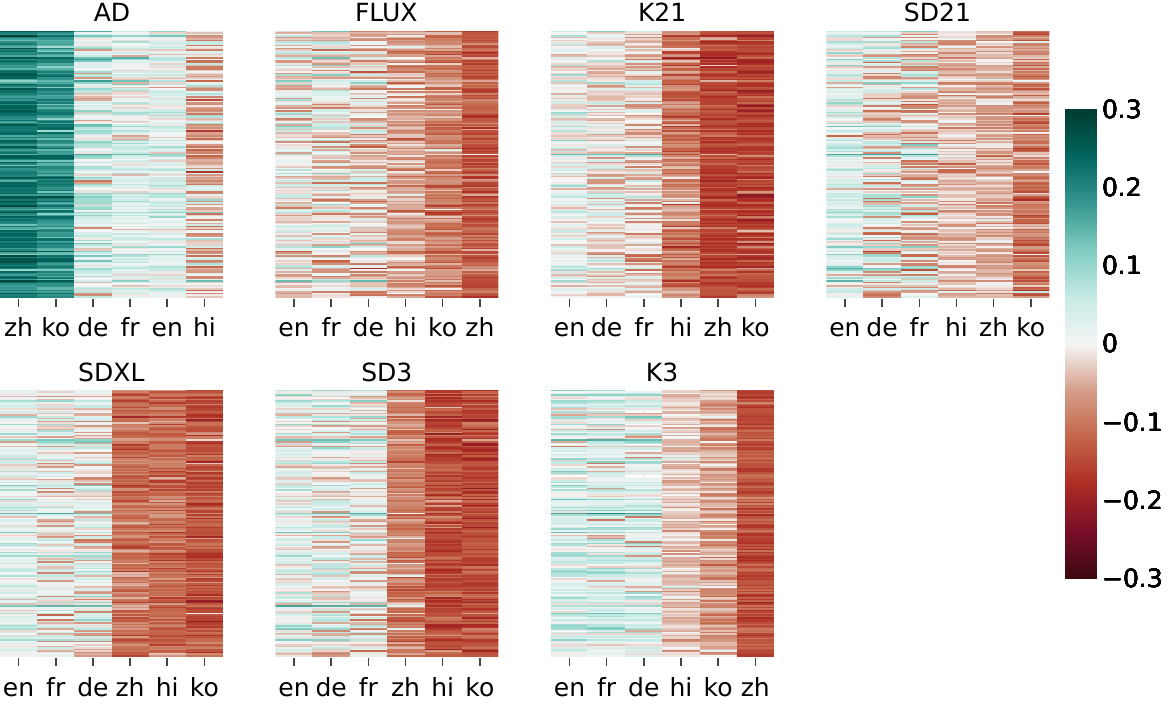}
    \caption{Heatmap of SoS-scores based on LAION embeddings for the concept ``house''.}
    \label{fig:house_heatmap}
\end{figure*}

    To demonstrate the robustness of our findings and the effectiveness of the SoS-score across different concepts, we extend our analysis to another culturally specific concept, i.e., the concept ``house''. Using a prompt template analogous to the main results -- ``A photo of a house a ${c_i}$ person lives in'' -- with three paraphrased formulations per prompt, we compute the corresponding SoS-scores across a subset of languages. For the concept house, these scores show a very strong alignment with those obtained for the concept ``person'', yielding a Pearson correlation coefficient of 87.6\%. Figure \ref{fig:house_heatmap} presents the detailed SoS-scores across cultures, languages, and models as heatmaps.

\subsection{Prompt Formulation}
To assess the robustness of SoS with respect to prompt formulation, we compute the mean absolute deviation (MAD) and Pearson correlation coefficient (PCC) between the SoS score distributions obtained from the three different prompt formulations (a, b, and c), which we report in Table~\ref{tab:prompt_formulation}. The high correlations and low deviations indicate that our SoS results are stable across these prompt variants. 

\begin{table}[h]
\centering
\small
\begin{tabular}{l|ccc}
\toprule
\textbf{Prompt Formulation} & \textbf{a \& b} & \textbf{a \& c} & \textbf{b \& c} \\ \midrule
MAD  & 0.0245 & 0.0250 & 0.0272 \\ \midrule
PCC & 0.9020 & 0.9040 & 0.8830 \\ \bottomrule
\end{tabular}
\caption{Variation across the three prompt formulations (a, b, and c). We report the mean absolute deviation (MAD) and the Pearson correlation coefficient (PCC) between the resulting SoS score distributions for images generated by each prompt formulation.}
\label{tab:prompt_formulation}
\end{table}

While our prompts are synthetically constructed and may not fully reflect real-world user inputs, this design choice is deliberate, as it minimizes variation across languages and allows observed differences to be more reliably attributed to language–culture effects rather than to prompt noise. However, given the high correlation between SoS score distributions, we do not expect substantial variation under more diverse prompt formulations, though future work could explore more naturalistic prompts drawn from real usage scenarios.

\subsection{Comparison of SoS and CLIPScore across languages}
The overall improvements on accuracy of SoS over the ClipScore are modest for non-English languages. However, our primary objective extends beyond raw accuracy to the precise identification of surface-level and semantic-level tendencies, where SoS exhibits clearer and more consistent gains. To additionally support the results presented in Figure \ref{tab:validation_results}, we provide a detailed per-language comparison of classification accuracy for both approaches in Table \ref{tab:clip_per_language_results}. In particular, SoS substantially outperforms CLIPScore for Japanese, Turkish, and Hindi, while achieving comparable performance for several other languages.

\begin{table}[]
\centering
\small
\begin{tabular}{lcc}
\toprule
\textbf{Language} & \textbf{SOS}       & \textbf{CLIPScore}      \\ \midrule
arabic & \textbf{0.78} & 0.73 \\
chinese & \textbf{0.96} & \textbf{0.96} \\
english & 0.51 & \textbf{0.86} \\
finnish & 0.75 & \textbf{0.81} \\
french  & \textbf{0.53} & 0.49  \\
german  & 0.76 & \textbf{0.80} \\
hindi   & \textbf{0.67} & 0.56  \\
italian & \textbf{0.74} & 0.70  \\
japanese & \textbf{0.85} & 0.74  \\
korean  & \textbf{0.67} & \textbf{0.67} \\
russian & \textbf{0.81} & 0.78  \\
spanish & \textbf{1.0} & \textbf{1.0}  \\
turkish & \textbf{0.96} & 0.87   \\
\bottomrule
\end{tabular}
\caption{Proportion of correctly categorized surface and semantic-level tendencies using SoS and CLIPScore.}

\label{tab:clip_per_language_results}
\end{table}

\subsection{Comparison of SoS and CLIPScore based on mCLIP}
To provide evidence that the limited performance of the CLIPScore on assessing semantic and surface-level tendencies is not due to its multilingual capabilities, we additionally report results obtained with mCLIP \cite{chen-etal-2023-mclip} in Table~\ref{tab:sos_mclip_results}. We find that while mCLIP slightly improves performance for non-English languages compared to CLIP, it does not surpass the best results achieved by SoS.

\begin{table}[]
\centering
\small
\begin{tabular}{lccc}
\toprule
 & \textbf{Acc} & $\mathbf{P_{sur}}$ & $\mathbf{P_{sem}}$ \\ \midrule
mCLIP & 77.6\% & 88.7\% & 84.8\% \\ 
SoS score & 74.0\% & 94.8\% & 84.8\% \\ \midrule
mCLIP \textbackslash en & \textbf{75.9} & 88.7 & 92.2            \\
SoS  $\setminus$  en & \textbf{79.1}\% & \textbf{94.8}\% & 94.8\%\\
\bottomrule
\end{tabular}
\caption{Comparison of mCLIP-based scores with SoS scores on the subset of human labelled instances.}
\label{tab:sos_mclip_results}
\end{table}

\clearpage
\twocolumn
\section{Linguistic Similarities}\label{app:linguistic_similarities} 

\subsection{Image Embedding Distribution} 
We examine the distribution of images created for different input languages when embedded in 2D. To this end, we perform a principal component analysis (PCA) \cite{MACKIEWICZ1993303} with a random seed of 42, reducing to 2 components. 

\textbf{European Languages Cluster together for all models. Also, multilingual models exhibit language clusters.}
Two things have already become very clear here. In contrast to all other models, AltDiffusion shows a more diverse distribution of images according to the prompted cultures and no clustering. Despite this, smaller clusters can be recognized for languages such as Amharic and Finnish, but they are not quite as clearly delineated. 
The plot of SD3, on the other hand, paints a very different picture. While the higher resource European languages such as English, Italian, Spanish, German, and French are also nicely distributed across each other in a cross-lingual space, you can see clearly separated clusters for some languages such as Arabic, Amharic, Hindi, and Turkish. Interestingly, you can even see that the languages Japanese, Korean and Chinese cluster on top of each other in a separate cluster, which already suggests a greater similarity of these images to each other in contrast to the other images. 
We also see similar patterns for the other non-explicitly multilingually trained models.

\begin{figure}[ht]
    \centering
    \includegraphics[width=0.8\linewidth]{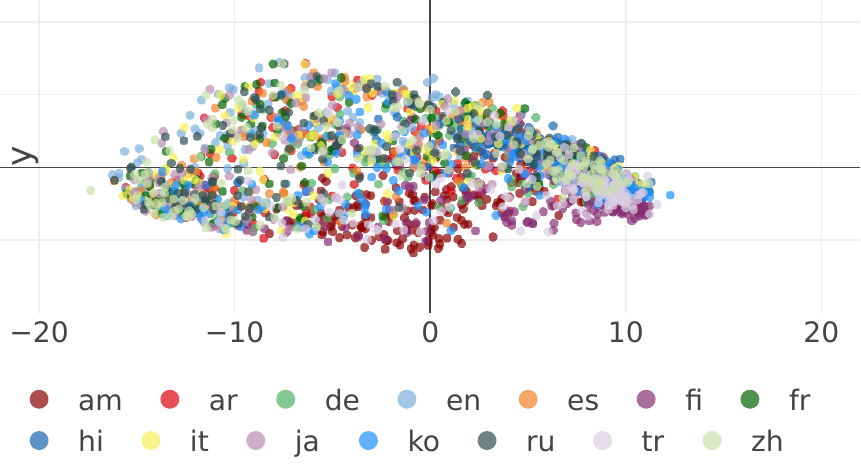}
    \caption{Distribution of image embeddings for images generated by AD.}
    \label{fig:distribution_ad9}
\end{figure}

\begin{figure}[ht]
    \centering
    \includegraphics[width=0.8\linewidth]{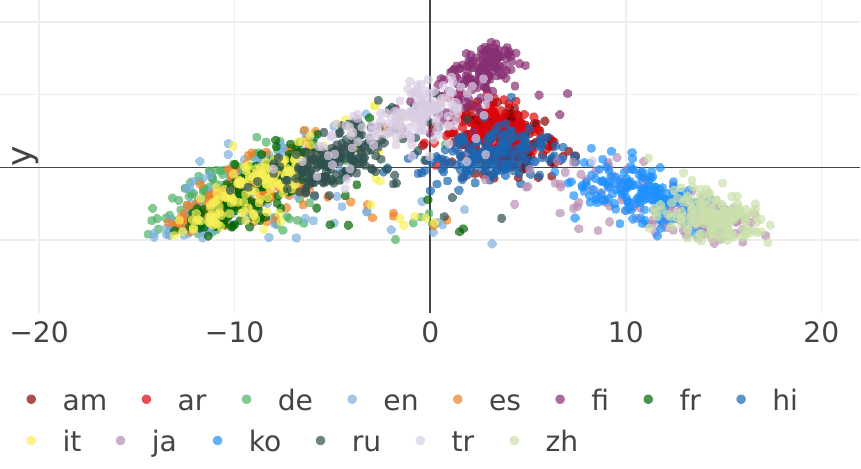}
    \caption{Distribution of image embeddings for images generated by FX.}
    \label{fig:distribution_blackforest}
\end{figure}

\begin{figure}[ht]
    \centering
    \includegraphics[width=0.8\linewidth]{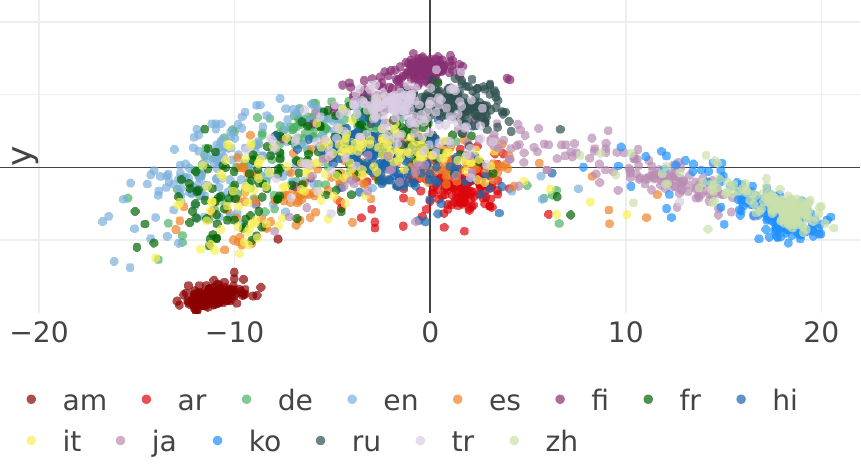}
    \caption{Distribution of image embeddings for images generated by SD21.}
    \label{fig:distribution_k21}
\end{figure}

\begin{figure}[ht]
    \centering
    \includegraphics[width=0.8\linewidth]{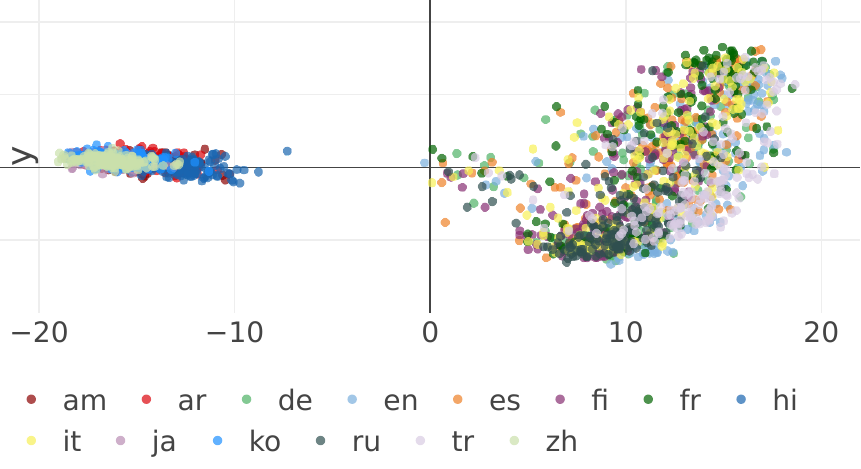}
    \caption{Distribution of image embeddings for images generated by K3.}
    \label{fig:distribution_k3}
\end{figure}

\begin{figure}[ht]
    \centering
    \includegraphics[width=0.8\linewidth]{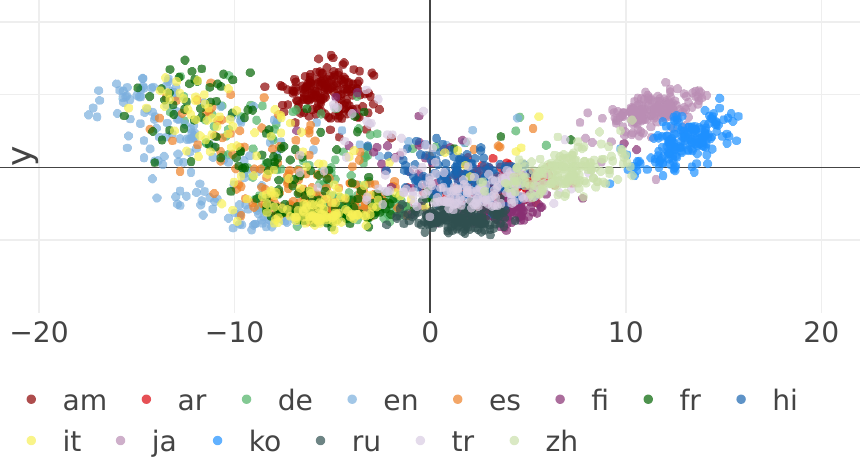}
    \caption{Distribution of image embeddings for images generated by SD21.}
    \label{fig:distribution_sd21}
\end{figure}

\begin{figure}[ht]
    \centering
    \includegraphics[width=0.8\linewidth]{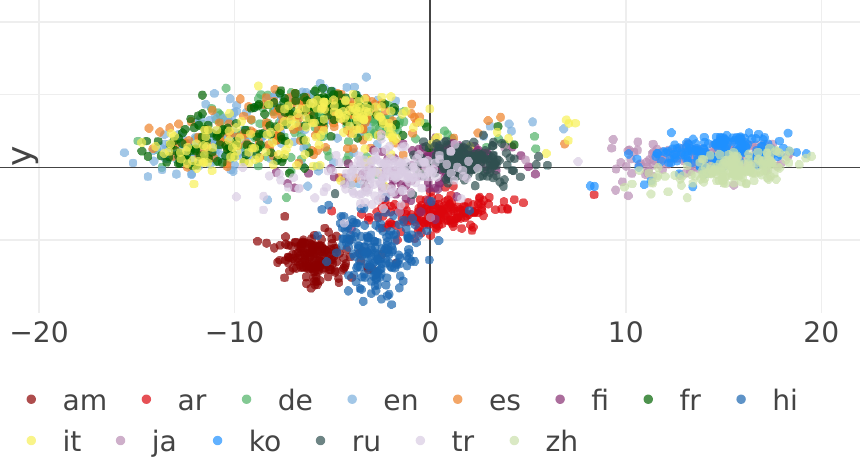}
    \caption{Distribution of image embeddings for images generated by SDXL.}
    \label{fig:distribution_sdxl}
\end{figure}

\begin{figure}[ht]
    \centering
    \includegraphics[width=0.8\linewidth]{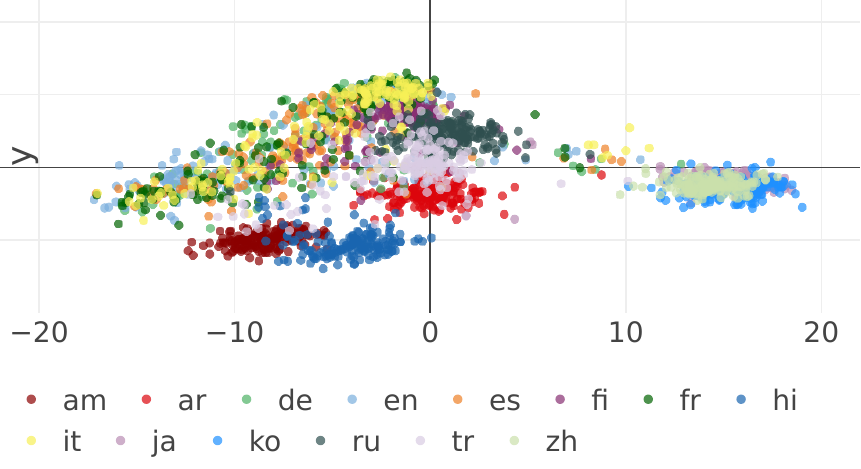}
    \caption{Distribution of image embeddings for images generated by SD3.}
    \label{fig:distribution_sd3}
\end{figure}

\clearpage

\subsection{Language Correlations} \label{app:langauge_corr}
We present the correlation analysis for all language pairs for each of the evaluated models.

\begin{figure}[ht]
    \centering
    \includegraphics[width=0.6\linewidth]{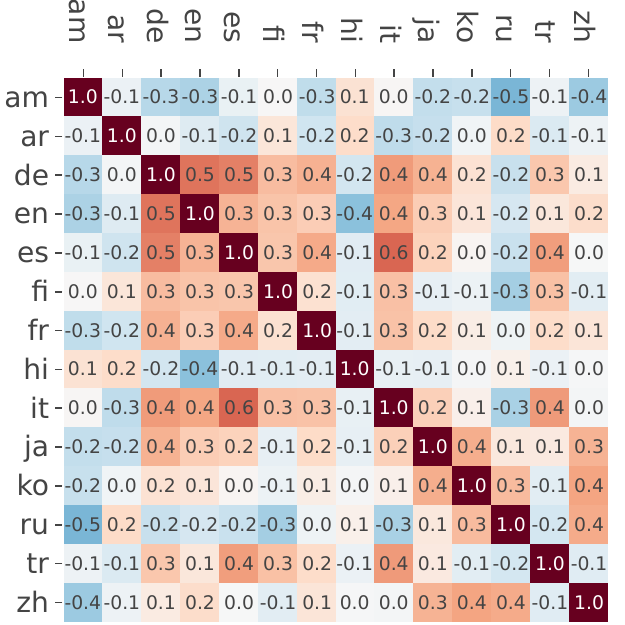}
    \caption{Per language Correlation of the images obtained by prompting SD21.}
    \label{fig:corr_sd21}
\end{figure}

\begin{figure}[ht]
    \centering
    \includegraphics[width=0.6\linewidth]{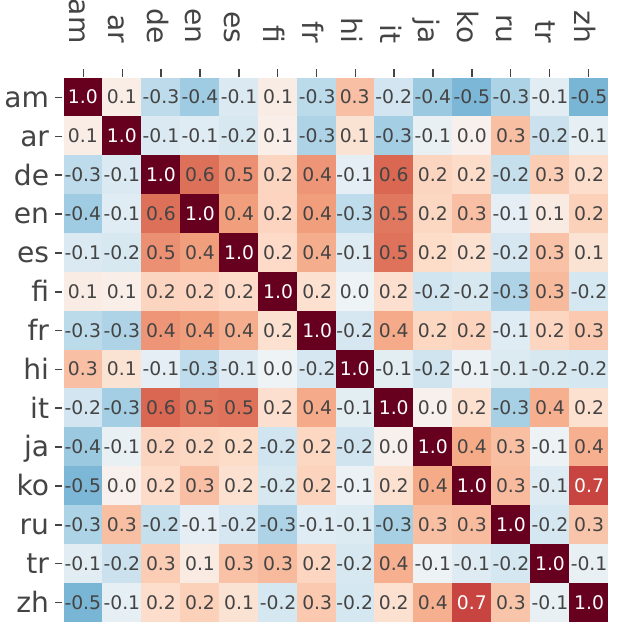}
    \caption{Per language Correlation of the images obtained by prompting SDXL.}
    \label{fig:corr_sdxl}
\end{figure}

\begin{figure}[ht]
    \centering
    \includegraphics[width=0.6\linewidth]{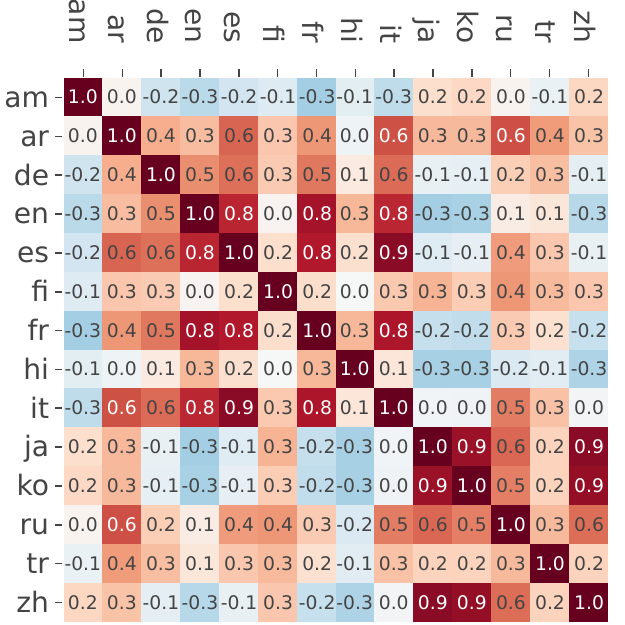}
    \caption{Per language Correlation of the images obtained by prompting AD.}
    \label{fig:corr_altdif}
\end{figure}

\begin{figure}[ht]
    \centering
    \includegraphics[width=0.6\linewidth]{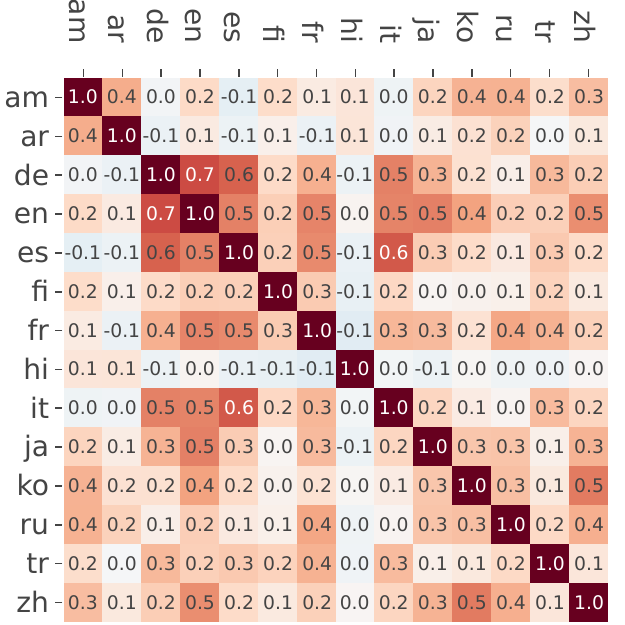}
    \caption{Per language Correlation of the images obtained by prompting FX.}
    \label{fig:corr_flux}
\end{figure}

\begin{figure}[ht]
    \centering
    \includegraphics[width=0.6\linewidth]{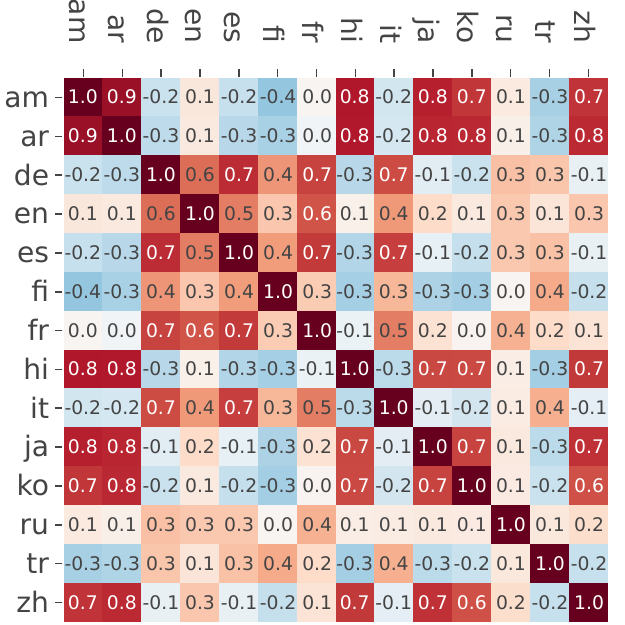}
    \caption{Per language Correlation of the images obtained by prompting K3.}
    \label{fig:corr_k3}
\end{figure}

\begin{figure}[ht]
    \centering
    \includegraphics[width=0.6\linewidth]{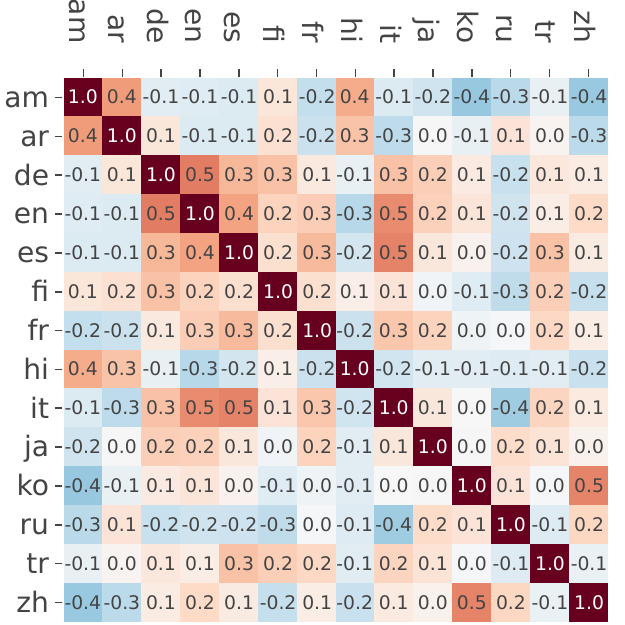}
    \caption{Per language Correlation of the images obtained by prompting K21.}
    \label{fig:corr_k21}
\end{figure}

\begin{figure}[ht]
    \centering
    \includegraphics[width=0.6\linewidth]{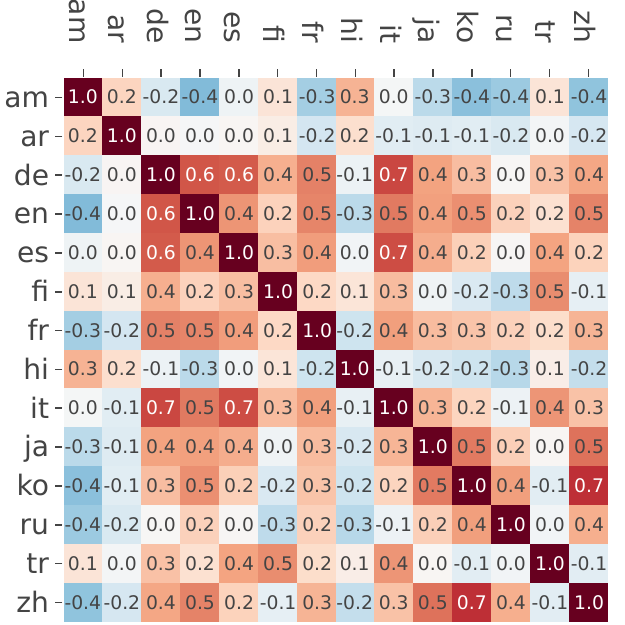}
    \caption{Per language Correlation of the images obtained by prompting SD3.}
    \label{fig:corr_sd3}
\end{figure}

\clearpage

\section{Colour Analysis} \label{app:colour_analysis}

In this section, we provide additional results on the color analysis. First, we provide the full results for the cultural identity group \textit{german} across languages and models. Second, we provide more examples across several cultural identities for images when prompting in German and images when prompting in Chinese. Finally, we provide distribution plots of the `value' in the colors HSV codes. The value represents the brightness of a color, ranging from 0 (complete darkness) to 1 (full brightness). We observe notable differences in brightness distribution across models. Specifically, images generated for European languages tend to feature darker tones compared to those produced for languages like Chinese and Hindi.

\subsection{Full results for one cultural identity}

\begin{figure}[h]
    \centering
    \includegraphics[width=0.9\linewidth]{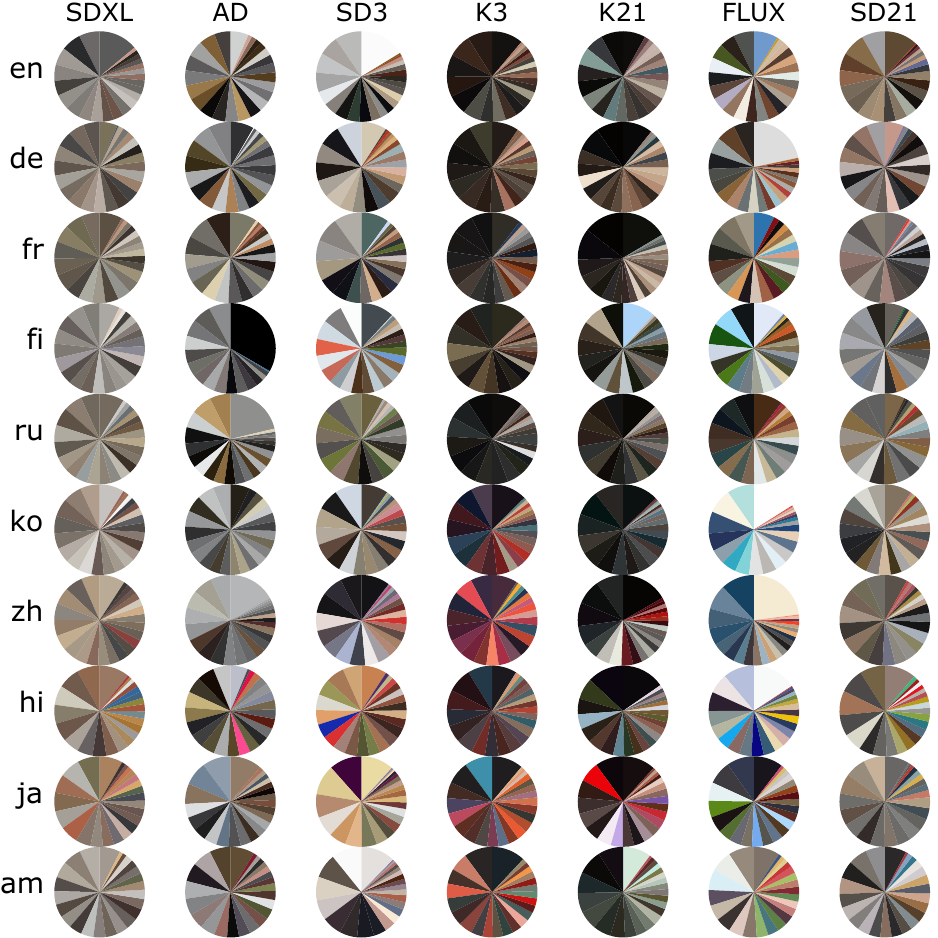}
    \caption{Most prominent colors across languages for the cultural identity \textit{German}.}
\end{figure}

\subsection{Results for multiple cultural identities for prompts in German and Chinese}
\begin{figure}[h]
    \centering
    \includegraphics[width=1\linewidth]{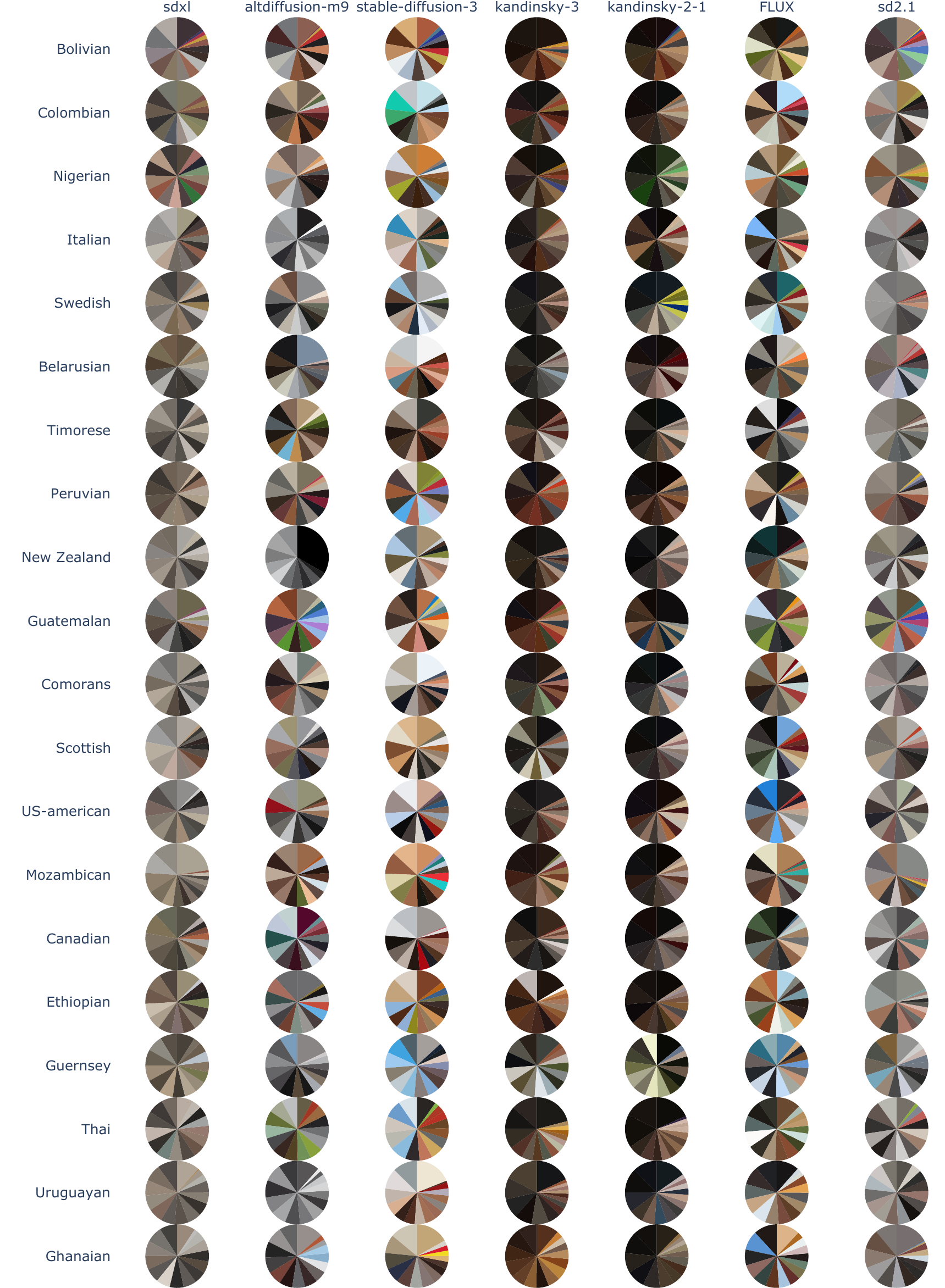}
    \caption{Prominent colors across 20 random Cultures when prompting in \textit{German}.}
    \label{fig:20_colors_german}
\end{figure}

\begin{figure}[h]
    \centering
    \includegraphics[width=1\linewidth]{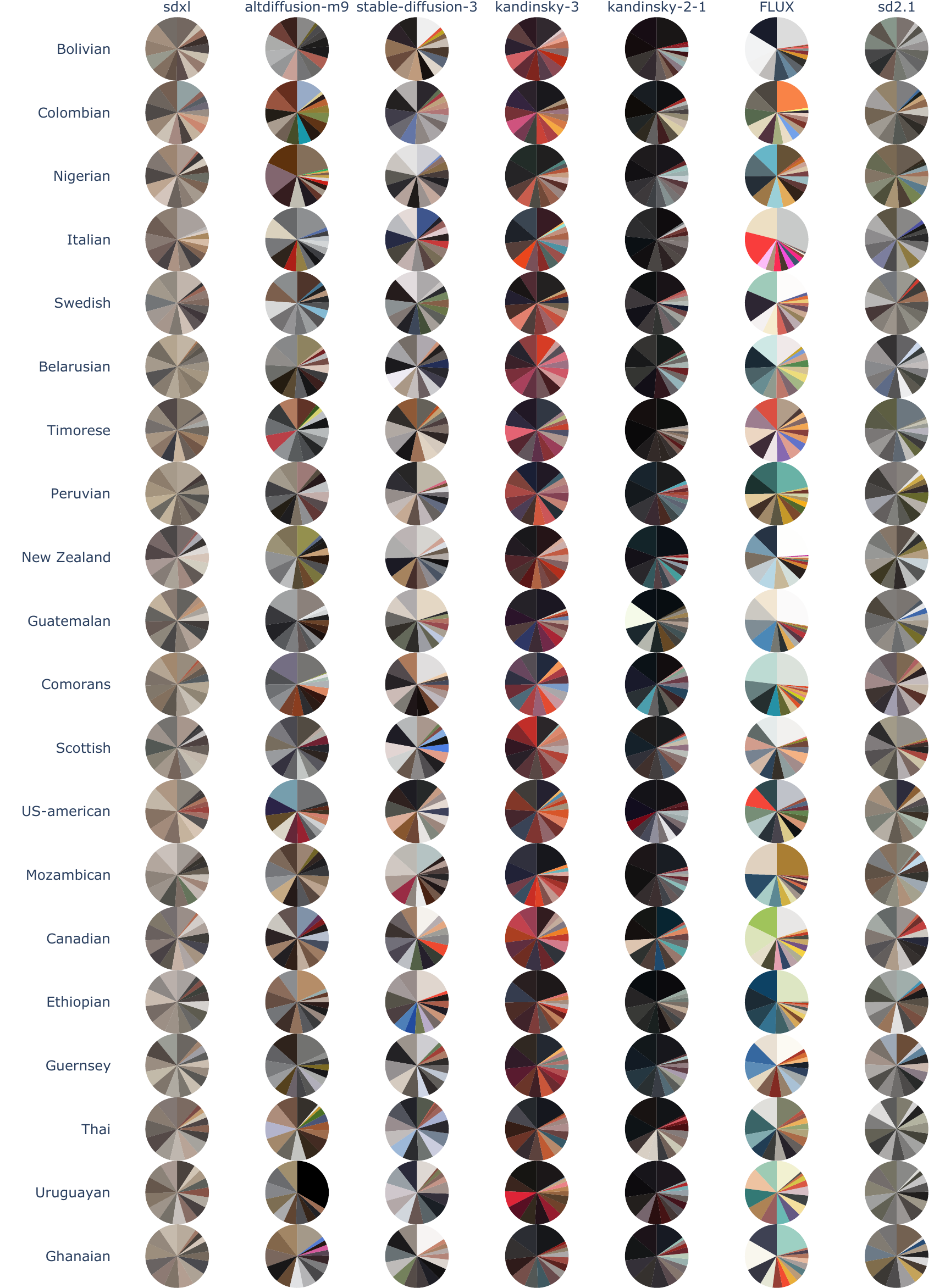}
    \caption{Prominent colors across 20 random Cultures when prompting in \textit{Chinese}.}
    \label{fig:20_colors_chinese}
\end{figure}

\clearpage
\subsection{Distribution of HSV-Values}

\begin{figure}[h]
    \centering
    \includegraphics[width=1\linewidth]{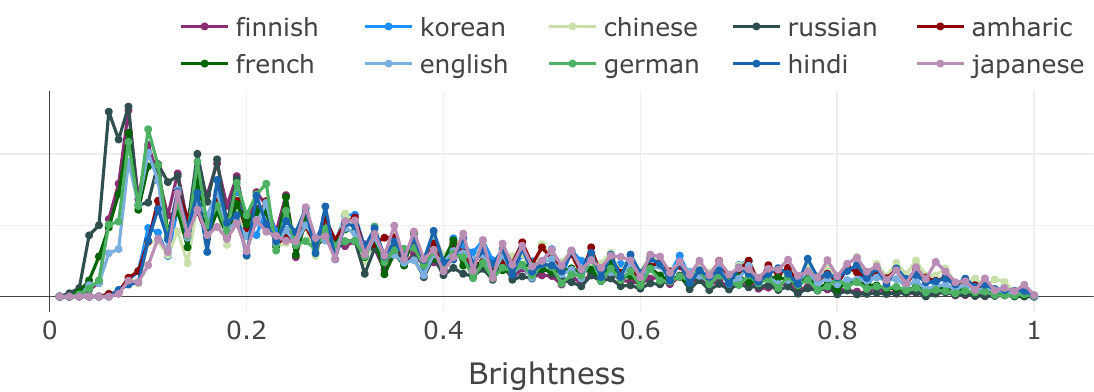}
    \caption{Value distribution of HSV values across the 8 most prominent color clusters across images for different languages obtained by prompting \textsc{K3}.}
    \label{fig:hsv_dist_k3}
\end{figure}

\begin{figure}[h]
    \centering
    \includegraphics[width=1\linewidth]{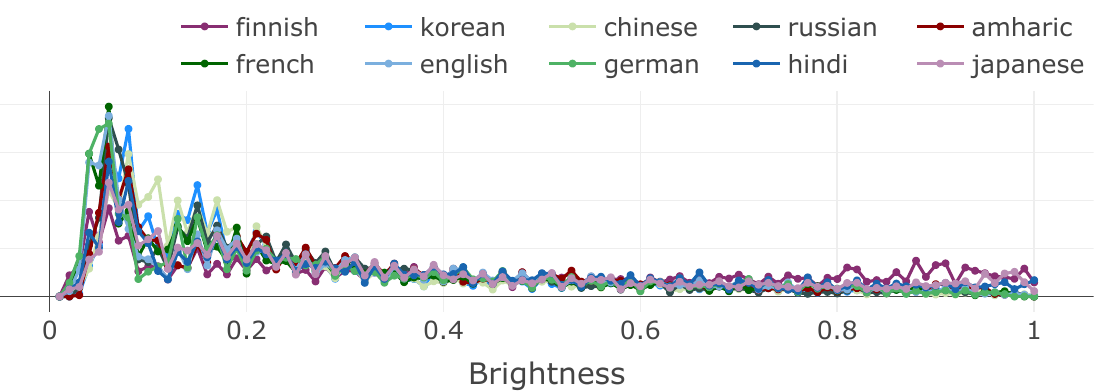}
    \caption{Value distribution of HSV values across the 8 most prominent color clusters across images for different languages obtained by prompting \textsc{K21}.}
    \label{fig:hsv_dist_altdif}
\end{figure}

\begin{figure}[h]
    \centering
    \includegraphics[width=1\linewidth]{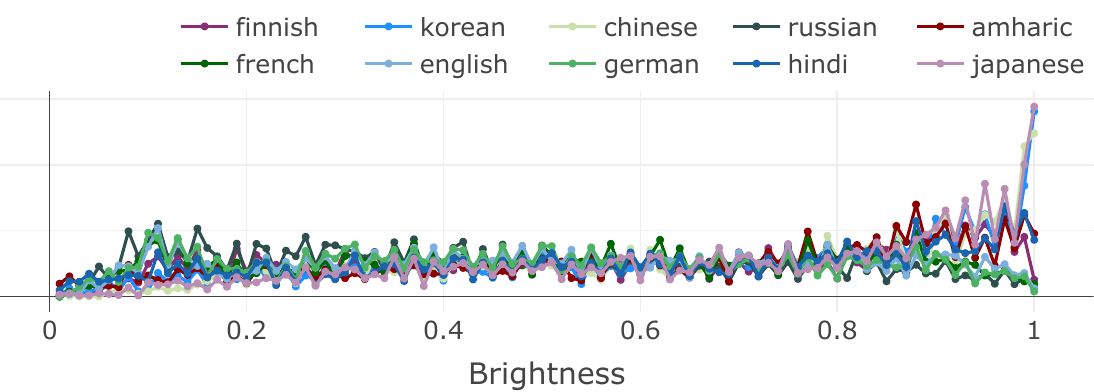}
    \caption{Value distribution of HSV values across the 8 most prominent color clusters across images for different languages obtained by prompting \textsc{FX}.}
    \label{fig:hsv_dist_flux}
\end{figure}

\clearpage

\onecolumn 

\section{VQA Analysis}\label{app:vqa_analysis}

\subsection{VQA Analysis using Fighting Words Method}

\begin{table}[h]
\setlength{\tabcolsep}{4pt} 
\renewcommand{\arraystretch}{1.1} 
    \centering
    \small
    \begin{tabular}{p{3.95cm}|p{11cm}}
    \toprule
    \textbf{Category} & \textbf{Terms} \\
    \midrule
Generic image describing terms & 
    item, product, object, hardship, abandonment, character, couple, 
   text, word, font, title, subject, depiction,  complexity,  texture, photograph, uneven, 
   person,  contours, picturesque, population, highrise, collage, room, portrait, individual, illustration,
   figure, abstract,  forehead, mood, outfit, hair, face, wall, pose, shoulder, people, shirt,
   head, wrap, group, pack, mute  \\
Quantitative and temporal terms & 
   20th, 19th, century, archival, digital, historical \\

Generic adjectives & 
   richly, tall, dynamic, dapple, official, blackandwhite, updo, minimalist, welllit, scenic, simple,  chaotic, 
   fantastical, bright, vibrant, grand, unsettling, classical, clear, long, short,  intense, densely,  rough, neutral, plain, undisturbed, gridlike, 
   casual, peaceful, tightly, contemplative, sparse, richness, heavy, overcast, dark, soft, brown,
   detailed, traditional \\

    Directional and relational terms & 
   north, south, east, underneath, outermost, central, directly, subject, title, fourth, second \\

Generic verbs & 
   elaborate, crash, wear, stand, enchanting, help, desolate, use,  neglect, highlight, stamp \\

Pronouns and person identifier &  man, mans \\
\bottomrule
    \end{tabular}
    \caption{List of filtered terms that were not validated by human annotators due to lack of visual appearance.}
    \label{tab:filtered_terms}
\end{table}

\twocolumn
\subsection{VQA Analysis Results}

We present the results of the VQA analysis for all language-model combinations that exhibited surface-level tension. We further present example images for each of them.

\begin{figure}[h]
    \centering
    \begin{subfigure}[b]{\linewidth}
         \centering
         \includegraphics[width=0.7\textwidth]{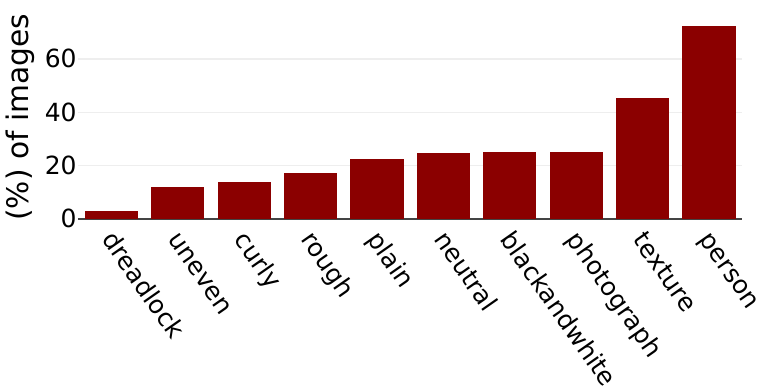}
     \end{subfigure}
     \hfill
     \begin{subfigure}[b]{\linewidth}
         \centering
         \includegraphics[width=0.7\textwidth]{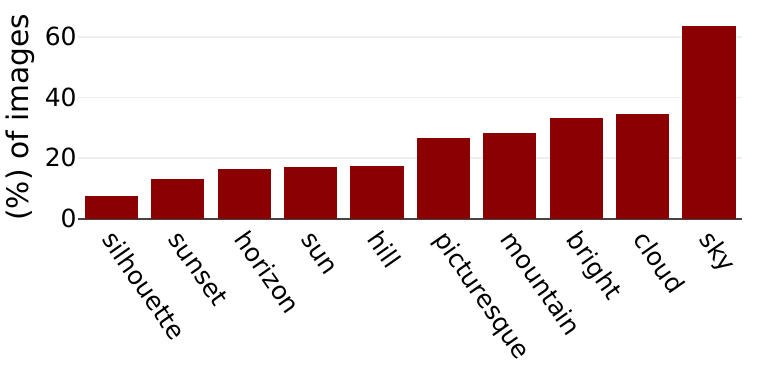}
     \end{subfigure}
     \hfill
     \begin{subfigure}[b]{\linewidth}
         \centering
         \includegraphics[width=0.7\textwidth]{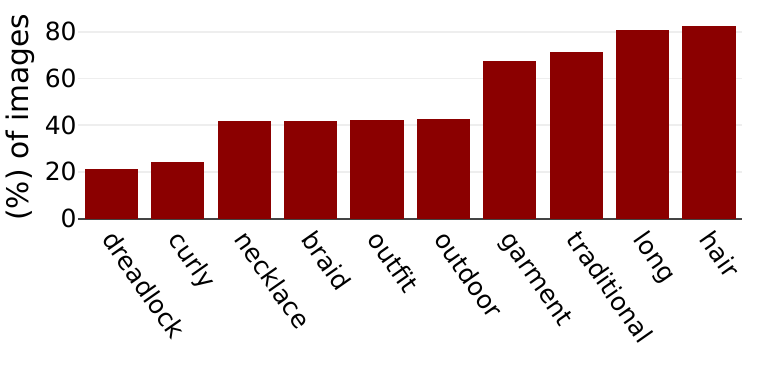}
     \end{subfigure}
     \hfill
     \begin{subfigure}[b]{\linewidth}
         \centering
         \includegraphics[width=0.7\textwidth]{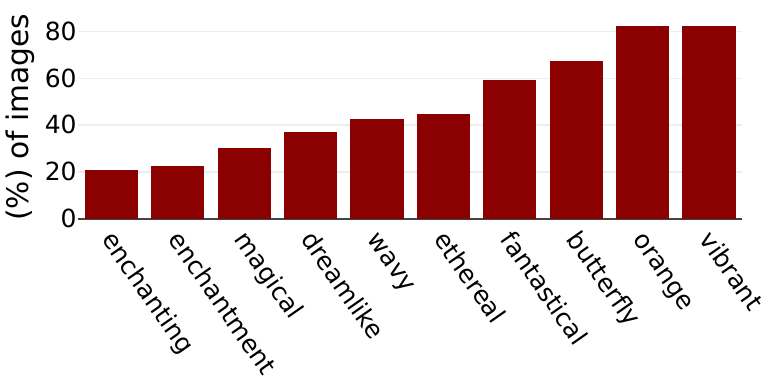}
     \end{subfigure}
     \hfill
     \begin{subfigure}[b]{\linewidth}
         \centering
         \includegraphics[width=0.7\textwidth]{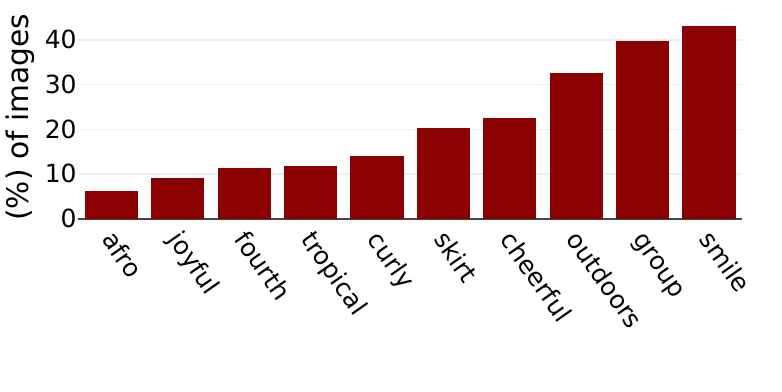}
     \end{subfigure}
     \hfill
     \begin{subfigure}[b]{\linewidth}
         \centering
         \includegraphics[width=0.7\textwidth]{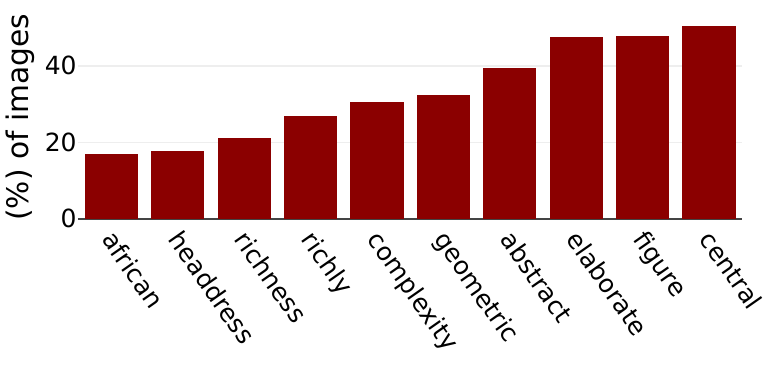}
     \end{subfigure}
     \hfill
     \begin{subfigure}[b]{\linewidth}
         \centering
         \includegraphics[width=0.7\textwidth]{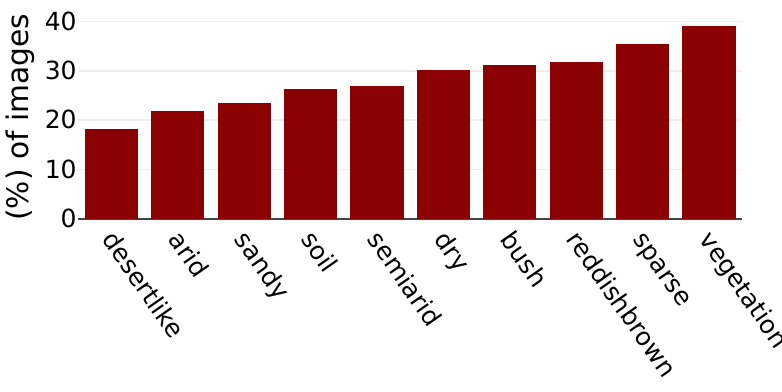}
     \end{subfigure}
    \caption{VQA results: We present the top 15 terms for images generated using input prompts in Amharic.}
    \label{fig:vqa_am}
\end{figure}

\begin{figure}[h]
    \centering
    \begin{subfigure}[b]{\linewidth}
         \centering
         \includegraphics[width=0.7\textwidth]{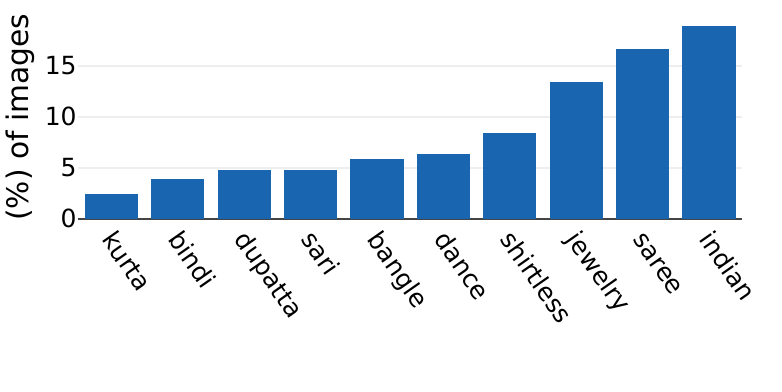}
     \end{subfigure}
     \hfill
     \begin{subfigure}[b]{\linewidth}
         \centering
         \includegraphics[width=0.7\textwidth]{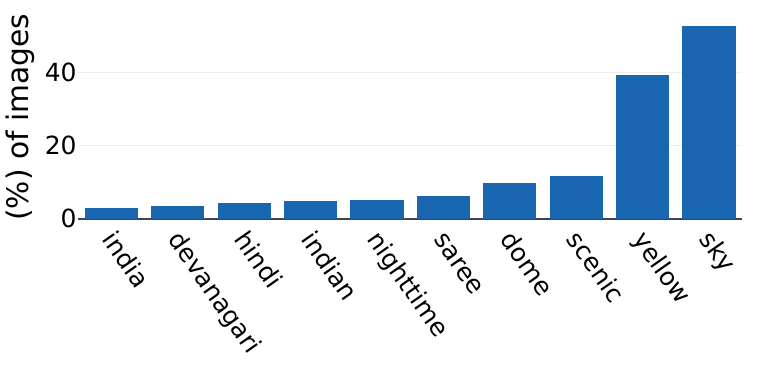}
     \end{subfigure}
     \hfill
     \begin{subfigure}[b]{\linewidth}
         \centering
         \includegraphics[width=0.7\textwidth]{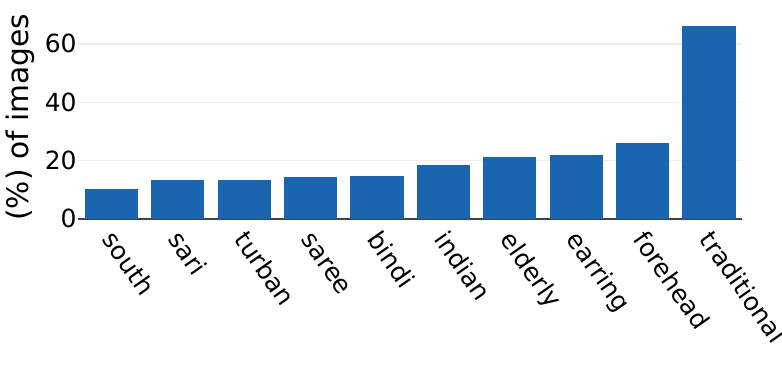}
     \end{subfigure}
     \hfill
     \begin{subfigure}[b]{\linewidth}
         \centering
         \includegraphics[width=0.7\textwidth]{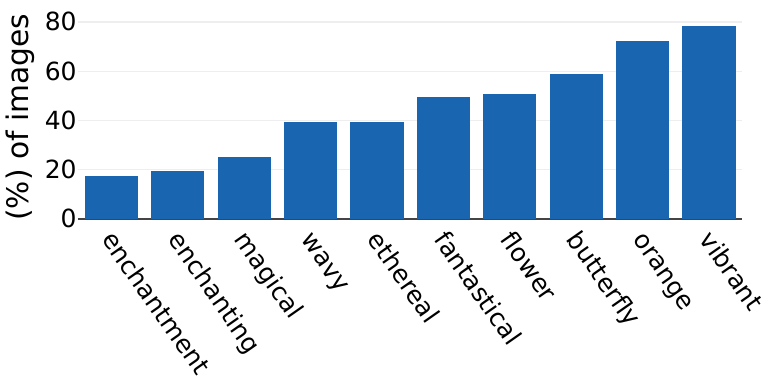}
     \end{subfigure}
     \hfill
     \begin{subfigure}[b]{\linewidth}
         \centering
         \includegraphics[width=0.7\textwidth]{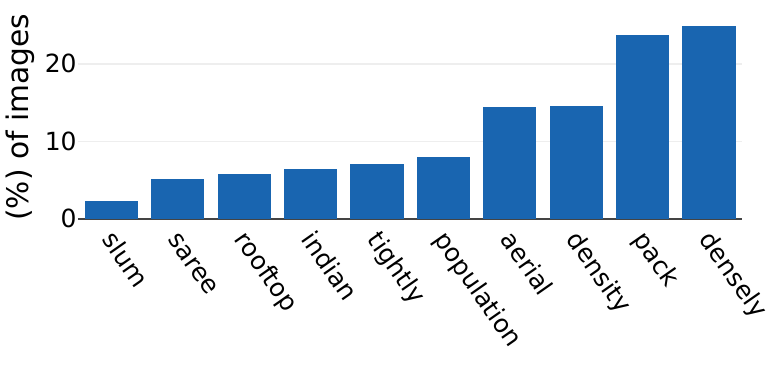}
     \end{subfigure}
     \hfill
     \begin{subfigure}[b]{\linewidth}
         \centering
         \includegraphics[width=0.7\textwidth]{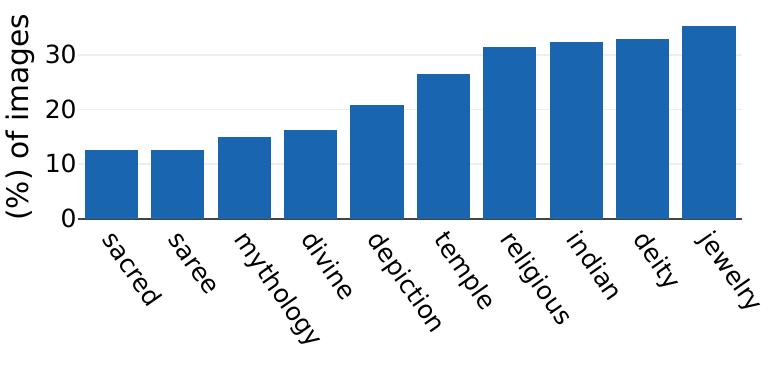}
     \end{subfigure}
     \hfill
     \begin{subfigure}[b]{\linewidth}
         \centering
         \includegraphics[width=0.7\textwidth]{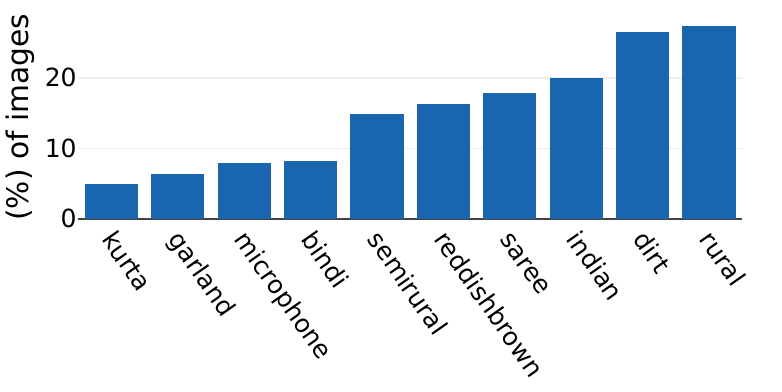}
     \end{subfigure}
    \caption{VQA results: We present the top 15 terms for images generated using input prompts in Hindi.}
    \label{fig:vqa_hi}
\end{figure}

\begin{figure}[h]
    \centering
    \begin{subfigure}[b]{\linewidth}
         \centering
         \includegraphics[width=0.7\textwidth]{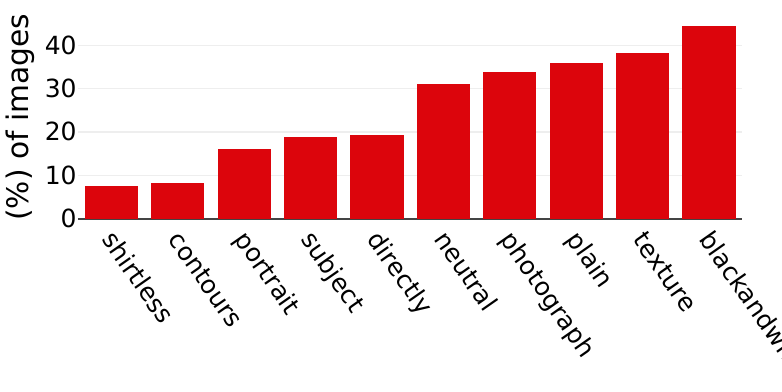}
     \end{subfigure}
     \hfill
     \begin{subfigure}[b]{\linewidth}
         \centering
         \includegraphics[width=0.7\textwidth]{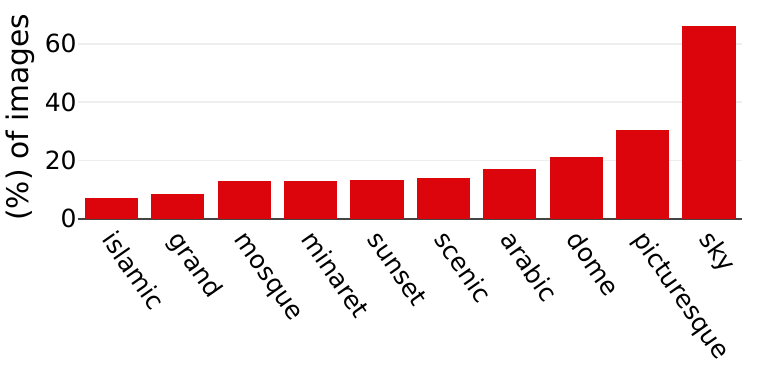}
     \end{subfigure}
     \hfill
     \begin{subfigure}[b]{\linewidth}
         \centering
         \includegraphics[width=0.7\textwidth]{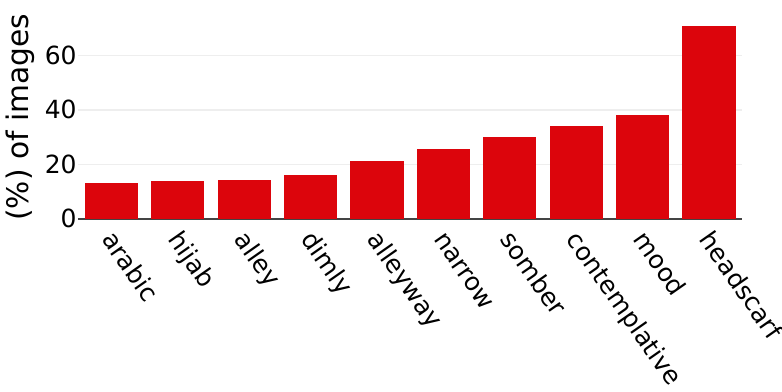}
     \end{subfigure}
     \hfill
     \begin{subfigure}[b]{\linewidth}
         \centering
         \includegraphics[width=0.7\textwidth]{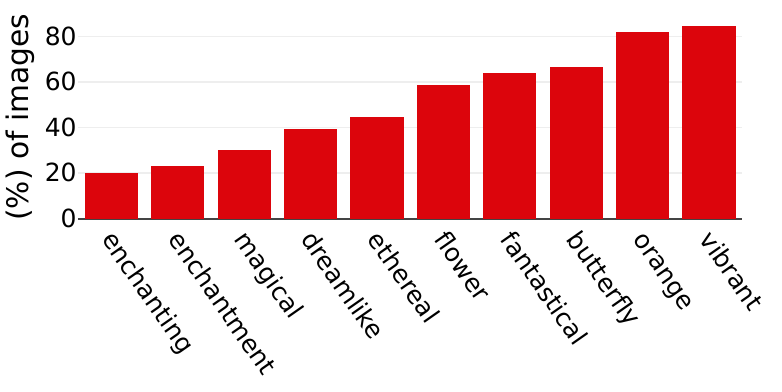}
     \end{subfigure}
     \hfill
     \begin{subfigure}[b]{\linewidth}
         \centering
         \includegraphics[width=0.7\textwidth]{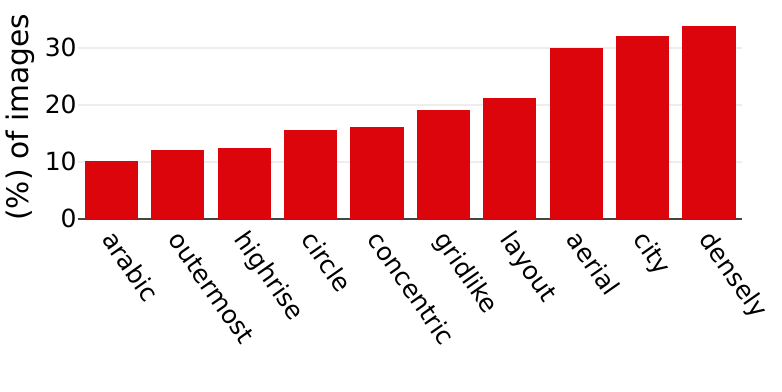}
     \end{subfigure}
     \hfill
     \begin{subfigure}[b]{\linewidth}
         \centering
         \includegraphics[width=0.7\textwidth]{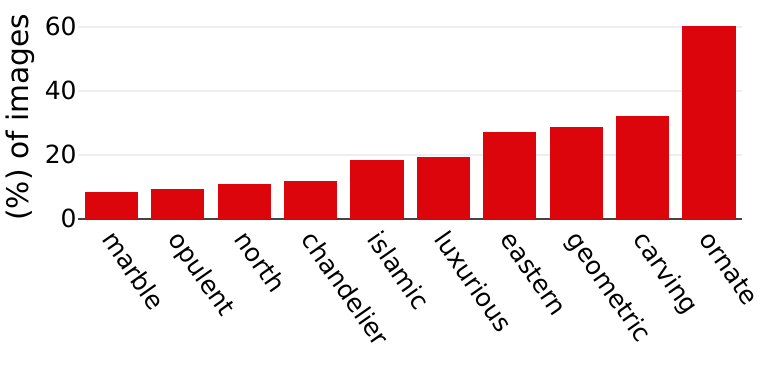}
     \end{subfigure}
     \hfill
     \begin{subfigure}[b]{\linewidth}
         \centering
         \includegraphics[width=0.7\textwidth]{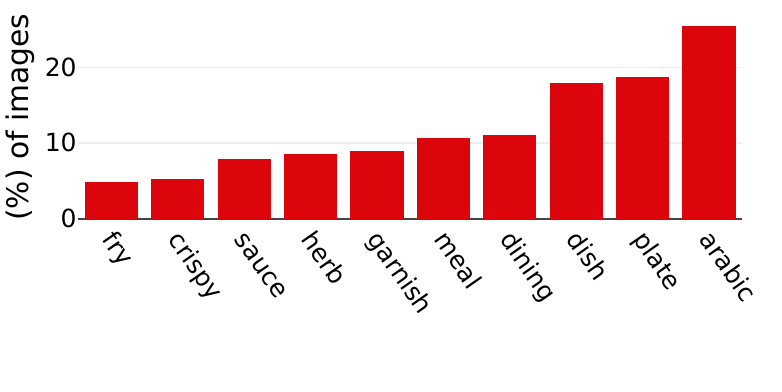}
     \end{subfigure}
    \caption{VQA results: We present the top 15 terms for images generated using input prompts in Arabic.}
    \label{fig:vqa_ar}
\end{figure}

\begin{figure}[h]
    \centering
    \begin{subfigure}[b]{\linewidth}
         \centering
         \includegraphics[width=0.7\textwidth]{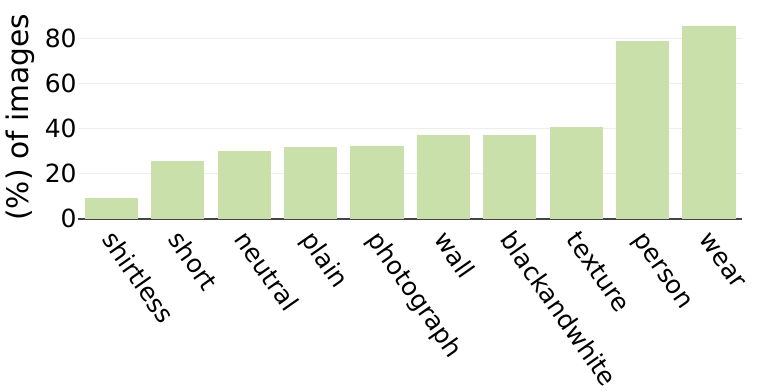}
     \end{subfigure}
     \hfill
     \begin{subfigure}[b]{\linewidth}
         \centering
         \includegraphics[width=0.7\textwidth]{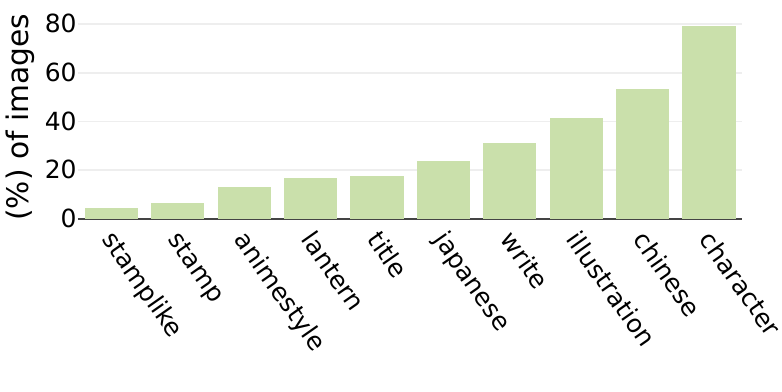}
     \end{subfigure}
     \hfill
     \begin{subfigure}[b]{\linewidth}
         \centering
         \includegraphics[width=0.7\textwidth]{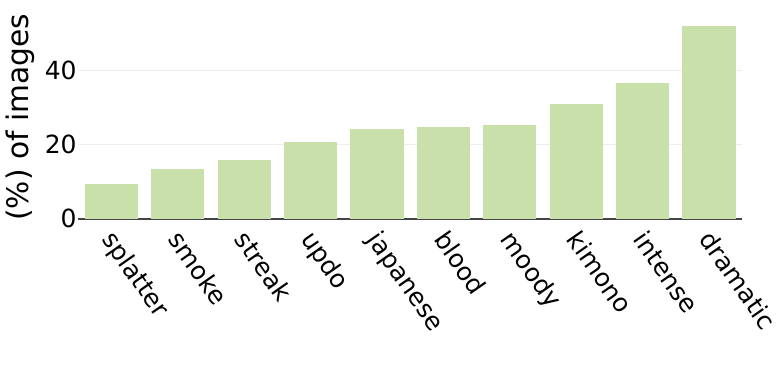}
     \end{subfigure}
     \hfill
     \begin{subfigure}[b]{\linewidth}
         \centering
         \includegraphics[width=0.7\textwidth]{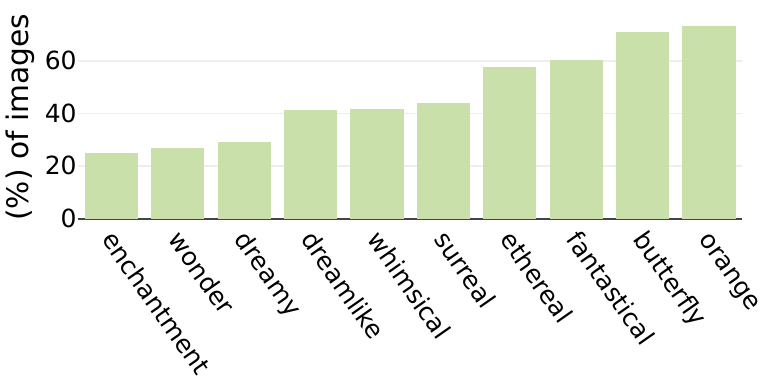}
     \end{subfigure}
     \hfill
     \begin{subfigure}[b]{\linewidth}
         \centering
         \includegraphics[width=0.7\textwidth]{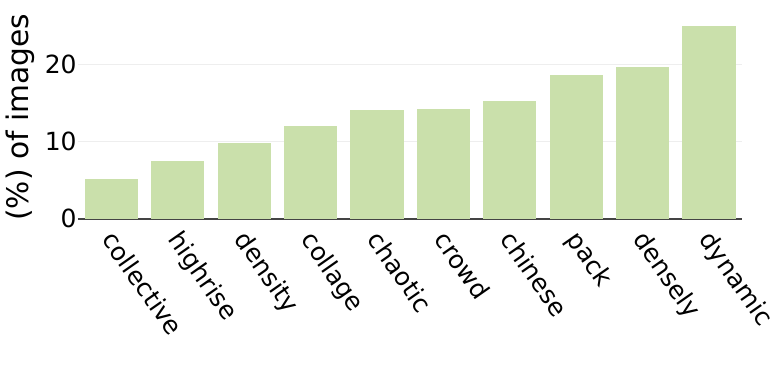}
     \end{subfigure}
     \hfill
     \begin{subfigure}[b]{\linewidth}
         \centering
         \includegraphics[width=0.7\textwidth]{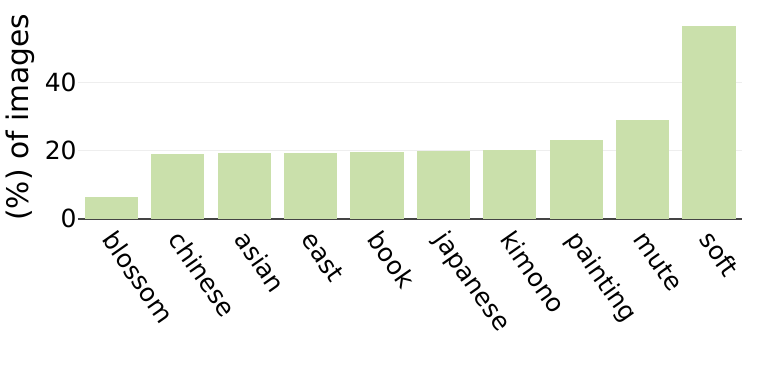}
     \end{subfigure}
     \hfill
     \begin{subfigure}[b]{\linewidth}
         \centering
         \includegraphics[width=0.7\textwidth]{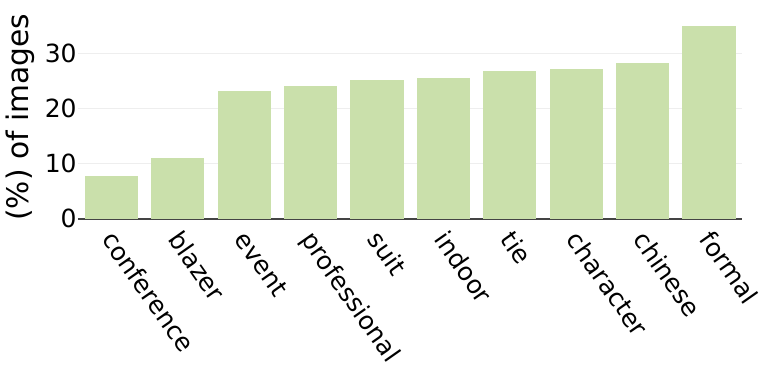}
     \end{subfigure}
    \caption{VQA results: We present the top 15 terms for images generated using input prompts in Chinese.}
    \label{fig:vqa_zh}
\end{figure}

\begin{figure}[h]
    \centering
    \begin{subfigure}[b]{\linewidth}
         \centering
         \includegraphics[width=0.7\textwidth]{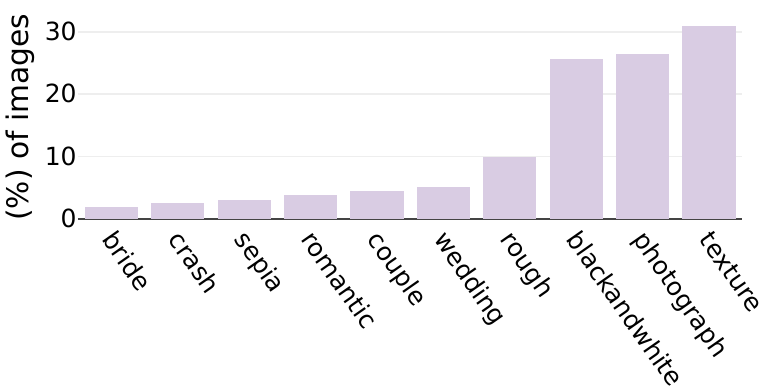}
     \end{subfigure}
     \hfill
     \begin{subfigure}[b]{\linewidth}
         \centering
         \includegraphics[width=0.7\textwidth]{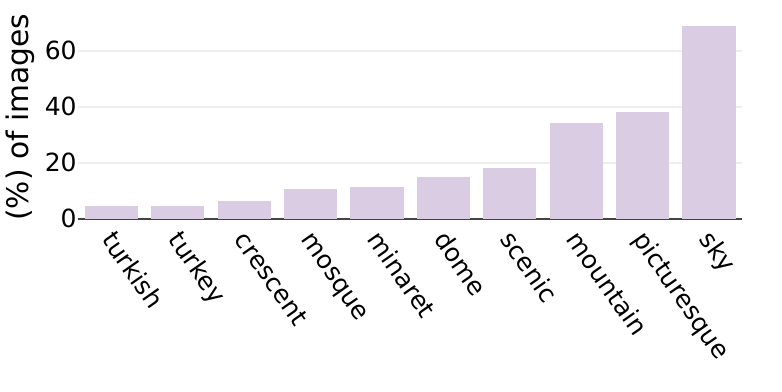}
     \end{subfigure}
     \hfill
     \begin{subfigure}[b]{\linewidth}
         \centering
         \includegraphics[width=0.7\textwidth]{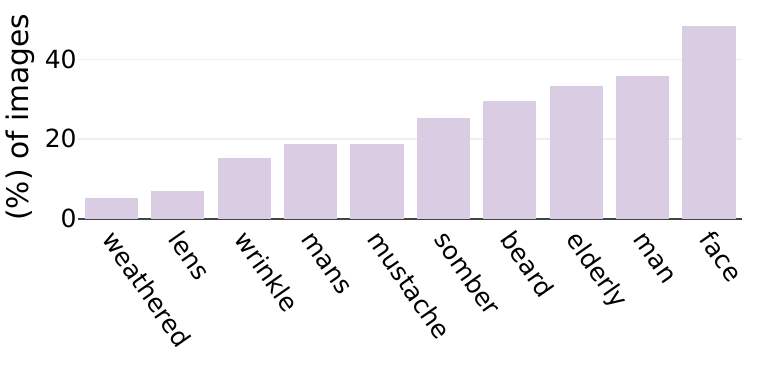}
     \end{subfigure}
     \hfill
     \begin{subfigure}[b]{\linewidth}
         \centering
         \includegraphics[width=0.7\textwidth]{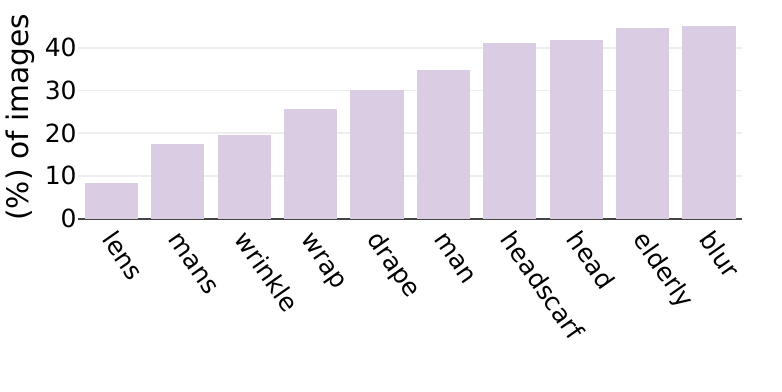}
     \end{subfigure}
     \hfill
     \begin{subfigure}[b]{\linewidth}
         \centering
         \includegraphics[width=0.7\textwidth]{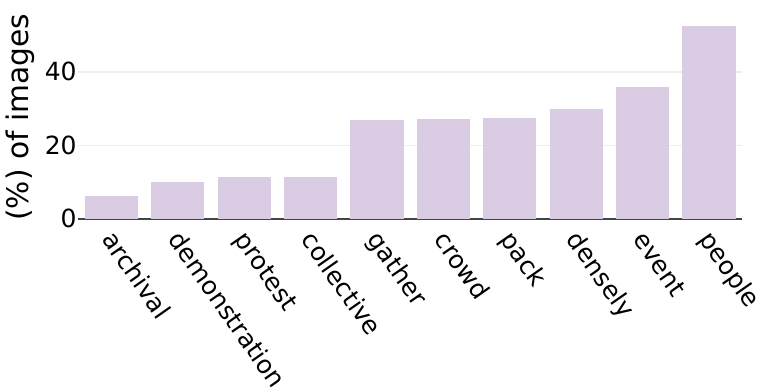}
     \end{subfigure}
     \hfill
     \begin{subfigure}[b]{\linewidth}
         \centering
         \includegraphics[width=0.7\textwidth]{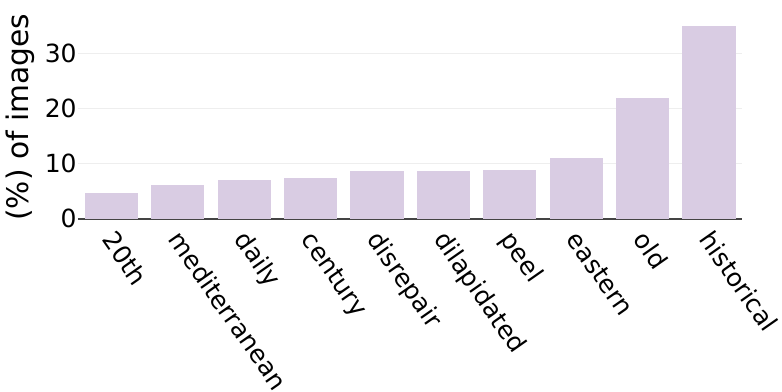}
     \end{subfigure}
     \hfill
     \begin{subfigure}[b]{\linewidth}
         \centering
         \includegraphics[width=0.7\textwidth]{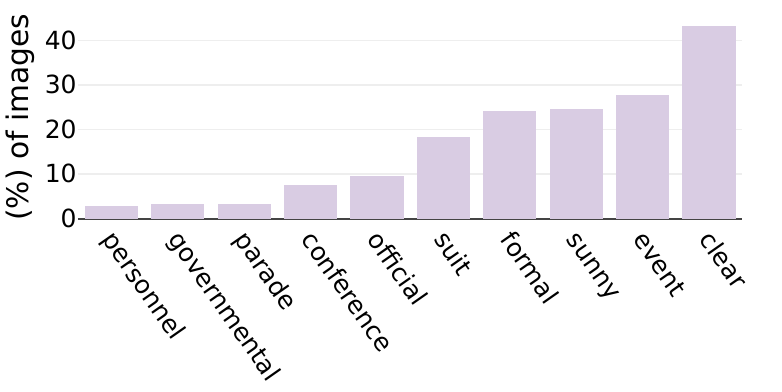}
     \end{subfigure}
    \caption{VQA results: We present the top 15 terms for images generated using input prompts in Turkish.}
    \label{fig:vqa_tr}
\end{figure}

\begin{figure}[h]
    \centering
    \begin{subfigure}[b]{\linewidth}
         \centering
         \includegraphics[width=0.7\textwidth]{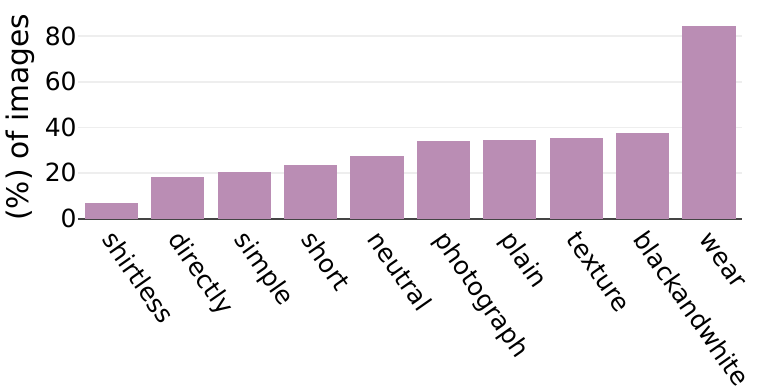}
     \end{subfigure}
     \hfill
     \begin{subfigure}[b]{\linewidth}
         \centering
         \includegraphics[width=0.7\textwidth]{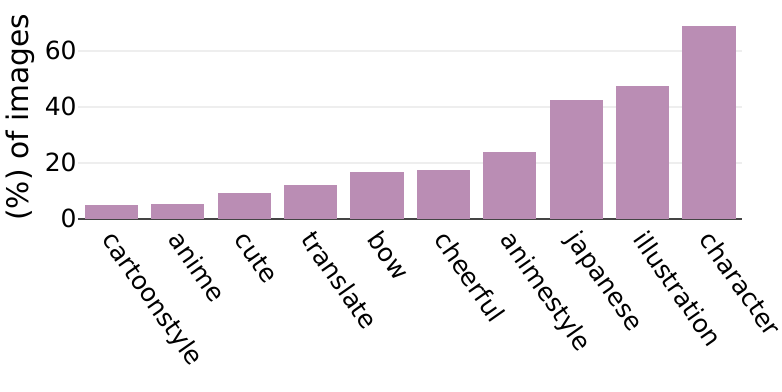}
     \end{subfigure}
     \hfill
     \begin{subfigure}[b]{\linewidth}
         \centering
         \includegraphics[width=0.7\textwidth]{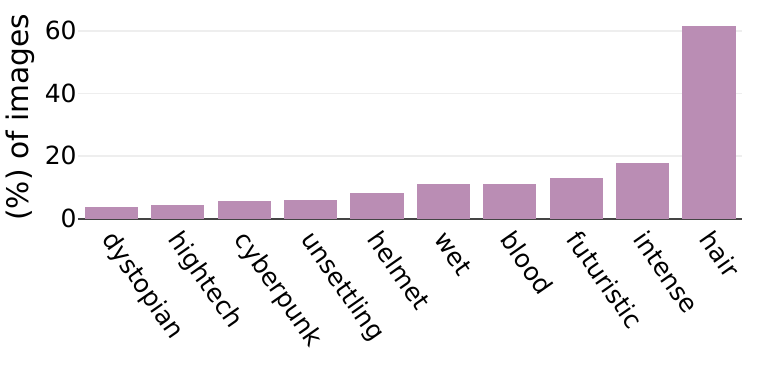}
     \end{subfigure}
     \hfill
     \begin{subfigure}[b]{\linewidth}
         \centering
         \includegraphics[width=0.7\textwidth]{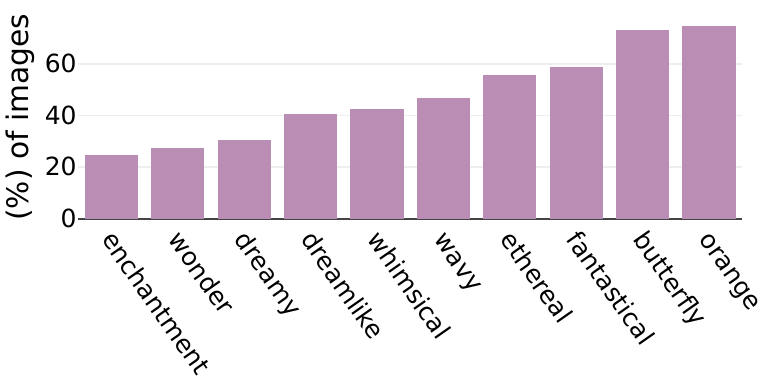}
     \end{subfigure}
     \hfill
     \begin{subfigure}[b]{\linewidth}
         \centering
         \includegraphics[width=0.7\textwidth]{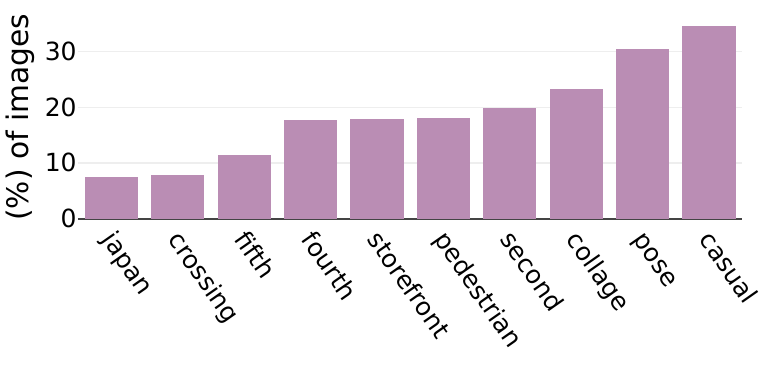}
     \end{subfigure}
     \hfill
     \begin{subfigure}[b]{\linewidth}
         \centering
         \includegraphics[width=0.7\textwidth]{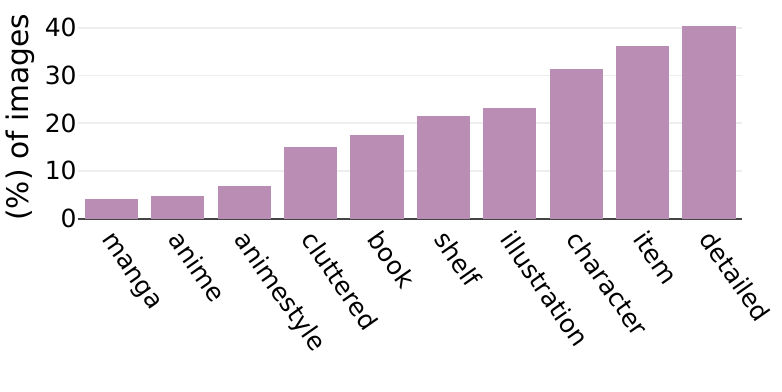}
     \end{subfigure}
     \hfill
     \begin{subfigure}[b]{\linewidth}
         \centering
         \includegraphics[width=0.7\textwidth]{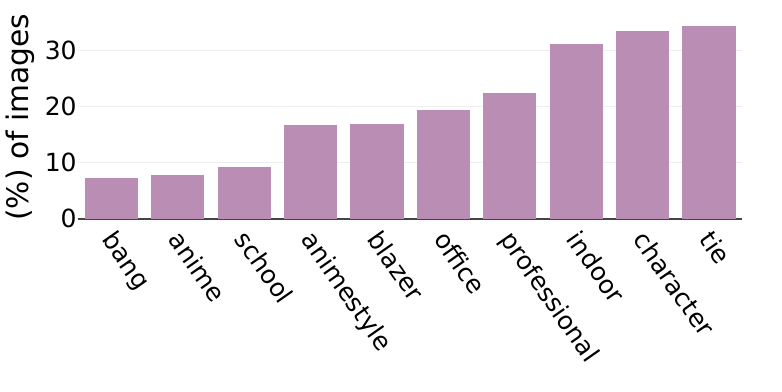}
     \end{subfigure}
    \caption{VQA results: We present the top 15 terms for images generated using input prompts in Japanese.}
    \label{fig:vqa_ja}
\end{figure}

\begin{figure}[h]
    \centering
    \begin{subfigure}[b]{\linewidth}
         \centering
         \includegraphics[width=0.7\textwidth]{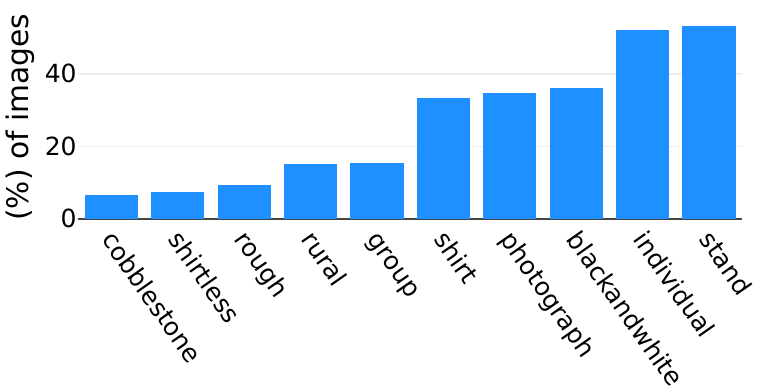}
     \end{subfigure}
     \hfill
     \begin{subfigure}[b]{\linewidth}
         \centering
         \includegraphics[width=0.7\textwidth]{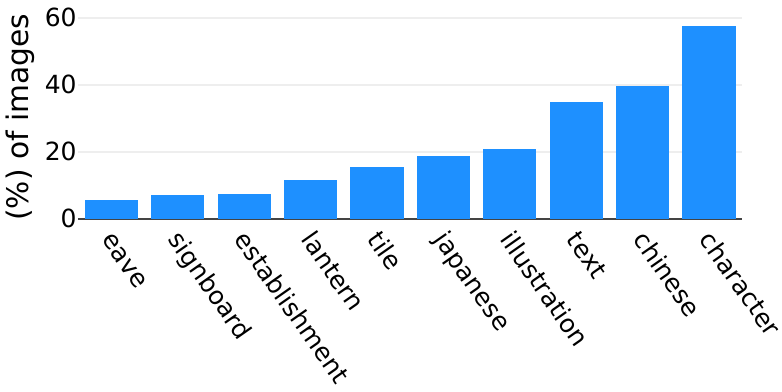}
     \end{subfigure}
     \hfill
     \begin{subfigure}[b]{\linewidth}
         \centering
         \includegraphics[width=0.7\textwidth]{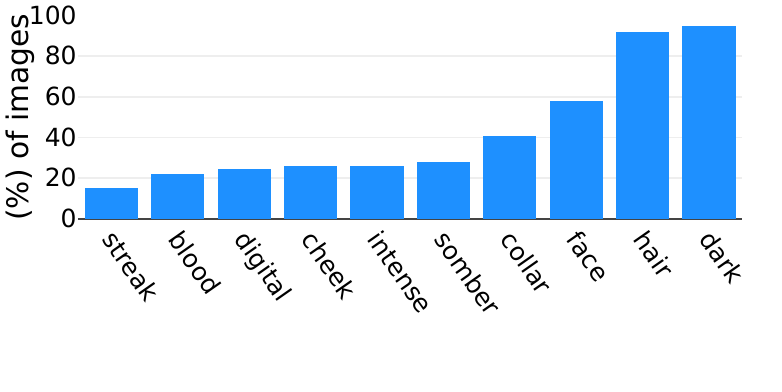}
     \end{subfigure}
     \hfill
     \begin{subfigure}[b]{\linewidth}
         \centering
         \includegraphics[width=0.7\textwidth]{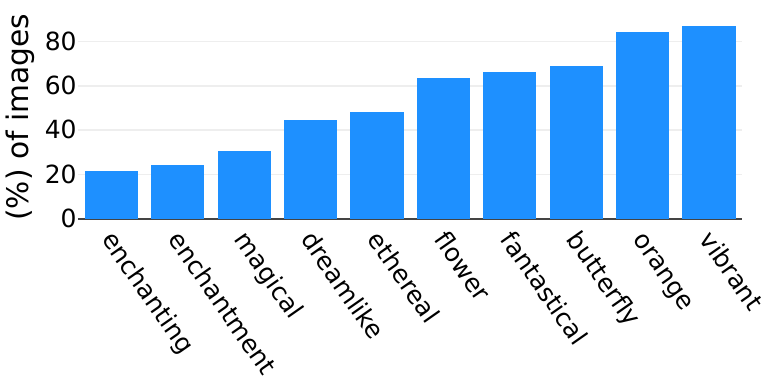}
     \end{subfigure}
     \hfill
     \begin{subfigure}[b]{\linewidth}
         \centering
         \includegraphics[width=0.7\textwidth]{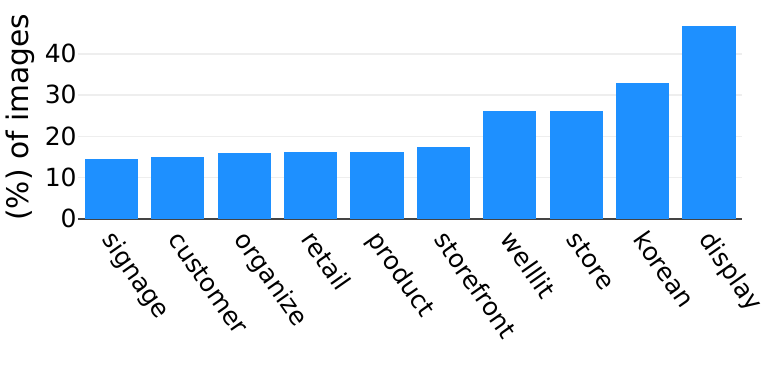}
     \end{subfigure}
     \hfill
     \begin{subfigure}[b]{\linewidth}
         \centering
         \includegraphics[width=0.7\textwidth]{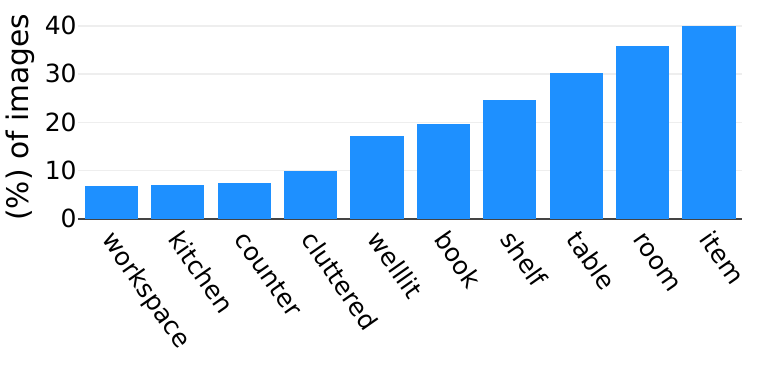}
     \end{subfigure}
     \hfill
     \begin{subfigure}[b]{\linewidth}
         \centering
         \includegraphics[width=0.7\textwidth]{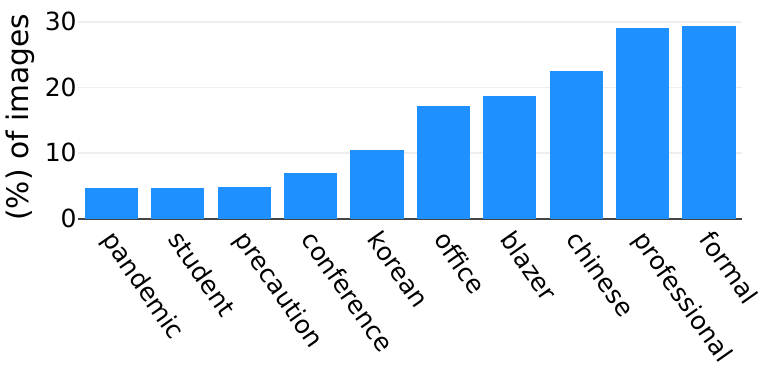}
     \end{subfigure}
    \caption{VQA results: We present the top 15 terms for images generated using input prompts in Korean.}
    \label{fig:vqa_ko}
\end{figure}

\clearpage
\subsection{Validation Correlation of SeeGULL Terms}

\begin{table}[h]
\setlength{\tabcolsep}{4pt} 
\renewcommand{\arraystretch}{1.1} 
    \centering
    \small
    \begin{tabular}{p{0.8cm}|p{4.5cm}|p{0.6cm}|p{8.5cm}}
    \toprule
    \textbf{Lang.} & \textbf{Associated Culture(s)} & \textbf{\#} &\textbf{Visual Stereotypes} \\
    \midrule
       zh & Chinese & 22 &  red, strong, china, yellow, \hlr{rice}, \hlr{tire}, bright, clear, \hlr{angry}, thin, short, blonde, tea, girl, beautiful, \hlr{brunette}, chinese, untidy, left, \hlr{drug}, dirty \\
       \midrule
     fi & Finnish &  1 & \hlr{sauna} \\
     \midrule
     ko & North Korean, South Korean & 6 & athletic, \hlr{taekwondo}, thin, kimchi, yellow, short \\
     \midrule
     ja & Japanese & 21 & attractive, sushi, sad, dog, rice, \hlr{sex}, tire, \hlr{sick}, \hlr{tsunami}, japanese, samurai, thin, short, car, monkey, \hlr{nervous}, robot, ninja, organize, white, fish \\
     \midrule
     hi & Indian & 35 & athletic, cow, attract, sad, watch, \hlr{thick}, red, \hlr{unorganized}, light, cricket, \hlr{hockey}, violent, elephant, open, snake, brown, \hlr{tall}, black, indian, \hlr{angry}, dark, organized, thin, old, beautiful, \hlr{sweaty}, \hlr{nervous}, blonde, curry, coast, untidy, \hlr{vegetarian}, white, colorful, dirty\\
     \midrule
     ru & Russian & 9 & \hlr{sauce}, blonde, snow, \hlr{vodka}, russian, drinking, \hlr{malnourished}, \hlr{strong}, unwashed\\
     \midrule
     ar & Bahraini, Egyptian, Jordanian, Kuwaiti, Lebanese, Libyan, Mauritanian, 
           Omani, Palestinian, Qatari, Saudi Arabian, Syrian, Tunisian, Emiratis, Yemeni & 14 & \hlr{kidnapping}, oil, dark, \hlr{malnourished}, violent, \hlr{unorganized}, pyramid, tall, hairy, \hlr{sexy}, barren, wet, dirty, beautiful\\
           \midrule
     tr & Turkish & 8 & \hlr{greasy}, \hlr{angry}, \hlr{fiery}, \hlr{violent}, untidy, organized, tall, \hlr{sexy} \\
     \midrule
     am & Ethiopian & 26 & \hlr{attractive}, zoo, thick, \hlr{torture}, strong, \hlr{die}, \hlr{coffee}, bald, tall, \hlr{sick}, black, child, \hlr{malnourished}, dark, \hlr{poor}, \hlr{underweight}, thin, village, short, \hlr{cannibal}, blonde, \hlr{poverty}, \hlr{undernourished}, \hlr{malnutrition}, \hlr{scrawny}, dirty \\  
     \bottomrule
 \end{tabular}
    \caption{Visual stereotypes of the SeeGULL dataset. We merge all cultural stereotypes of countries where the language is the official language, remove multi-token examples and non visually depictable adjectives. Highlighted red terms are terms that are never matched within our analysis. }
    \label{tab:seagull_terms}
\end{table}

\onecolumn

\section{Example Images} \label{app:example_images}
In this section, we present example images generated by the T2I models. The first subsection shows images generated from different text-encoder representations of an input prompt. The second subsection presents examples spanning multiple languages, models, and prompted cultural identities.

\subsection{Examples of layer-wise images}
One limitation we set out to address with SoS, relative to CLIPScore, is the reliance on a textual anchor. Text-based matching presupposes both high-quality images and faithful realization of the captioned concepts. These assumptions break down for generations guided by early layers of the text encoder, which often yield low-quality outputs or random visual noise. In contrast, SoS operates directly on images without requiring a caption anchor. To illustrate the issue, we show two example generations produced from different layers of the text encoder for the concept ``Baharini man''. The early layer outputs lack coherent semantics, so image-to-caption alignment is uninformative, whereas SoS remains applicable.

\begin{figure}[h]
    \centering
    \begin{subfigure}[t]{0.5\textwidth}
        \centering
        \includegraphics[width=0.8\linewidth]{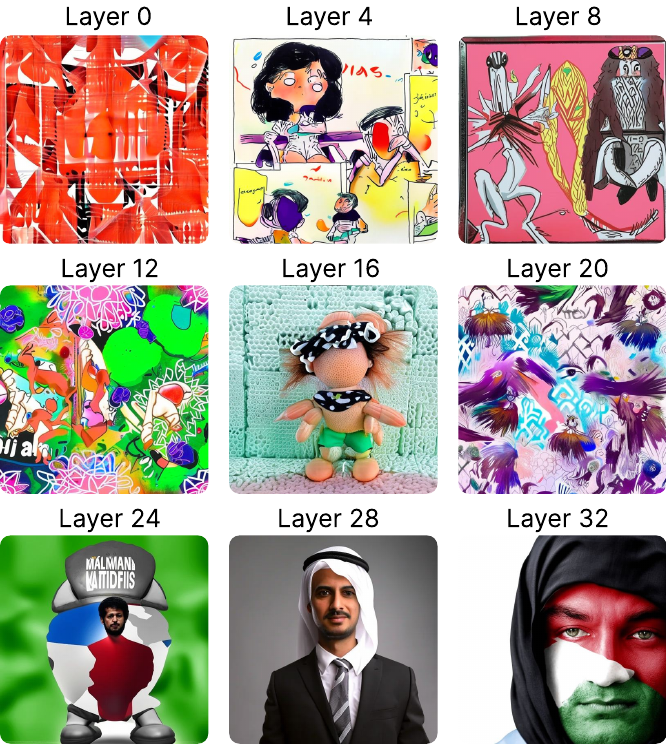}
        \caption{Per layer images generated by K3.}
    \end{subfigure}%
    ~ 
    \begin{subfigure}[t]{0.5\textwidth}
        \centering
        \includegraphics[width=0.8\linewidth]{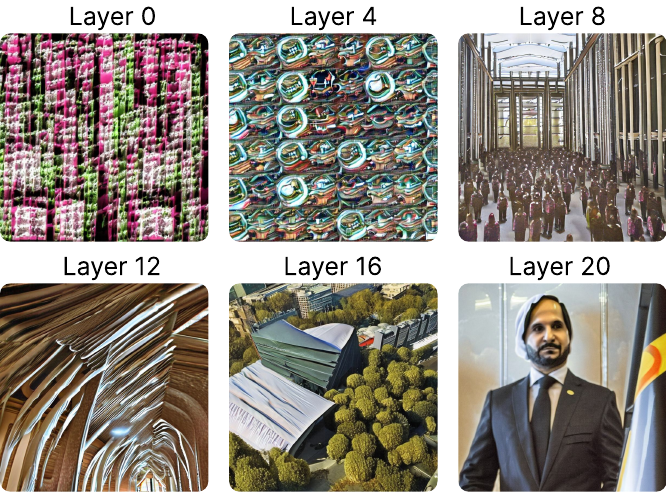}
    \caption{Per layer images generated by SD21.}
    \end{subfigure}
    \caption{Example images generated from successive text-encoder layers by K3 (a) and SD21 (b) for the prompt ``Baharini man''. For both T2I models, early-layer representations exhibit low visual quality and weak semantic coherence.}
\end{figure}

\subsection{Example images generated by T2I models}

\begin{figure}[htbp]
    \centering
    \includegraphics[width=0.85\textwidth]{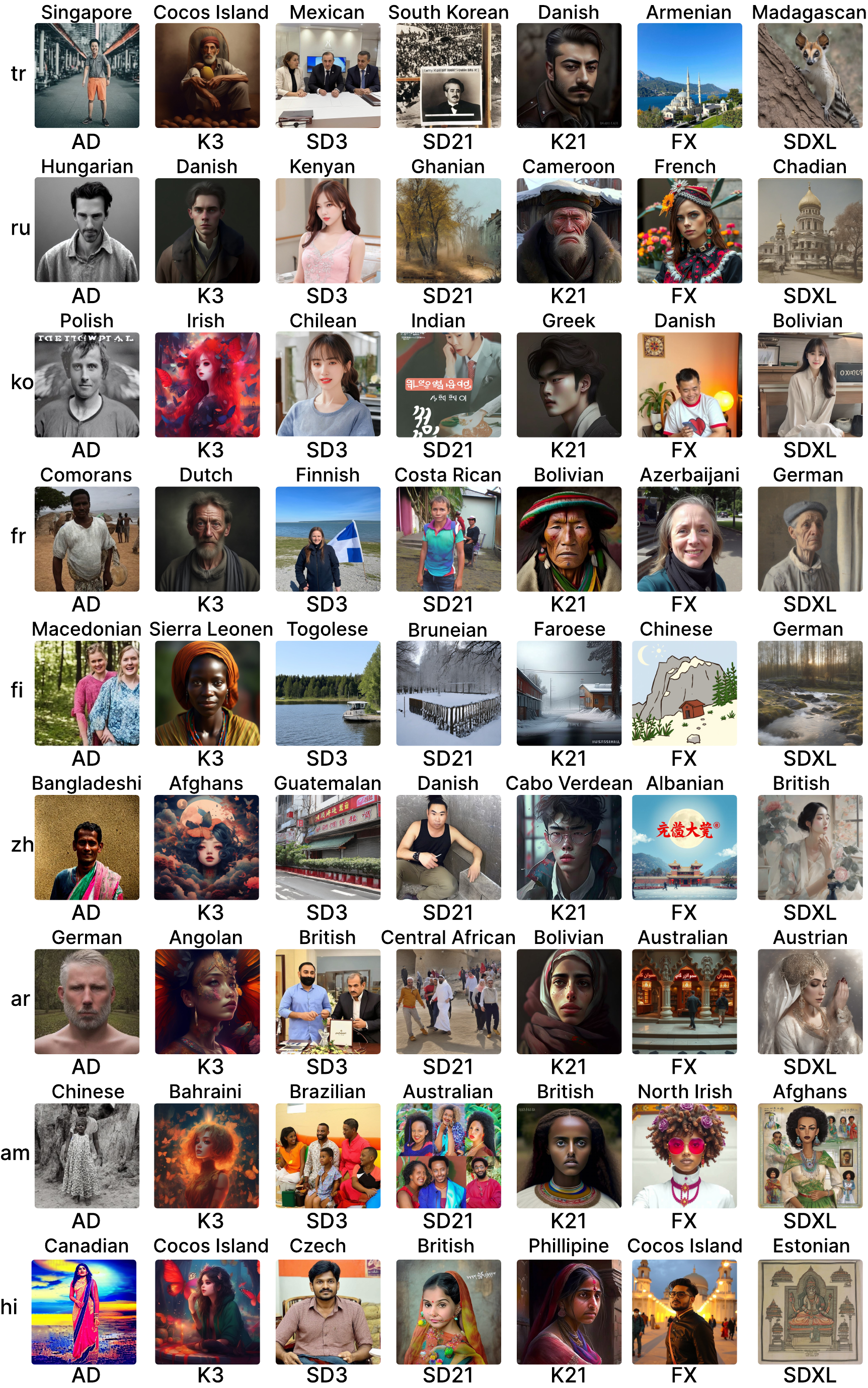}
    \caption{Example images across languages and models. Above each image, we indicate the target cultural identity to be generated; below each image, the T2I model used; and on the left, the language code of the input prompt.}
    \label{fig:example_images}
\end{figure}

\clearpage


\end{document}